\documentclass[manuscript]{acmart}

\usepackage{amsmath,amsfonts}
\usepackage{algorithmic}
\usepackage{algorithm}
\usepackage{array}
\usepackage[caption=false,font=normalsize,labelfont=sf,textfont=sf]{subfig}
\usepackage{textcomp}
\usepackage{stfloats}
\usepackage{url}
\usepackage{verbatim}
\usepackage{graphicx}
\usepackage{color}
\usepackage{booktabs}
\usepackage{enumitem}
\usepackage{tabularx} % Add this in the preamble
\usepackage{multirow}
\usepackage{makecell}
\usepackage[T1]{fontenc}
\usepackage{overpic}
\usepackage{hyperref}
\usepackage{url}
\usepackage{adjustbox}
\usepackage{pifont}
\usepackage{cellspace}
\usepackage{colortbl}
\usepackage{wrapfig}

\definecolor{myblue}{RGB}{16, 78, 139}
\usepackage{multicol}
\ifodd 1
\newcommand{\com}[1]{\textbf{\color{red}(COMMENT: #1)}} %comment of the 
\else

\newcommand{\com}[1]{}
\fi

\usepackage{url} 
\definecolor{ptcolor_new}{RGB}{212, 128, 1}
\definecolor{rscolor_new}{RGB}{111, 177, 61}

\newcommand{\why}[1]{{\color{black}{#1}}}
\newcommand{\mpt}[1]{{\color{black}{#1}}}
\newcommand{\grs}[1]{{\color{black}{#1}}}

\def\fig{Fig.}

\def\eg{e.g.}

\AtBeginDocument{%
  }

\setcopyright{acmcopyright}
\renewcommand\footnotetextcopyrightpermission[1]{}
\acmJournal{JACM}
\acmVolume{37}
\acmNumber{4}
\acmArticle{111}
\acmMonth{8}

\begin{document}

% \title{Towards Mobile Sensing with Event Cameras on High-agility Resource-constrained Platforms: A Survey}
% \title{Event Camera Meets Resource-Constrained Mobile Computing: \\ Abstraction, Algorithms, Acceleration, and Applications}
% \title{Event Camera Meets Resource-Aware Mobile Computing: \\Abstraction, Algorithm, Acceleration, Application}
% \title{Event Camera Meets Mobile Embodied Perception: \\ Abstraction, Algorithm, Acceleration, Application}
\title[Event Camera Meets Mobile Embodied Perception: Abstraction, Algorithm, Acceleration, Application]{Event Camera Meets Mobile Embodied Perception: \\ Abstraction, Algorithm, Acceleration, Application}
% Event Camera Meets Resource-Aware Mobile Computing: Abstraction,Algorithm,Acceleration,Application 


\author{Haoyang Wang, Ruishan Guo, Pengtao Ma, Ciyu Ruan, Xinyu Luo, Wenhua Ding, Tianyang Zhong}
% \email{haoyang-22@mails.tsinghua.edu.cn}
\affiliation{%
  \institution{Shenzhen International Graduate School, Tsinghua University}
  \city{Shenzhen}
  \country{China}
}

% \author{Haoyang Wang}
% \affiliation{%
%   \institution{Shenzhen International Graduate School, Tsinghua University}
%   \city{Shenzhen}
%   \country{China}
% }
% \email{haoyang-22@mails.tsinghua.edu.cn}

% \author{Ruishan Guo}
% \email{grs24@mails.tsinghua.edu.cn}
% \affiliation{%
%   \institution{Shenzhen International Graduate School, Tsinghua University}
%   \city{Shenzhen}
%   \country{China}
% }

% \author{Pengtao Ma}
% \email{mpt25@mails.tsinghua.edu.cn}
% \affiliation{%
%   \institution{Shenzhen International Graduate School, Tsinghua University}
%   \city{Shenzhen}
%   \country{China}
% }

% \author{Ciyu Ruan}
% \email{rcy23@mails.tsinghua.edu.cn}
% \affiliation{%
%   \institution{Shenzhen International Graduate School, Tsinghua University}
%   \city{Shenzhen}
%   \country{China}
% }

% \author{Xinyu Luo}
% \email{luo-xy23@mails.tsinghua.edu.cn}
% \affiliation{%
%   \institution{Shenzhen International Graduate School, Tsinghua University}
%   \city{Shenzhen}
%   \country{China}
% }

% \author{Wenhua Ding}
% \email{dingwh24@mails.tsinghua.edu.cn}
% \affiliation{%
%   \institution{Shenzhen International Graduate School, Tsinghua University}
%   \city{Shenzhen}
%   \country{China}
% }

% \author{Tianyang Zhong}
% \email{zhongty25@mails.tsinghua.edu.cn}
% \affiliation{%
%   \institution{Shenzhen International Graduate School, Tsinghua University}
%   \city{Shenzhen}
%   \country{China}
% }

\author{Jingao Xu}
% \email{jingaoxu@hku.hk}
\affiliation{%
  \institution{The University of Hong Kong}
  \city{Hong Kong}
  \country{China}
}

\author{Yunhao Liu}
% \email{yunhao@tsinghua.edu.cn}
\affiliation{%
  \institution{Global Innovation Exchange, Tsinghua University}
  \city{Beijing}
  \country{China}
}

\author{Xinlei Chen}
\email{chen.xinlei@sz.tsinghua.edu.cn}
\authornote{Xinlei Chen is the corresponding author}
\affiliation{%
  \institution{Shenzhen International Graduate School, Tsinghua University}
  \country{China}
}

\thanks{This paper was supported by the Natural Science Foundation of China under Grant 62371269, Guangdong Innovative and Entrepreneurial Research Team Program 2021ZT09L197, Shenzhen Low-Altitude Airspace Strategic Program Portfolio Z253061 and Meituan Academy of Robotics Shenzhen.
% This paper was supported by the Low-Altitude Airspace Strategic Program Portfolio No. Z253061, 
% the National Key R\&D program of China No. 2022YFC3300703,
% Natural Science Foundation of China under Grant 62371269, 
% Guangdong Innovative and Entrepreneurial Research Team Program (2021ZT09L197), Meituan Academy of Robotics Shenzhen.
}

\renewcommand{\shortauthors}{Wang, H. et al.}

\begin{abstract}
With the evolution of mobile embodied intelligence, agents such as drones and autonomous robots are transitioning toward high agility. 
This shift imposes stringent demands on embodied perception, requiring high-accuracy and low-latency feedback loops for reliable interaction. 
Event-based vision has emerged as a transformative paradigm.
Its microsecond-level temporal resolution and high dynamic range render it ideal for embodied perception tasks on high-agility mobile platforms.
However, asynchronous nature, substantial noise, lack of persistent semantic information, and large data volume pose challenges for processing on resource-constrained mobile agents. 
This paper surveys the literature from 2014-2025 and presents a comprehensive overview of event-based mobile embodied perception.
We organize review around four key pillars: event \textit{abstraction} methods, perception \textit{algorithm} advancements, hardware and software \textit{acceleration} strategies, and mobile \textit{applications}. 
We discuss critical tasks including visual odometry, object tracking, optical flow, and 3D reconstruction, while highlighting challenges associated with sensor fusion and real-time deployment. 
Furthermore, we outline future research directions, such as improving event cameras with advanced optics and leveraging neuromorphic computing for efficient processing.
To support ongoing research, we provide an open-source \textit{Online Sheet} with recent developments. 
We hope this survey serves as a reference, facilitating adoption of event-based vision across diverse mobile embodied applications.
\end{abstract}



% \received{17 May 2025}
% \received[revised]{8 September 2025}

\maketitle

\vspace{-0.2cm}
\section{Introduction}

\begin{figure}[t]
    \centering
        \includegraphics[width=1\columnwidth]{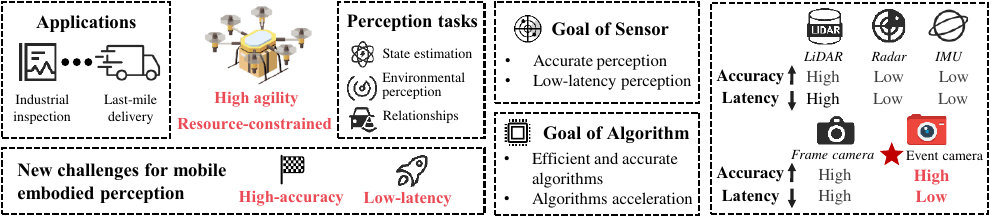}
    \vspace{-0.7cm}
     \caption{
     Mobile agents are used in various applications, with key tasks including state estimation, environment perception, and understanding agent-environment interactions. As agents become more agile, mobile perception faces higher demands for accuracy and latency. This requires tight coordination between sensors and algorithms: \textit{(i)} sensors must capture high-precision data with minimal delay; \textit{(ii)} algorithms must efficiently process data within resource constraints. Traditional sensors fall short of these needs, whereas event cameras, capable of asynchronously capturing pixel-level intensity changes with microsecond latency, offer transformative potential. This survey presents a comprehensive review of event cameras and development of efficient algorithms.
     % designed for them.
     % Mobile agents are employed in a wide range of applications. 
     % Among these, the most critical steps involve estimating the agents' states, perceiving surrounding environmental structures, and understanding the relationships between agents and their environment. 
     % As mobile agents advance toward high-agility design, these mobile platforms impose new demands on mobile perception, particularly in terms of high accuracy and low latency. 
     % This necessitates close coordination between sensors and perception algorithms: $(i)$ Sensors must acquire high-accuracy data with low latency; $(ii)$ perception algorithms must process the data efficiently and accurately under constrained resources. 
     % Existing sensors fail to fully meet these requirements; Event cameras, capable of asynchronously capturing pixel-level intensity changes with $\mu s$-level latency, hold the potential to revolutionize mobile perception. 
     % This survey provides a comprehensive review of event cameras and the development of efficient and accurate algorithms tailored to them.
     }
    \label{intro}
    \vspace{-0.5cm}
\end{figure} 

% \\ \role{haoyang}
% \textbf{Mobile perception.} 
% With the ongoing advancements in sensor technology and the proliferation of sophisticated computing capabilities within embedded systems, mobile agents (\eg, drones and autonomous vehicles) have emerged as some of the most groundbreaking innovations in recent years \cite{ren2025safety, zhou2022swarm, wang2025ultra}.
% As illustrated in \fig \ref{intro}, these agents are increasingly being employed in a variety of novel applications, including last-mile delivery \cite{wang2022micnest}, and industrial inspection \cite{xu2022swarmmap, wang2024transformloc, liu2024mobiair}, within the context of smart city scenarios.
% To accomplish the aforementioned tasks, which require extensive interaction with the external environment, these mobile agents must be equipped with the capability to $(i)$ \textit{perceive their own state} \cite{chen2023self, xu2020edge, li2024edgeslam2}, $(ii)$ \textit{comprehend their surroundings} \cite{cui2024alphalidar, he2023vi}, and $(iii)$ \textit{understand their relationships to the environment} \cite{he2022automatch}.
% This includes awareness of their own pose (\eg, location and orientation), environmental structure and the spatial-temporal relationships between mobile agents and objects within the environment \cite{cui2024vilam}. 
% Achieving these goals has garnered widespread interest within the mobile computing community.
% \textbf{Mobile perception.}

% \textbf{Mobile embodied perception.}
In the era of embodied AI, mobile agents, ranging from drones and autonomous vehicles to humanoid robots, are expected to operate autonomously in the wild \cite{chen2024ddl,zhou2022swarm, jian2023path}. 
Unlike passive perception systems, these embodied agents must actively perceive their environment to make real-time decisions, forming a tight coupling between perception and action~\cite{ren2025safety, wang2025ultra}.
% With ongoing advancements in sensor technology and the proliferation of sophisticated computing capabilities within embedded systems, mobile agents (\eg, drones and autonomous vehicles) have emerged as the most groundbreaking innovations in recent years .  
As illustrated in \fig \ref{intro}, these agents are increasingly deployed in a variety of applications, including last-mile delivery \cite{wang2025ultra}, industrial inspection \cite{wang2024transformloc, xu2022swarmmap, liu2024mobiair}, rapid relief-and-rescue \cite{chen2024soscheduler}, aerial imaging \cite{leng2024recent} and sky networking \cite{xu2024scalable}, particularly within smart city scenarios.  
To perform these tasks, which require extensive interaction with the external environment, mobile agents must possess the ability to:  
$(i)$ \textit{perceive their own state}, including location and orientation \cite{wang2025aerial, chen2023self, wang2022h},  
$(ii)$ \textit{comprehend their surroundings}, (\eg, environmental structure and map) \cite{cui2024alphalidar, zhao2025flight}, and  
$(iii)$ \textit{understand their relationship with the environment}, such as the spatio-temporal relationships between mobile agents and objects within their surroundings \cite{he2022automatch, wang2025enabling}.  
% This entails awareness of their own pose (\eg, location and orientation), environmental structure, and the spatiotemporal relationships between mobile agents and objects within their surroundings \cite{cui2024vilam}. 
Achieving these capabilities has become a focal point of interest within mobile embodied intelligence community.  

In the context of smart city scenarios, the demand for mobile agents is progressively increasing, with their mission profiles evolving toward execution of 4D tasks, which are characterized as deep, dull, dangerous, and dirty \cite{francis2022gas}. 
The expanding scale of cities also necessitates that mobile agents complete various tasks within shorter time frames, driving their evolution toward high-speed operation \cite{ren2025safety, floreano2015science}.
Consequently, the development of mobile agents is exhibiting new trends toward high agility.
% As cities expand, mobile agents must complete an increasing variety of tasks within shorter time frames, driving advancements in high-speed operation \cite{ren2025safety, loquercio2021learning, floreano2015science}.
% This shift has led to a growing emphasis on high agility in mobile device development.
For instance, DJI’s industrial inspection drones can cruise at speeds of 21 $m/s$ \cite{DJIMAVIC4}, while Wing’s delivery drones fly at 30 $m/s$ to deliver packages \cite{Wing}.

% $(ii)$ \textit{Efficient design}.
% The complex operational environment of cities demands that the design of mobile agents balances performance and efficiency, prompting the development of mobile agents toward lightweight and efficient designs \cite{shen2024sunlight, zhou2022swarm, bai2024agile}.
% For example, Meituan's delivery drone manages to maintain a weight of less than 10$kg$ \cite{he2023acoustic}, and DJI's professional photography drone, the Mavic series, achieves a weight of less than 1.5$kg$ \cite{DJICompare}. 
% These specifications underscore the focus on enhancing accuracy and efficiency of perception to meet the demands of complex urban scenarios.
% and the execution of 4D tasks

% \textbf{New challenges for mobile embodied perception.}
% As mobile agents evolve toward high-agility operation and efficient design, mobile perception is required to advance toward \textit{high-accuracy} and \textit{high-efficiency}, enabling these \textit{resource-constrained} mobile agents to perceive their state and surroundings more efficiently, thereby facilitating faster responses and adjustments.  
% This progression sets new objectives for the sensors and data processing algorithms that execute mobile perception tasks:  
% $(i)$ For sensors, it is essential to acquire higher-precision raw data with lower latency.  
% $(ii)$ For perception algorithms, efficient processing of raw data is required to achieve greater accuracy in mobile perception tasks while optimizing performance for resource-limited platforms.
As mobile agents evolve toward high agility design, mobile embodied perception is required to advance toward \textit{high accuracy} and \textit{low latency}, enabling these mobile agents to perceive their state and surroundings in $millimeter$-level accuracy with $millisecond$-level latency, thereby facilitating faster responses and more precise adjustments.  
% This progression establishes new objectives for the sensors and data processing algorithms that execute mobile perception tasks:  
This evolution establishes new objectives for the sensors and data processing algorithms involved in mobile embodied perception tasks:  
$(i)$ \textit{On the sensor input front}, it is essential to acquire higher accuracy raw data with lower latency.  
$(ii)$ \textit{On the data processing algorithms front}, efficient processing of sensor measurements is essential to enhance accuracy in mobile perception tasks while optimizing performance on resource-constrained mobile platforms.

% \textbf{Existing sensors for mobile embodied perception.} 
% As mobile agents evolve toward high-speed operation and efficient designs, the
% However, existing works are increasingly inadequate for meeting the mobile perception' demands of high-accuracy and low-latency for environmental understanding and interaction.
% However, existing sensors are increasingly insufficient in meeting the high accuracy and low latency demands of mobile perception for environmental perception and interaction of high agility agents.
However, existing sensors increasingly struggle to meet the high accuracy and low latency demands of mobile perception, especially for environmental perception and interaction in high-agility agents.
\textit{(i) Radar-based solutions} employ sensors such as LiDAR \cite{ren2025safety, zhu2024swarm}, mmWave radar \cite{sie2024radarize}, and ultrasound radar, which emit signals and estimate distances based on reflections. These methods track distance changes to update agent positions and infer spatial relationships. However, they suffer from either high latency or limited accuracy: LiDAR achieves millimeter-level accuracy but requires point accumulation into frames at low frequencies (e.g., 10 Hz), introducing delays up to 100 ms; mmWave radar offers millisecond latency but lacks sufficient spatial resolution for millimeter accuracy.
\textit{(ii) Frame camera-based methods} use monocular \cite{zhang2022edgeis, cao2023edge} and stereo cameras \cite{qin2019a} for self-localization and environment mapping via SLAM \cite{li2024tmc_edgeslam2}. 
% Yet, these approaches are computationally demanding and limited by low temporal resolution (<30 Hz), relatively high latency (>30 ms), motion blur, and standard dynamic range (60 dB), decoupling perception from the continuous dynamics of the real world, making them inadequate for high-agility mobile platforms.
However, these approaches are computationally intensive and constrained by low temporal resolution (<30 Hz), high latency (>30 ms), motion blur, and limited dynamic range (60 dB). These limitations decouple perception from the continuous dynamics of the real world, rendering them unsuitable for high-agility mobile agents.
Embodied perception demands a sensor that is natively responsive to motion.
\textbf{New sensor: Event camera.}
The event cameras are novel bio-inspired sensors that outputs pixel-wise intensity changes in an asynchronous manner, offer a paradigm shift \cite{gehrig2024low, gallego2020event}. 
Unlike frame cameras, it generates output based on scene dynamics rather than a global clock that is independent of the scene, just as biological organisms process visual information asynchronously to survive in dynamic worlds \cite{rebecq2019high}. 
The event cameras offer four key advantages that align well with requirements of mobile perception tasks for high-agility mobile agents: 
% \textit{High temporal resolution.}
\why{
$(i)$ The \textit{$\mu$s-level temporal resolution} refers to the time interval between two consecutive samples. A higher temporal resolution implies a smaller interval, enabling event cameras to capture high-speed motions without motion blur and thereby supporting accurate perception during fast operations \cite{falanga2020dynamic}.
$(ii)$ The \textit{$\mu$s-level perception latency} denotes the time required for the sensor to respond to a change in illumination by producing an output. A lower latency allows environmental changes to be reported to mobile agents almost instantaneously \cite{he2024microsaccade}.
}
% $(i)$ The \textit{$\mu s$-level temporal resolution} enables the capture of high-speed motions without motion blur, supporting accurate perception during high-speed operations \cite{falanga2020dynamic}. 
% $(ii)$ The \textit{$\mu s$-level perception latency} allows the report of environmental changes to mobile agents almost instantaneously \cite{he2024microsaccade}.
$(iii)$ The \textit{high dynamic range (HDR)}, which is 140 $dB$ compared to 60 $dB$ for standard cameras, making it effective in diverse lighting conditions \cite{zou2024eventhdr, rebecq2019high}. 
% \textit{Low power consumption.}
$(iv)$ The \textit{low power consumption} (\eg, 0.5 $W$) makes it particularly suitable for efficient designed mobile agents \cite{scheerlinck2020fast}. 
These advantages position event cameras as a promising technology to empower mobile agents designed for high-speed operation and efficiency.
Event cameras offer high-accuracy, low-latency data acquisition but face three main processing challenges:
\textit{(i)} Event cameras' sensitivity to illumination causing significant noise,
% \textit{(ii)} lack of inherent semantic information hindering feature extraction,
% \textit{(ii)} lack of stable and persistent texture and they cannot maintain high informational output all the time, as the readings occur at image edges but depend on both the motion and the scene texture, where No events are recorded at edges parallel to the camera motion, making feature extraction and accurate and long-term data association very difficult,
\why{
\textit{(ii)} Event data lacks stable, persistent texture and cannot provide consistent information, as events are generated only at image edges depending on motion and scene texture, complicating feature extraction and long-term data association.
}
\textit{(iii)} Large data volume leading to high computational load on mobile agents.
Efficient and accurate event data processing is thus crucial for resource-constrained, agile mobile agents to perceive their state and environment.
This survey reviews event processing algorithms across six stages: event representation, denoising, filtering and feature extraction, matching, mapping, and hardware/software acceleration.

\begin{table*}[t]
    \centering
    \caption{Summary of topics covered in various studies}
    \vspace{-0.3cm}
    \resizebox{\textwidth}{!}{
    \normalsize
    \renewcommand{\arraystretch}{2}
    % \begin{tabular}{>
    % {\centering\arraybackslash}m{0.06\textwidth}>
    % {\centering\arraybackslash}m{0.078\textwidth}>
    % {\centering\arraybackslash}m{0.088\textwidth}>
    % {\centering\arraybackslash}m{0.078\textwidth}>
    % {\centering\arraybackslash}m{0.078\textwidth}>
    % {\centering\arraybackslash}m{0.078\textwidth}>
    % {\centering\arraybackslash}m{0.078\textwidth}>
    % {\centering\arraybackslash}m{0.078\textwidth}>
    % {\centering\arraybackslash}m{0.078\textwidth}>
    % {\centering\arraybackslash}m{0.078\textwidth}>
    % {\centering\arraybackslash}m{0.078\textwidth}>
    % {\centering\arraybackslash}m{0.078\textwidth}>
    % {\centering\arraybackslash}m{0.078\textwidth}>
    % {\centering\arraybackslash}m{0.078\textwidth}>
    % {\centering\arraybackslash}m{0.078\textwidth}>
    % {\centering\arraybackslash}m{0.078\textwidth}}
    % \begin{tabular}{*{16}{>{\centering\arraybackslash}X}}
    \begin{tabular}{*{16}{>{\centering\arraybackslash}m{0.103\textwidth}}}
    \hline
    % \toprule
    \cellcolor{myblue}\textcolor{white}{\textbf{Topic}} & 
    \cellcolor{myblue}\textcolor{white}{\textbf{Design of event camera}} & 
    \cellcolor{myblue}\textcolor{white}{\textbf{Adv.   of event camera}} & 
    \cellcolor{myblue}\textcolor{white}{\textbf{Gen. model of event}} & 
    \cellcolor{myblue}\textcolor{white}{\textbf{HW design}} & 
    \cellcolor{myblue}\textcolor{white}{\textbf{Products}} & 
    \cellcolor{myblue}\textcolor{white}{\textbf{Datasets}} & 
    \cellcolor{myblue}\textcolor{white}{\textbf{Repr.}} & 
    \cellcolor{myblue}\textcolor{white}{\textbf{Denoising}} & 
    \cellcolor{myblue}\textcolor{white}{\textbf{Filtering and feature ext.}} & 
    \cellcolor{myblue}\textcolor{white}{\textbf{Matching}} & 
    \cellcolor{myblue}\textcolor{white}{\textbf{Mapping}} & 
    \cellcolor{myblue}\textcolor{white}{\textbf{Accel.}} & 
    \cellcolor{myblue}\textcolor{white}{\textbf{App.}} & 
    \cellcolor{myblue}\textcolor{white}{\textbf{Adv. in mobile comp.}} & 
    \cellcolor{myblue}\textcolor{white}{\textbf{Challenge in mobile comp.}} \\ 
    % \hline

    \cellcolor{gray!10} {\large \textbf{\renewcommand{\arraystretch}{1.3}\begin{tabular}{@{}c@{}}
        \cite{gallego2020event} \\
        (2020)
        \end{tabular}}}  &  \cellcolor{gray!10} \checkmark & \cellcolor{gray!10} \checkmark & \cellcolor{gray!10} \checkmark & \cellcolor{gray!10} \checkmark & \cellcolor{gray!10} \checkmark & \cellcolor{gray!10} & \cellcolor{gray!10} \checkmark & \cellcolor{gray!10} & \cellcolor{gray!10} \checkmark & \cellcolor{gray!10} \checkmark & \cellcolor{gray!10} \checkmark & \cellcolor{gray!10} & \cellcolor{gray!10} & \cellcolor{gray!10} & \cellcolor{gray!10} \\

    {\large\textbf{\renewcommand{\arraystretch}{1.3}\begin{tabular}{@{}c@{}}
        \cite{zheng2023deep} \\
        (2023)
        \end{tabular}}} &  & \checkmark &  &  &  &  & \checkmark &  & \checkmark & \checkmark & \checkmark &  & \checkmark &  &  \\

    \cellcolor{gray!10} {\large\textbf{\renewcommand{\arraystretch}{1.3}\begin{tabular}{@{}c@{}}
        \cite{chakravarthi2024recent} \\
        (2024)
        \end{tabular}}}  & \cellcolor{gray!10} & \cellcolor{gray!10} \checkmark & \cellcolor{gray!10}\checkmark & \cellcolor{gray!10} & \cellcolor{gray!10}\checkmark & \cellcolor{gray!10}\checkmark & \cellcolor{gray!10} & \cellcolor{gray!10} & \cellcolor{gray!10} & \cellcolor{gray!10} & \cellcolor{gray!10} & \cellcolor{gray!10} & \cellcolor{gray!10}\checkmark & \cellcolor{gray!10} & \cellcolor{gray!10} \\

    {\large\textbf{\renewcommand{\arraystretch}{1.3}\begin{tabular}{@{}c@{}}
        \cite{novo2024neuromorphic} \\
        (2024)
        \end{tabular}}} &  & \checkmark &  &  &  & \checkmark & \checkmark &  & \checkmark & \checkmark & \checkmark &  &  &  &  \\

    \cellcolor{gray!10} {\large\textbf{\renewcommand{\arraystretch}{1.3}\begin{tabular}{@{}c@{}}
        \cite{shariff2024event} \\
        (2024)
        \end{tabular}}} & \cellcolor{gray!10} & \cellcolor{gray!10}\checkmark & \cellcolor{gray!10}\checkmark & \cellcolor{gray!10}\checkmark & \cellcolor{gray!10}\checkmark & \cellcolor{gray!10} & \cellcolor{gray!10}\checkmark & \cellcolor{gray!10}\checkmark & \cellcolor{gray!10} & \cellcolor{gray!10} & \cellcolor{gray!10} & \cellcolor{gray!10} & \cellcolor{gray!10}\checkmark & \cellcolor{gray!10} & \cellcolor{gray!10} \\

    {\large\textbf{\renewcommand{\arraystretch}{1.3}\begin{tabular}{@{}c@{}}
        This \\
        survey
        \end{tabular}}} & \checkmark & \checkmark & \checkmark & \checkmark & \checkmark & \checkmark & \checkmark & \checkmark & \checkmark & \checkmark & \checkmark & \checkmark & \checkmark & \checkmark & \checkmark \\
   
\bottomrule
\end{tabular}
}
\vspace{0.02em}
 \begin{flushleft} 
    % \scriptsize 
    \textit{\textbf{Note}: 
    % The abbreviations employed in this comparison are as follows: Avg. for Advantage, Gen. model for Generation model, HW design for Hardware design, Repr. for Representations, feature ext. for feature extraction, Accel. for Acceleration, App. for Application, mobile comp. for mobile computing.
    Abbreviations used: Avg. (Advantage), Gen. model (Generation model), HW (Hardware), Repr. (Representations), feature ext. (Feature extraction), Accel. (Acceleration), App. (Application), mobile comp. (Mobile computing).
    }
\end{flushleft}
\vspace{-0.6cm}
\label{summary}
\end{table*}

\why{As shown in Tab. \ref{summary}, this survey extends previous surveys by focusing on \textit{how event cameras enable high-agility, resource-constrained mobile agents to achieve high-accuracy and low-latency self-state estimation and environmental understanding} and extending prior surveys by incorporating literature from 2023–2025. 
It outlines the processing workflow of event data and reviews the advancements at each stage of this workflow. 
Using the key metrics of accuracy and efficiency in mobile computing, we summarize various methods in each stage to provide a deeper analysis of cutting-edge research, and assess each stage under latency, accuracy, and power constraints.
}
It also covers work on event-based hardware and software acceleration, offering insights for deploying event cameras on resource-constrained mobile agents. 
Finally, we discuss applications of event cameras on mobile agents.

% \textbf{Contribution.} 
The main contribution of the survey paper is summarized as follows.

% \begin{itemize}
%     \item We present a comprehensive review of how event cameras enhance high-agility resource-constrained mobile agents by enabling high-accuracy, low-latency self-state estimation and environmental perception.
%     \item We provide a comprehensive introduction to event generation models, event camera hardware design, commercial products, as well as event-based datasets. 
%     \item We highlight unique advantages of event cameras in mobile perception, as well as the specific challenges they encounter. 
%     \item We categorize event stream processing methods into several stages and provide a comprehensive review of each, including event stream representation, data processing algorithms, acceleration techniques, and applications on event-based mobile platforms.
%     \item We present our insights and solutions for future trends, with a particular focus on bio-inspired event camera hardware design, algorithm development, and hardware-software co-optimization techniques.
% \end{itemize}

% \noindent $(1)$ We extend previous surveys and present a comprehensive review of how event cameras enhance high-agility resource-constrained mobile agents by enabling high-accuracy, low-latency self-state estimation and environmental perception. 
\why{
\noindent $(1)$ We extend prior surveys by presenting a comprehensive review of how event cameras empower high-agility, resource-constrained mobile agents to achieve accurate, low-latency self-state estimation and environmental perception.
% To better accommodate the trend of mobile device design toward high-mobility design, event camera-based perception technology has emerged, enabling ms-level temporal resolution and latency, as well as a low-power consumption perception mode. 

% We cover event-based abstraction, algorithms, acceleration, and applications.

% \noindent $(2)$ We provide a comprehensive introduction to event generation models, event camera hardware design, commercial event camera products, as well as event-based datasets.
\noindent $(2)$ We present an in-depth introduction to event generation models, the design of event camera hardware, existing commercial products, and benchmark datasets for event-based vision.}
% The introduction of these aspects facilitates device selection and enables the use of public datasets for convenient implementation and reproduction of event camera mobile perception.

\noindent $(3)$ We highlight unique advantages and specific challenges of event cameras in mobile embodied perception.
% \noindent $(3)$ We highlight unique advantages of event cameras in mobile perception, as well as specific challenges they encounter.
% This will provide readers with a deeper understanding of the benefits of applying event cameras in mobile perception and the critical issues that need to be addressed.

% \noindent \textit{(4)} We categorize event stream processing methods into several stages and provide a comprehensive review of each, including event stream representation, data processing algorithms, acceleration, and event-based mobile platform applications. 
% \noindent $(4)$ We categorize event stream processing methods into several stages and provide a comprehensive review of each, including event stream representation, data processing algorithms, acceleration techniques, and applications on event-based mobile platforms. We also assess each stage under latency, accuracy, and power constraints.

\noindent $(4)$ We categorize event stream processing methods into distinct stages and present a comprehensive review of each, covering event stream representation, data processing algorithms, acceleration strategies, and applications on event-based mobile agents.
Furthermore, we evaluate these stages with respect to latency, accuracy, and power constraints.

% (\eg, denoising, filtering and feature extraction, matching and mapping)
% Furthermore, we conduct a comprehensive survey on related studies at different stages using accuracy and efficiency-two key metrics in mobile perception-as evaluation criteria.
\noindent $(5)$ We present our insights and solutions for future trends, with a particular focus on bio-inspired event camera hardware design, algorithm development, and hardware-software co-optimization techniques.
% We review recent research efforts in addressing practical challenges from both hardware and software perspectives and present our insights and solutions for future trends, with a particular focus on bio-inspired event camera hardware design, algorithm development, and hardware-software co-optimization techniques.

\begin{figure}[t]
    \centering
        \includegraphics[width=0.8\columnwidth]{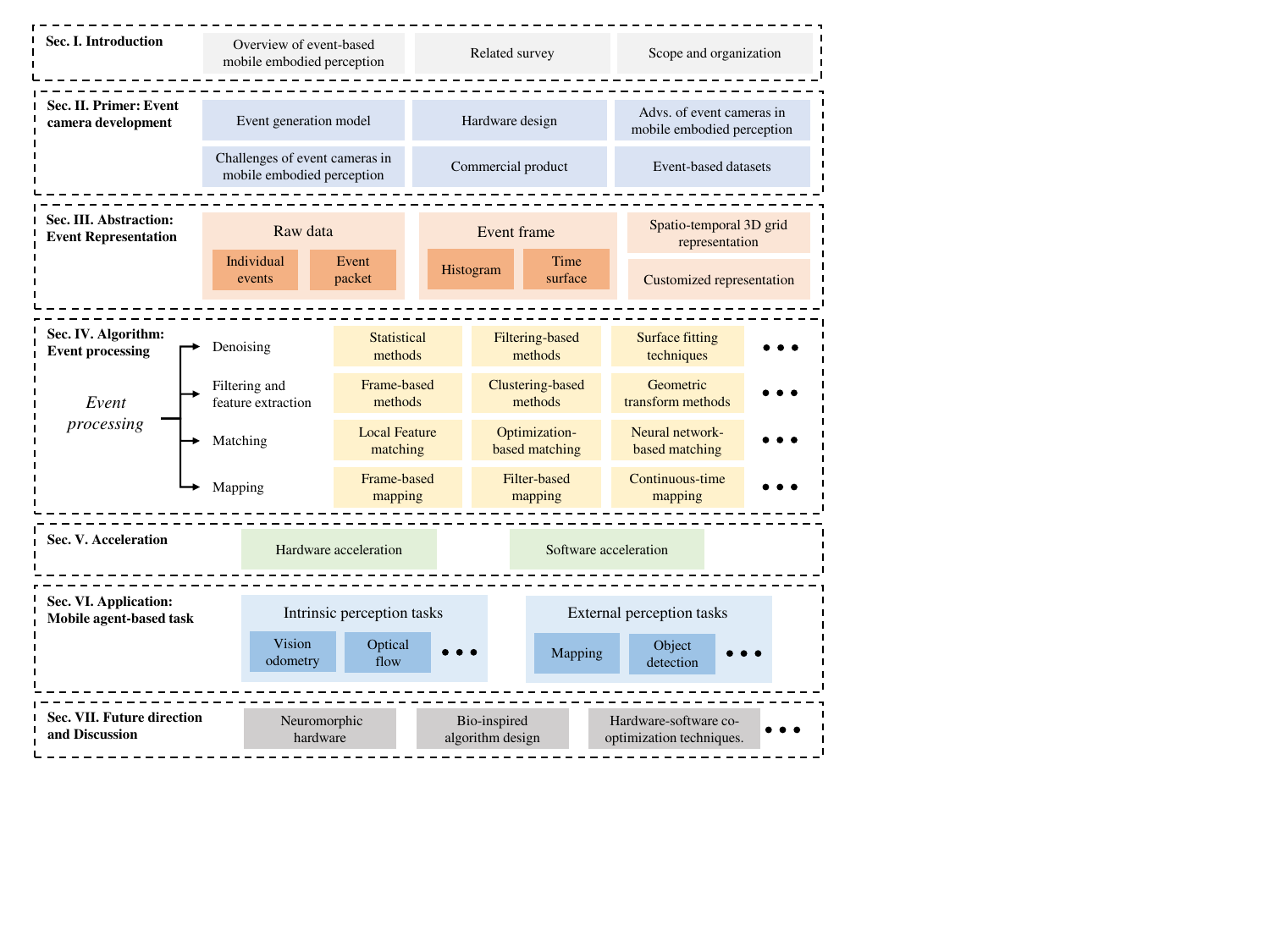}
    \vspace{-0.25cm}
     \caption{Structure of this survey. }
    \label{outline}
    \vspace{-0.7cm}
\end{figure}

% \textbf{Organization.}
% \fig \ref{outline} illustrates the structure of this survey. 
% In Sec. \ref{2}, we introduce the primer of event cameras, including the principles behind event generation, the hardware design of event cameras, existing products, and datasets. 
% We also emphasize the advantages and challenges of applying event cameras to mobile agents. 
% Sec. \ref{3} discusses methods for representing event data, while Sec. \ref{4} reviews existing work in event data processing, including event-based denoising, filtering, matching, and mapping. This survey highlights the performance of current methods in terms of accuracy and efficiency. 
% In Sec. \ref{5}, we present hardware and software acceleration techniques that enable event cameras to operate effectively on resource-constrained mobile agents.
% Sec. \ref{6} covers the application scenarios of event cameras in mobile agents. 
% In Sec. \ref{8}, we propose potential future research directions, followed by a conclusion in Sec. \ref{9}.

% \textbf{Online resource.} 
This survey presents a comprehensive review of event-based embodied perception systems, focusing on key technological advancements and practical applications.
To further support the research community, we have established an open-source \textit{Online Sheet} \footnotemark[1], which is adapted from \cite{Recent}. 
This online sheet will be regularly updated, ensuring access to latest developments and fostering continued innovation in event-based embodied perception systems.

\footnotetext[1]{\href{https://docs.qq.com/sheet/DRFRaUGNnQVNyb2d1?tab=BB08J2}{\color{blue}{Event-based mobile embodied perception resource}}}

% \vspace{0.5em} 
% \textbf{Organization.}
\fig \ref{outline} illustrates the survey structure.
Sec. \ref{2} introduces event camera fundamentals, including principles, hardware, products, datasets, and their pros and cons on mobile agents.
Sec. \ref{3} covers event stream \textbf{Abstraction} methods, while Sec. \ref{4} reviews event processing \textbf{Algorithms} such as denoising, filtering, matching, and mapping.
Sec. \ref{5} discusses hardware/software \textbf{Acceleration} for resource-constrained agents, and Sec. \ref{6} highlights mobile \textbf{Applications} of event cameras.
Sec. \ref{8} outlines future directions, with conclusions in Sec. \ref{9}.

\vspace{-0.2cm}
\section{Primer: Event camera development} \label{2}

% \begin{figure*}[t]
%     \centering
%     \includegraphics[width=0.5\linewidth]{Figs/chapter2_1.pdf}
%     \vspace{-0.4cm}
%     \caption{
%     % \notice{Need Redraw! Ruishan, Please! okk} refer to Fig 2 of Recent Event Camera Innovations: A Survey
%     Principle of the Event Cameras: Events are generated based on changes in logarithmic light intensity over time.
%     }
%     \label{software}
%     \vspace{-0.85cm}
% \end{figure*}  

\subsection{Event generation model}

% Unlike traditional cameras that capture images at regular intervals, event cameras operate based on changes in log brightness at individual pixels to generate events. This unique feature allows event cameras to detect changes at each pixel with extremely high speed, effectively mitigating motion blur often encountered in traditional cameras when capturing fast-moving objects, as they do not require all pixels to be exposed simultaneously.

% An event includes the pixel location where the change occurred, the time elapsed since the last event at that pixel, and the polarity change in log brightness. The event generation model can be described by the following formula:
% \begin{center}
%   $e_{k}=(x_{k},t_{k},p_{k})$
% \end{center} 
% Where the $e_{k}$ represents an event, $x_{k}$ denotes the pixel location, $t_{k}$ indicates the elapsed time, and $p_{k}$ refers to the polarity of the change.
    
% The change in log brightness can be expressed as follows:
% \begin{center}
%      $\Delta L(x_{k},t_{k})=L(x_{k},t_{k})-L(x_{k},t_{k}-\Delta t_{k}) $
% \end{center} 
% Where \( L \) represents the change in log brightness.

% An event is generated only when the absolute value of $\Delta L(x_{k},t_{k})$ reaches the threshold \( C \). Compared to traditional cameras, this event-driven image generation method significantly reduces redundant information, thereby accelerating processing speed, conserving computational resources, and enabling the integration of such cameras into embedded systems.If the value of $\Delta L(x_{k},t_{k})$ is negative, $p_{k}$ is -1; otherwise, $p_{k}$ is 1.

\begin{wrapfigure}{r}{0.5\textwidth} % r 表示图在右边，l 表示图在左边
  \centering
  \includegraphics[width=0.5\textwidth]{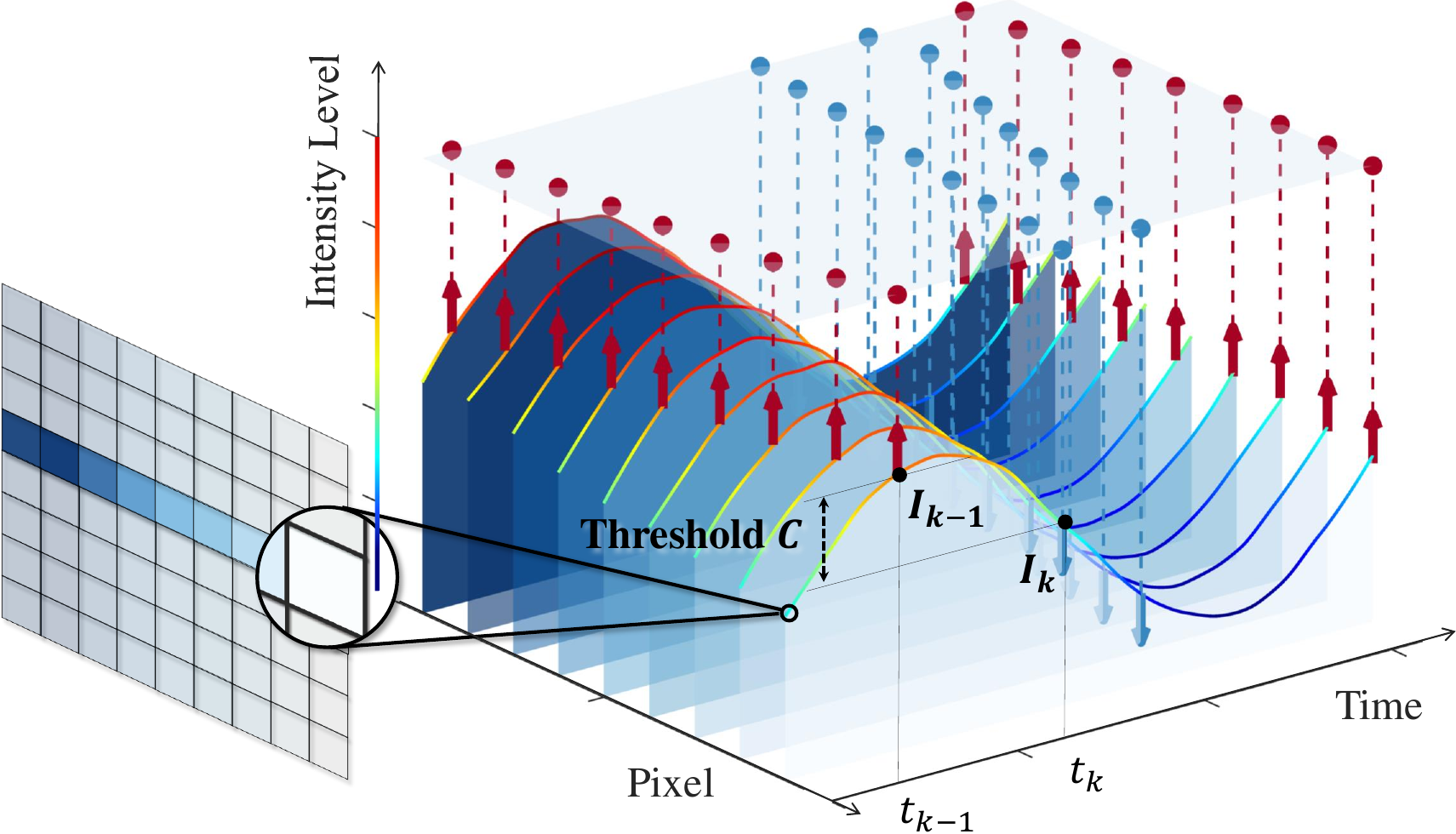}
  \caption{Principle of the Event Cameras: Events are generated based on changes in logarithmic light intensity over time.}
  \vspace{-0.4cm}
  \label{software}
\end{wrapfigure}

Unlike conventional frame cameras that capture images at fixed intervals, event cameras operate asynchronously by detecting changes in log intensity at individual pixels, generating events only when changes occur.
This characteristic enables event cameras to achieve exceptionally high temporal resolution and effectively mitigate motion blur, particularly in scenarios involving fast-moving objects.

% as they do not require simultaneous exposure of all pixels.

Each event is defined by the pixel location where the change occurs, the timestamp, and the polarity. 
Formally, an event can be represented as $e_{k} = (x_{k}, t_{k}, p_{k})$, 
% \[
% e_{k} = (x_{k}, t_{k}, p_{k}),
% \]  
where \(e_{k}\) denotes the event, \(x_{k}\) specifies the pixel location, \(t_{k}\) represents the timestamp, and \(p_{k}\) indicates the polarity of the change.
The change in log intensity is given by $\Delta L(x_{k}, t_{k}) = L(x_{k}, t_{k}) - L(x_{k}, t_{k} - \Delta t_{k})$, 
% \[
% \Delta L(x_{k}, t_{k}) = L(x_{k}, t_{k}) - L(x_{k}, t_{k} - \Delta t_{k}),
% \]  
where \(L(x_{k}, t_{k})\) represents the log intensity at pixel \(x_{k}\) and time \(t_{k}\). As shown in \fig \ref{software}, an event is triggered only when \(|\Delta L(x_{k}, t_{k})|\) exceeds a predefined threshold \(C\). The polarity \(p_{k}\) is assigned as \(+1\) if \(\Delta L(x_{k}, t_{k}) > 0\), and \(-1\) otherwise.
This event-driven paradigm substantially reduces redundant data, enhancing processing efficiency, conserving computational resources, and enabling deployment in resource-constrained systems such as embedded mobile agents \cite{xu2023taming}.
In practice, threshold $C$ can be adjusted to meet specific application needs and impacts event camera performance.
A high $C$ reduces sensitivity, may missing subtle changes, while a low $C$ increases noise-triggered events, causing redundancy.

\vspace{-0.2cm}
\subsection{Hardware design}

In this part, we will introduce hardware design of modern event cameras, as illustrated in Tab. \ref{cameratype}.

\begin{figure}
    \centering
    \includegraphics[width=0.9\linewidth]{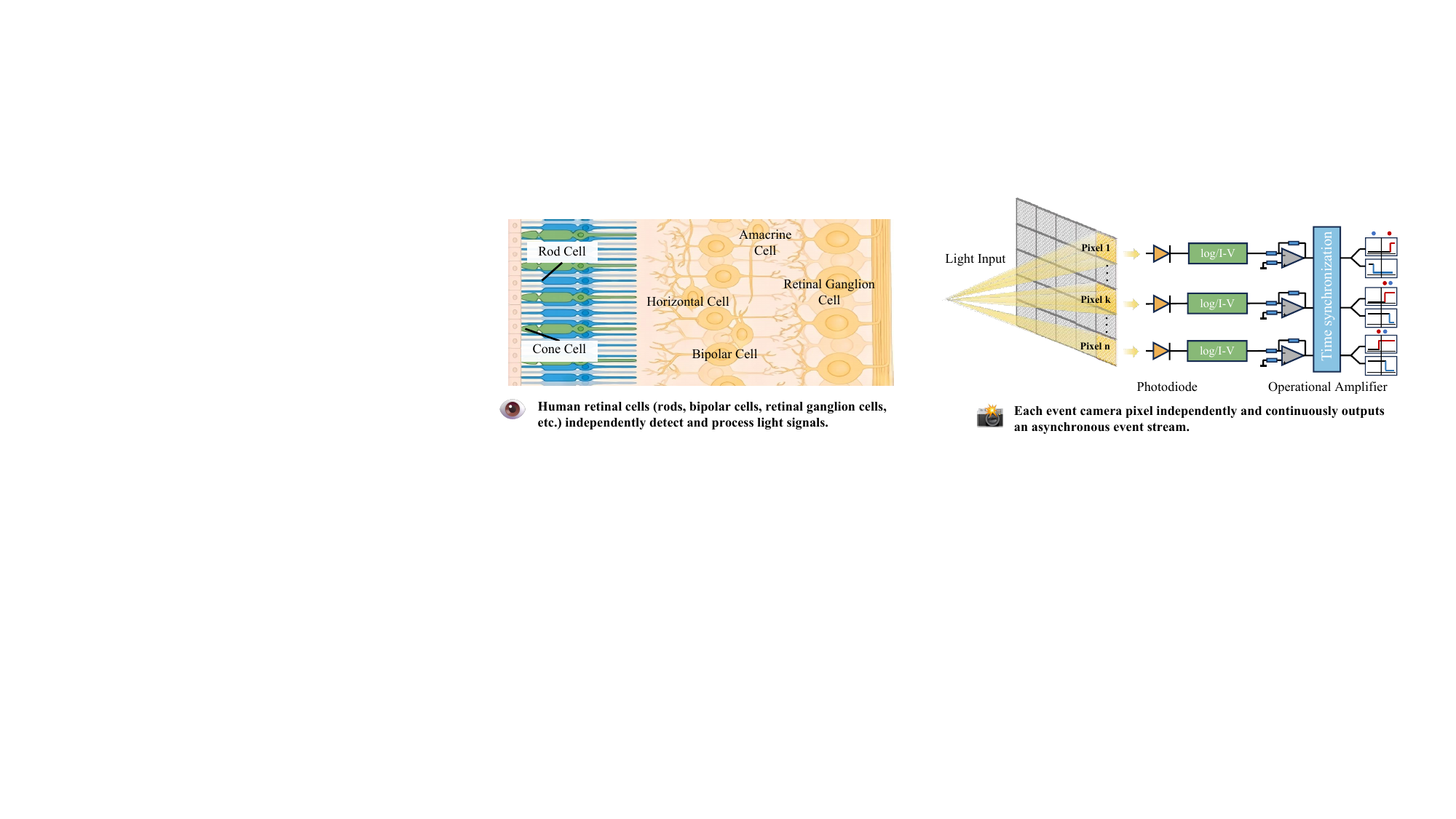}
    \vspace{-0.2cm}
    \caption{
    % \notice{Need Redraw! Ruishan, Please!okk} refer to Fig 2 of Recent Event Camera Innovations: A Survey
    Working Mechanism of Event Cameras: Inspired by the rod cells in the human eye, event camera operates at the pixel level, independently transforming light into voltage signals to capture intensity variations.
    }
    \label{hardware}
    \vspace{-0.7cm}
\end{figure}

% why：这我改了，pengtao，你该其他的就好的～   haodi
\why{
\textbf{General hardware architecture}.
As shown in \fig \ref{hardware}, event cameras employ CMOS sensors for low-latency operation, which involves three steps: $(i)$ incident light generates electron–hole pairs, $(ii)$ electrons are collected under an electric field, and $(iii)$ readout circuits convert them into voltage signals for logarithmic brightness computation. The event circuit filters noise, applies a threshold, and triggers an event when exceeded, transmitting it to the processor.
}

\textbf{DVS event camera} \cite{lichtsteiner2008128}. 
The Dynamic Vision Sensor (DVS), inspired by the silicon retina, detects brightness changes via capacitance coupling and resets after each measurement. 
It outputs only changes, enabling smaller pixel sizes but restricting output to event data, which limits information extraction in static scenes.

\textbf{ATIS event camera} \cite{posch2010qvga}. 
The Active Time-Image Sensor (ATIS) uses subpixels to measure absolute brightness, doubling pixel area compared to DVS but enabling wide dynamic and static ranges for robust imaging under extreme lighting. Its limitation lies in potential misalignment between absolute brightness (averaged across pixels) and event data (triggered per pixel), especially during high-speed motion.

\textbf{DAVIS event camera} \cite{brandli2014240}.
The Dynamic and Active Vision Sensor (DAVIS) is capable of outputting both absolute brightness and event-based data. In DAVIS, pixels and subpixels share the same sensor, enabling a more compact design. 
As a result, the pixel area is smaller than that of the ATIS, with only a modest 5\% increase in size compared to the DVS. 
However, the sampling speed of the DAVIS circuit is slower than that of the DVS circuit. 
Beyond foundational architectures, modern event cameras are optimized for mobile agents. 
Key trends include more versatile readouts (e.g., hardware-generated event-accumulation views in the CeleX-V \cite{chen2019live}), mobile-friendly interfaces like MIPI CSI-2 (on the CeleX5-MIPI \cite{celex5mipi_datasheet}), and ultra-low power optimization specifically targeting battery-powered mobile agents (\eg, Prophesee's GENX320 \cite{prophesee_genx320_datasheet}). 
A detailed comparison of commercial products is presented in Sec. \ref{comparison}.

\begin{table*}[t]
    \centering
    \caption{Classification of event camera types}
    \vspace{-0.3cm}
    \footnotesize
    \begin{tabular}{@{}lcccc@{}}
    % \begin{tabularx}{\textwidth}{@{}>{\raggedright\arraybackslash}p{2.5cm}XXXX@{}}
        \toprule
        \textbf{Indicator} & \textbf{Original event camera\cite{mahowald1994silicon}} & \textbf{DVS} \cite{lichtsteiner2008128} & \textbf{ATIS} \cite{posch2010qvga} & \textbf{DAVIS}  \cite{brandli2014240}\\ \midrule
        Development origin & Mahowald \& Mead, Caltech & Inspired by silicon retina & Evolved from DVS & Advanced DVS/ATIS \\
        Output data type & Log brightness & Change in brightness & Change + absolute brightness & Change + absolute brightness \\
        Sensor type & Large pixels, CMOS & CMOS & CMOS & CMOS \\
        Pixel structure & Single-pixel design & Smaller, simpler pixels & Dual-subpixel & Shared pixel/subpixel \\
        Brightness measurement & Continuous-time & Change only & Absolute + change & Absolute + change \\
        Dynamic range & Limited & Narrow & High & Moderate \\
        Event synchronization & Basic & Fast reset & Potential mismatch & Slow sampling \\
        Noise filtering & Minimal & Simple & Complex & Advanced \\ \bottomrule
    \end{tabular}
    % \end{tabularx}
    \vspace{-0.6cm}
    \label{cameratype}
\end{table*}

\vspace{-0.2cm}
\subsection{Advantages of event cameras in mobile embodied perception}
% \notice{Emphasis the potential in high speed mobile platform(application)}
% Compared to frame cameras, event cameras can greatly enhance mobile agents by improving embodied perception capabilities, boosting efficiency, and broadening application scope. Specifically, their advantages include:
% Compared to frame cameras, event cameras offer numerous potential advantages for mobile agents. 
% They have the potential to significantly enhance the embodied perception capabilities of mobile agents, improve operational efficiency, and expand their range of applications.
% Specifically, the advantages that event cameras provide to mobile agents include:

% $(i)$ The \textit{$\mu$s-level temporal resolution} refers to the time interval between two consecutive samples. A higher temporal resolution implies a smaller interval, enabling event cameras to capture high-speed motions without motion blur and thereby supporting accurate perception during fast operations \cite{falanga2020dynamic}.

\why{
\textbf{High temporal resolution.}
% The high agility of mobile agents leads to rapid environmental changes, causing motion blur in standard frame cameras and making timely embodied perception about changes in the environment and self-state difficult for radar- or camera-based solutions due to their low spatial-temporal resolution. 
% Temporal resolution refers to the time interval between two consecutive samples, where a higher temporal resolution implies a smaller interval.
% Event cameras, with microsecond resolution and the ability to detect changes without motion blur, offer crucial support for quickly embodied perception changes in the environment and device state.
The high agility of mobile agents induces rapid environmental changes, often causing motion blur in frame cameras and limiting the responsiveness of radar- or frame camera-based solutions due to their low spatio-temporal resolution.
Temporal resolution, defined as the interval between consecutive samples, improves with shorter intervals.
Event cameras, with microsecond-level resolution and motion-blur-free embodied perception, enable timely detection of both environmental dynamics and agent state.

% $(ii)$ The \textit{$\mu$s-level embodied perception latency} denotes the time required for the sensor to respond to a change in illumination by producing an output. A lower latency allows environmental changes to be reported to mobile platforms almost instantaneously \cite{he2024microsaccade}.
\textbf{Low perception latency.}
The high agility of mobile agents demands rapid awareness of both environmental and self-state changes. 
Perception latency, defined as the time required for a sensor to respond to environmental variations, is relatively long in frame cameras due to their global exposure time (20 ms), which delays reactions and increases collision risk. In contrast, event cameras employ independent pixels that trigger events immediately upon brightness changes, achieving sub-millisecond latency and enabling mobile agents to detect and respond to changes instantaneously.
}
% The high agility of mobile agents requires rapid awareness of environmental and self-state changes to respond effectively.
% embodied perception latency denotes the time required for the sensor to respond to a change in the environment by producing an output.
% Standard frame cameras have long embodied perception latency due to a global exposure time (about 20 $ms$), which delays reaction times and increases the risk of collisions. 
% In contrast, event cameras operate with independent pixels that generate events immediately upon brightness changes, without requiring a global exposure time.
% This results in sub-millisecond latency, allowing mobile agents to detect changes and respond quickly.

% The high agility of mobile agents requires rapid awareness of environmental and self-state changes to respond effectively.
% Standard frame cameras depend on a global exposure time (about 20 $ms$) and additional processing time (10-20 $ms$), which delays reaction times and increases the risk of collisions. 
% In contrast, event cameras operate with independent pixels that generate events immediately upon brightness changes, without requiring a global exposure time.
% This results in sub-millisecond latency, allowing mobile agents to detect changes and respond quickly.

\textbf{High dynamic range (HDR).}
Mobile agents are increasingly used in challenging environments, such as low-light nighttime and bright daytime settings, requiring reliable embodied perception across varying lighting conditions. 
Frame cameras have a dynamic range of about 60 $dB$, making them less effective in extreme lighting. 
Event cameras operate on a logarithmic scale with independent pixels, offering a high dynamic range with \textgreater 120 $dB$. 
This enables them to adapt to both extremely dark and bright conditions, making mobile agents suitable for a wider range of scenarios. 

\textbf{Low power consumption.}
Mobile agents in complex urban environments often prioritize efficiency but face limited computation and power. 
Frame cameras require heavy processing and energy, whereas event cameras transmit only brightness changes, reducing redundant data and easing computational and power demands to improve efficiency.

% Mobile agents, designed for operation in complex urban environments, often prioritize efficiency, which can limit their computational capabilities and power reserves. 
% Standard frame cameras typically demand substantial computational resources to process image data, leading to increased processing times and additional power consumption for mobile agents. 
% In contrast, event cameras transmit only changes in brightness, thereby eliminating redundant data.
% This elimination in data redundancy alleviates the computational and power burdens on mobile agents, enhancing their operational efficiency.

% In summary, event cameras offer numerous advantages for mobile agents, expanding their application scenarios and enhancing efficiency. 
% At the same time, mobile agents can leverage benefits of event cameras to unlock full potential.

\vspace{-0.2cm}
\subsection{Challenges of event cameras in mobile embodied perception}
% \vspace{-0.1cm}
% Since event cameras operate fundamentally differently from frame cameras which measure per-pixel brightness changes as events asynchronously, integrating event cameras into mobile agents presents several challenges:
Since event cameras operate fundamentally differently from frame-based cameras by capturing per-pixel brightness changes asynchronously as events, their integration into mobile agents poses several challenges:
% is non-trivial:
% and presents significant challenges:\\
% rather than capturing absolute brightness at a constant rate

\why{
\textit{(i) How to mitigate event bursts and accurately extract features from event data}, given the high-speed operation of mobile agents and event cameras' lack of stable, persistent semantic information?
Event cameras are highly sensitive to illumination changes, with even minor variations triggering numerous events. 
On high-agility mobile agents, rapid scene changes captured by onboard event cameras can trigger event bursts, generating thousands of events in a short time.
Taking the DAVIS 346 event camera (346 × 260 resolution) as an example, let $v$ denote the translational velocity of the drone and $w$ its angular velocity.
Under normal flight conditions ($2 \text{m/s} \leq v \leq 4 \text{m/s}$, $5^\circ/\text{s} \leq w \leq 15^\circ/\text{s}$), the event generation rate is 298 $e/ms$. During rapid translation ($18\text{m/s} \leq v \leq 25\text{m/s}$, $5^\circ/\text{s} \leq w \leq 15^\circ/\text{s}$), the rate increases to 945 $e/ms$, while in rapid rotation ($2\text{m/s} \leq v \leq 4\text{m/s}$, $75^\circ/\text{s} \leq w \leq 95^\circ/\text{s}$), it further rises to 1437 $e/ms$ \cite{xu2023taming}.
In contrast, the data rate of a frame-based camera is independent of motion dynamics and scene content. Regardless of whether the scene is static or highly dynamic, frames are sampled at a fixed rate, \eg, 1MP × 30fps = 30M pixels/s, resulting in a stable and uniform data stream.
This comparison highlights the non-uniform nature of event data: static scenes generate almost no events, while high-speed motion can trigger explosive outputs far exceeding frame rates. 
This sparsity–burst duality, coupled with the lack of stable texture or semantic cues, makes efficient processing difficult, as meaningful signals are often buried in motion-induced noise.
% This comparison highlights the non-uniform data characteristics of event cameras: in static scenes, almost no events are triggered, leading to much lower data volume compared to frame cameras; yet under high-speed motion, event outputs can increase explosively, with instantaneous event rates far exceeding those of frame cameras. 
% Such sparsity–burst duality, driven by high temporal resolution, poses significant challenges for efficient event data processing.
% Moreover, the absence of stable and persistent texture or semantic information further complicates the extraction of critical environmental cues, as meaningful signals are often overshadowed by large numbers of motion-induced irrelevant events.

\textit{(ii) How to efficiently process a large volume of event data} given on-board constrained resources?
Mobile agents typically rely on low-power embedded systems for efficiency, inherently limiting their computational capacity.
For example, a smartphone SoC such as the Qualcomm Snapdragon 8Gen3 has a typical power budget of 5–7$W$, with its integrated NPU delivering around 30$TOPS$ at INT8 precision.
By contrast, a laptop equipped with an Intel Core i7-13700H can consume up to 45$W$ per CPU package and provide 1–2 $TFLOPS$ of FP32 compute power. 
For high-performance drones, the NVIDIA Jetson Orin NX module consumes 15–25$W$ and delivers 100 $TOPS$ of AI compute, optimized for onboard vision tasks.
In comparison, a NVIDIA GeForce RTX 4060Ti (160$W$) offers up to 22 $TFLOPS$ FP32 and 288 $TOPS$ INT8, while the RTX 4090 (450$W$) reaches 83 $TFLOPS$ FP32 and over 1.3 $POPS$ INT8. 
Clearly, mobile agents trade computational resources for energy efficiency.
Meanwhile, onboard event cameras capture rapid scene changes, generating large volumes of events that demand efficient processing. 
For instance, the iniVation DVXplorer can reach peak output rates above 1$Meps$, the Sony IMX636 sensor used in Prophesee Gen4.1 exceeds 10$Meps$, and the high-speed CelePixel Taurus supports up to 240$Meps$.
Such workloads heavily strain limited computational resources. 
In a typical scenario, an algorithm processing a 5$Meps$ event stream with 100 $GOP/s$ demand is trivial for an RTX 4060Ti (288 $TOPS$), leaving ample headroom for other tasks. 
For a smartphone NPU (30 $TOPS$), however, achieving low latency and high energy efficiency is challenging due to memory sharing constraints and thermal throttling.
}

\vspace{-0.2cm}
\subsection{Commercial product \& comparison} \label{comparison}
% \vspace{-0.1cm}
% \subsubsection{Commercial companies and products}
% Event cameras are increasingly entering the commercial market, with several leading companies spearheading the development and production of these advanced sensors. Notable manufacturers include Prophesee and Inivation, each providing specialized products tailored to diverse application scenarios. 
\mpt{
% Event cameras are increasingly entering the commercial market, with several leading companies spearheading the development and production of these advanced sensors. 
% Notable manufacturers include Prophesee, iniVation, CelePixel, Lucid Vision Labs, and Insightness, each providing specialized products tailored to diverse application scenarios, particularly for mobile and embedded systems.

Event cameras are increasingly entering the commercial market, with several companies offering distinct products. Representative suppliers include iniVation, Prophesee, Lucid Vision Labs and CelePixel. 
Their commercial offerings emphasize features relevant to mobile embodied perception contexts, such as low-power operation, compact modules, mobile-friendly interfaces, or robust industrial connectivity.

}

\begin{table*}[t]
\centering
\caption{Specifications of event camera models for mobile embodied perception applications}
\vspace{-0.3cm}
\newcolumntype{C}[1]{>{\centering\arraybackslash}m{#1}}

{\scriptsize
\begin{tabular}{ C{0.08\textwidth} C{0.12\textwidth} C{0.08\textwidth} C{0.08\textwidth} C{0.08\textwidth} C{0.10\textwidth} C{0.20\textwidth} }
\toprule
\textbf{Supplier} & \textbf{Model} & \textbf{Resolution} & \textbf{\makecell{Dynamic \\ Range (dB)}} & \textbf{\makecell{Pixel Size \\ (µm)}} & \textbf{\makecell{Power \\ Consumption}} & \textbf{Mobile/Embedded Features} \\
\midrule

\multirowcell{4}{iniVation} 
& DVXplorer & 640×480 & 90-120 & 9.0 & Max 12W & Multi-camera synchronization; Robust aluminum casing. \\
& DVXplorer Micro & 640×480 & 110 & 9.0 & <140mA @ 5VDC & Compact and lightweight design. \\
& DAVIS346 & 346×260 & 120 & 18.5 & Typical 180mA @ 5VDC & Hybrid output (provides both event and frame data).  \\
& DAVIS346 AER & 346×260 & 120 & 18.5 & Typical 180mA @ 5VDC &  Integrated IMU for self-contained visual-inertial perception. \\
\hline
\multirowcell{3}{Prophesee}
& EVK4 HD & 1280×720 & >86 & 4.86×4.86 & Typical 0.5W & High-resolution sensor with external trigger support. \\
& EVK5 HD & 1280×720 & >110 & 4.86×4.86 & Typical 0.5W & On-board advanced processing; Hardware trigger support. \\
& GENX320 & 320×320 & >120 & 6.3 & 3mW & Ultra-low power consumption; Compact, mobile-optimized design. \\
\hline
\multirowcell{2}{CelePixel} 
& CeleX-V & 1280×800 & - & 9.8 & 400mW & Multi-mode output (Event, Grayscale, Accumulated frames). \\
& CeleX5-MIPI & 1280×800 & - & 9.8 & - & Designed for direct, low-level System-on-Chip (SoC) integration. \\
\hline
\makecell{Lucid Vision \\Labs} & TRT009S-EC & 1280×720 & 120 & 4.86 & - & Industrial-grade robustness and reliable data streaming. \\
\bottomrule
\end{tabular}
}
\parbox{\textwidth}{\footnotesize} 
\label{productions}
\vspace{-0.68cm}
\end{table*}

% These companies are at the forefront of innovation in mobile embodied perception areas such as robotics, autonomous vehicles, industrial automation, and machine vision, capitalizing on the benefits of event-based embodied perception technologies to enable real-time, high-speed performance in demanding applications.

% Event cameras are gaining traction in the commercial market, with several companies at the forefront of developing and producing these specialized sensors. 
% Among the prominent manufacturers are Prophesee, Inivation, and Celepixel, each offering unique products designed for various application scenarios.
% These companies are driving innovation in fields such as robotics, autonomous vehicles, industrial automation, and machine vision, leveraging the advantages of event-based embodied perception technologies for real-time, high-speed applications.

\textbf{Inivation} is a prominent leader in the event camera industry, specializes in developing high-resolution, energy-efficient event cameras \cite{inivation}.
Inivation’s flagship sensors, like the DAVIS240 and DAVIS346, combine event-based and frame-based data, enabling real-time analysis of both modalities. With resolutions up to 1 megapixel, they rank among the highest-resolution event cameras available. Designed for excellent low-light performance and outdoor durability, these sensors are ideal for applications in robotics, autonomous systems, and more.

% Their flagship sensors, such as the DAVIS240 and DAVIS346, uniquely combine event-based and traditional frame-based data, providing a valuable platform for researchers to analyze and compare these two modalities in real-time. Inivation's cameras offer resolutions of up to 1 megapixel (1 MP), positioning them among the highest-resolution event cameras available. 
% Designed for exceptional low-light performance and durability in outdoor environments, these sensors are well-suited for diverse applications, including robotics and autonomous systems.

\textbf{Prophesee} is a leading innovator in event-based vision technology, renowned for its cameras' ultra-high dynamic range exceeding 120 dB \cite{prophesee}. 
% This makes them particularly suitable for environments with variable lighting, such as autonomous driving and industrial automation. Prophesee’s flagship Metavision sensor, along with its EVK4 HD and EVK5 HD cameras, offers microsecond-level temporal resolution, enabling the capture of fast-moving objects with minimal latency. This high precision supports accurate real-time monitoring and processing, providing clear advantages for applications requiring rapid response and high accuracy.
\mpt{
Their flagship Metavision sensors, including the EVK4 HD and EVK5 HD cameras, offer microsecond-level temporal resolution for applications requiring rapid response and high accuracy, such as autonomous driving and industrial automation. 
Recently, they have introduced the GENX320, a new sensor designed with features particularly beneficial for mobile embodied perception applications. With its compact 320×320 resolution, small 6.3µm pixels, and ultra-low power consumption, the GENX320 addresses key constraints in battery-powered mobile agents, making it well-suited for applications such as AR/VR headsets, drones, and other embedded agents \cite{prophesee_genx320_datasheet}.

}

\mpt{
% \textbf{CelePixel} offers multi-mode cameras like the CeleX-V\cite{chen2019live}, capable of outputting events, full frames, and event-accumulation views. The CeleX5-MIPI\cite{celex5mipi_datasheet}. variant provides a MIPI CSI-2 interface, crucial for direct, low-power integration with mobile SoCs 

% \textbf{Lucid Vision Lab} provides the industrial-grade TRT009S-EC, based on the Sony IMX636 sensor. It uses a GigE Vision interface with Power over Ethernet (PoE), ensuring robust data streaming for mobile robotic platforms where stability is prioritized over minimal power use \cite{lucid_trt009sec}.

\textbf{Lucid Vision Labs}, a manufacturer of industrial cameras \cite{lucid_vision_labs_cn}, offers the TRT009S-EC Triton camera built on the Sony IMX636 event sensor. It utilizes a GigE Vision interface with Power over Ethernet (PoE) to ensure robust and reliable data streaming. This design prioritizes operational stability for mobile robotic agents over the ultra-low power consumption typical of other mobile-optimized sensors.

\textbf{CelePixel} developed smart sensory platforms with a unique on-chip processing architecture. While its official website is inactive, technical resources remain available via software repositories.
Their CeleX-V sensor \cite{chen2019live} is notable for a multi-mode capability, outputting events, full frames, and hardware-generated event-accumulation views.
The CeleX5-MIPI \cite{celex5mipi_datasheet} variant targets mobile integration, providing a MIPI CSI-2 interface for low-power connection.
% to SoCs.

}

% Tab. \ref{productions} compares various event cameras across key specifications, including resolution, latency, dynamic range, pixel size, power consumption, interface type, weight, casing material, synchronization support, and additional features. Resolutions range from $346 \times 260$ to $1280 \times 720$, with latencies between 1 $\mu s$ and 800 $\mu s$, indicating varying processing speeds. Dynamic ranges span 86 $dB$ to 120 $dB$, while pixel sizes range from 4.86 $\mu m$ to 18.5 $\mu m$, impacting sensitivity and detail.
% Power consumption varies significantly, from 0.5 $W$ (typical for EVK models) to 12 $W$ for the DVXplorer. Most cameras use USB 3.0 interfaces, with some adopting Type-C or micro-B connectors. Weights range from 16 $g$ to 120 $g$, with aluminum casings being common, except for the DVXplorer Micro, which uses POM plastic. Synchronization support is standard in most models, except for the DAVIS346 and DAVIS346 AER. Additional features like integrated IMUs, neuromorphic support, and adjustable sensor configurations further enhance their application versatility.

\textbf{Comparation of products.}
\mpt{
Tab. \ref{productions} compares commercial event cameras, revealing significant diversity in their specifications. Resolutions range from 320x320 to high-definition 1280x800 pixels, while a high dynamic range (often >110 dB) is a common strength for challenging lighting. A critical differentiator is power consumption, which spans from just 3 mW for mobile-optimized models to 12 W for high-performance ones. This reflects divergent design goals, from ultra-low-power, compact agents to performance-focused industrial models with advanced features like integrated IMUs, hybrid outputs, and multi-camera synchronization for complex robotic systems.
}

\vspace{-0.2cm}
\subsection{Event-based datasets} 
% \notice{add more datasets! Refer to Recent Event Camera Innovations: A Survey}

% The event-based datasets collectively serve as important benchmarks in the field of autonomous driving, robotics, and visual perception (Tab. \ref{datasets}). 
% They provide diverse sensor data, including event cameras, LiDAR, and IMUs, enabling researchers to develop and test algorithms that are capable of handling dynamic, high-speed, and low-light environments. 
% By using these datasets, developers can improve the accuracy, efficiency, and robustness of their algorithms for mobile embodied perception tasks (\eg, localization, object detection, and SLAM).
As shown in Tab.~\ref{datasets}, event-based datasets serve as key benchmarks for robotics and visual perception.
% They combine data from event cameras, LiDAR, and IMUs, supporting algorithm development for dynamic, high-speed, and low-light environments. 
% These datasets help improve accuracy and efficiency in mobile embodied perception tasks such as localization and SLAM.

\textbf{MVSEC \cite{8288670}.} 
The MVSEC (Multi Vehicle Stereo Event Camera) dataset integrates data from event camera, stereo frame-based camera, LiDAR, IMU, motion capture, and GPS, making it a comprehensive resource for tasks such as stereo vision, SLAM, and autonomous driving. It provides synchronized stereo event streams and frame-based stereo images captured from vehicles navigating diverse driving environments, including urban roads and highways. 
% The dataset's real-world scenarios and multimodal sensor data make it invaluable for testing and evaluating event camera algorithms in dynamic and complex conditions typical of autonomous driving applications.

\textbf{DVS-Pedestrian \cite{sakhai2024pedestrianintentionpredictionadverse}.} 
The DVS-Pedestrian dataset is a benchmark dataset designed specifically for pedestrian detection and tracking, captured using a DVS in various urban environments. It features sequences of pedestrians performing various actions (walking, running, standing) under different lighting conditions and backgrounds. 
% Additionally, the dataset also provides three event-to-frame encoding methods: Frequency, Surface of Active Events(SAE), and Leaky Integrate-and-Fire (LIF), along with their respective Python implementations. A subset of the raw event data is pre-processed into frames(20ms intervals) using the SAE encoding method.

% This makes the DVS-Pedestrian dataset particularly useful for evaluating pedestrian detection algorithms that leverage the unique advantages of event cameras, such as high temporal resolution and low latency.

% \textbf{Gen1 \cite{detournemire2020largescaleeventbaseddetection}.}  
% The Gen1 dataset is a synthetic dataset designed for the development and evaluation of event-based vision systems. It includes high-resolution event camera data, captured in various dynamic scenarios that simulate real-world environments. The dataset provides both monocular and stereo event streams, offering temporal and spatial information with high precision. The synthetic nature of Gen1 allows for controlled conditions and a wide range of variations, such as changes in lighting, motion speed, and scene complexity. 
% Gen1 is particularly useful for evaluating event-based algorithms in tasks like motion estimation, depth estimation, and visual odometry, where real-time performance and accuracy in dynamic environments are critical.

\textbf{DDD20 \cite{9294515}.}  
The DDD20 (Dynamic Driving Dataset 2020) features synchronized frame camera, event camera, and IMU data from real-world urban and highway environments. It includes vehicles, pedestrians, cyclists and supports key tasks like object detection, tracking, and motion prediction. Additionally, DDD20 provides vehicle control signals (steering, throttle, braking) and is ideal for evaluating frame-event fusion in driving assistance systems.

% The DDD20 (Dynamic Driving Dataset 2020) is a comprehensive dataset designed for the development and testing of autonomous driving systems. It includes synchronized sensor data from stereo cameras, LiDAR, GPS, and IMU, collected in real-world driving environments. The dataset captures dynamic urban and highway scenes, featuring a wide range of road users such as vehicles, pedestrians, cyclists, and various road obstacles. With its high-resolution video, point cloud data, and detailed ground-truth annotations, DDD20 is an invaluable resource for tasks like object detection, tracking, semantic segmentation, and motion prediction. 
% The diverse and challenging driving scenarios make it an ideal benchmark for evaluating autonomous driving algorithms in realistic, dynamic conditions.

\textbf{DSEC \cite{9387069}.} 
The DSEC (Dynamic and Static Environment for Cars) dataset is a large-scale dataset for autonomous driving research, designed to test algorithms in both dynamic and static environments. It contains data from multiple sensors, including event cameras, LiDAR, and RGB cameras, captured from vehicles moving through urban streets and highways. The DSEC dataset is particularly useful for tasks like visual odometry, 3D reconstruction, and object tracking in dynamic environments. Its multimodal nature allows for the development of algorithms that can handle challenges of autonomous driving, such as dealing with fast-moving objects and varying light conditions.
% This dataset is an important resource for advancing autonomous driving technologies, providing real-world data for complex driving scenarios.

\textbf{TUM-VIE \cite{9636728}.} 
The TUM-VIE (TUM Visual-Inertial Evaluation) dataset is designed for evaluating VIO and SLAM algorithms. It contains synchronized data from event cameras, monocular cameras and IMUs, recorded during various motion scenarios, including both indoor and outdoor environments. This dataset is particularly useful for testing algorithms that combine visual and inertial data to estimate the camera’s position and orientation in real time.

\textbf{VECtor \cite{Gao_2022}.} The VECtor Event Dataset is the first SLAM benchmark dataset captured using a fully synchronized multi-sensor setup, including event-based and regular stereo cameras, RGB-D sensors, LiDAR, and IMU, with complete 6-DoF ground truth for diverse scenarios. It captures the full spectrum of motion dynamics and environmental conditions while providing precise calibration, specifically designed to address challenges unique to dynamic vision sensors.

\begin{table*}[t]
\centering
\vspace{-0.2cm}
\caption{Event-based datasets}
\vspace{-0.3cm}
\scriptsize
% \footnotesize
\label{tab:dataset_overview}
\begin{tabular}{>{\raggedright\arraybackslash}m{2.0cm}>
{\centering\arraybackslash}m{0.3cm}>
{\centering\arraybackslash}m{1.8cm}>
{\centering\arraybackslash}m{0.8 cm}>
{\centering\arraybackslash}m{1.8 cm}>
{\centering\arraybackslash}m{1.8 cm}>
{\centering\arraybackslash}m{1.0 cm}>
{\centering\arraybackslash}m{1.8 cm}}
\toprule
\textbf{Dataset name} & \textbf{Year} & \textbf{Data volume} & \textbf{Perspective} & \textbf{Participants} & \textbf{Lighting conditions} & \textbf{Annotation count} & \textbf{Application scenario} \\
\midrule
MVSEC \cite{8288670}               & 2018 & -                       & Dynamic & Pedestrians, vehicles                  & Daytime                  & -        & Driving, handheld scenes \\
DVS-Pedestrian \cite{sakhai2024pedestrianintentionpredictionadverse}      & 2019 & 0.1 hours, 4.6K annotations & Dynamic & Pedestrians                           & Daytime                  & 4.6K     & Walking street \\
% Gen1 \cite{detournemire2020largescaleeventbaseddetection}                & 2020 & 39 hours, 255K annotations & Dynamic & Pedestrians, vehicles                  & Daytime, night           & 255K     & Driving \\
DDD20 \cite{9294515}               & 2020 & 51 hours                & Dynamic & Pedestrians, vehicles                  & Daytime, night           & -        & Driving \\
1 Mpx Automotive \cite{perot2020learning} & 2020 & ~15 hours & Dynamic & Cars, pedestrians, two-wheelers & Daytime, night & 25M bboxes & Object Detection \\
% 1 Megapixel \cite{perot2020learning}         & 2020 & 15 hours, 25M annotations & Dynamic & Pedestrians, vehicles                  & Daytime, night           & 25M      & Driving \\
DSEC  \cite{9387069}               & 2021 & 1 hour, 390K annotations & Dynamic & Pedestrians, vehicles, scenes         & Daytime, night           & 390K     & Driving \\
TUM-VIE \cite{9636728}      & 2021 & 21 video clips          & Dynamic  & Static objects                       & Standard light, low-light & -        & 3D perception and navigation \\
FE108 \cite{zhang2021object} & 2021 & 1.5 hours & Dynamic & 21 object types & LL, HDR, fast motion & 208K frames & Object Tracking \\
VECtor \cite{Gao_2022}      & 2022 & 12 video clips          & Dynamic  & Static objects                       & Standard light, low-light, HDR & -        & SLAM \\

% PEDRo\cite{bbpprsPedro2023}                & 2023 & 0.5 hours, 43K annotations & Dynamic & Pedestrians                           & Daytime, night           & 43K      & Robotics detection \\
eTraM \cite{verma2024etrameventbasedtrafficmonitoring}               & 2024 & 10 hours, 2M annotations & Static  & Vehicles, pedestrians, micro-mobility & Daytime, night, twilight & 2M       & Intersections, roadways, streets \\
LLE-VOS\cite{li2024event}              & 2024 & 70 video clips          & Dynamic & Pedestrians, other targets             & Normal, low-light        & 5600     & Gym, classroom, zoo \\
REVD \cite{kim2024frequency} & 2024 & 21 sequences & Dynamic & Various scenes & Varied & 21 paired seq. & Video Deblurring \\
SDE \cite{liang2024towards} & 2024 & 91 sequences & Dynamic & Indoor, outdoor scenes & Low-light, normal & 30K+ pairs & Low-Light Enhancement \\
% LLE-DAVIS\cite{li2024event}            & 2024 & 90 video clips          & Dynamic & Various scene objects                 & Standard light, low-light & 6118     & Synthetic dataset \\

\bottomrule
\end{tabular}
\vspace{-0.7cm}
\label{datasets}
\end{table*}

\textbf{eTraM \cite{verma2024etrameventbasedtrafficmonitoring}.}  
% The eTraM (Event-based Traffic Monitoring) dataset is a comprehensive collection of event-based data designed for the development and evaluation of traffic monitoring and analysis systems using event cameras. It includes synchronized event camera data captured in urban road settings. The dataset features dynamic scenes with various types of vehicles, pedestrians, and complex traffic scenarios, providing high temporal resolution and spatial precision. 
The eTraM (Event-based Traffic Monitoring) dataset a fully event-based traffic perception benchmark captured using a high-resolution Prophesee EVK4 HD event camera. It features annotated traffic data with diverse vehicles, pedestrians, and micro-mobility objects under challenging lighting and weather conditions, including high glare, overexposure, underexposure, nighttime, twilight, and rainy days.
% eTraM is particularly useful for tasks such as real-time traffic flow analysis, vehicle tracking, and anomaly detection, offering a rich resource for researchers working on event-based perception and event-driven processing in autonomous driving applications.

\textbf{LLE-VOS \cite{li2024event}.}  
The LLE-VOS (Low-Light Event-based Video Object Segmentation) dataset is designed for event-based video object segmentation in low-light conditions, providing synchronized event and frame data with ground-truth masks. It captures challenging scenarios including night-time and indoor scenes.
% The LLE-VOS (Low-Light Event-based Video Object Segmentation) dataset is a specialized dataset designed for video object segmentation tasks in low-light environments using event cameras. It provides synchronized event-based data and traditional frame-based video, captured under challenging lighting conditions, including night-time and indoor scenes. The dataset features various dynamic objects, such as moving vehicles, pedestrians, and other environmental elements, with ground-truth object masks for segmentation tasks. 
% The LLE-VOS dataset is particularly useful for developing and evaluating video object segmentation algorithms that can work robustly in low-light and high-motion scenarios.
% This makes it an important resource for applications in autonomous driving, surveillance, and robotics, where visibility in poor lighting conditions is crucial.

\mpt{
\textbf{FE108} \cite{zhang2021object}. The FE108 dataset is a large-scale, frame-event-based resource for object tracking. It comprises 108 sequences with 21 object types under challenging conditions like low light, HDR, and fast motion. With high-frequency ground truth for both domains, it is highly suitable for evaluating multi-modal tracking algorithms.

\textbf{1 Mpx Automotive Detection Dataset} \cite{perot2020learning}. This dataset is the first large-scale, high-resolution benchmark for automotive object detection. It provides over 14 hours of recordings with more than 25 million bounding box annotations for cars, pedestrians, and two-wheelers, making it ideal for training robust detectors for autonomous driving.

\textbf{REVD} \cite{kim2024frequency}. The Real-world Event Video Deblurring (REVD) dataset is the first real-world benchmark for its task. It offers synchronized high-resolution blurred videos, sharp ground-truth counterparts, and event streams captured in scenes with extreme motion blur, serving as a critical resource for video restoration algorithms.

\textbf{SDE dataset} \cite{liang2024towards}. The SDE dataset is a large-scale, real-world benchmark for low-light image enhancement. It consists of over 30,000 spatially and temporally aligned image-event pairs captured in varied lighting conditions. This precise alignment enables the development of robust enhancement techniques for real-world scenarios.

}

\mpt{
Event-based datasets provide essential benchmarks for robotics and autonomous driving (e.g., MVSEC, DSEC, DDD20), SLAM/VIO (e.g., TUM-VIE, VECtor), and traffic monitoring (e.g., eTraM). They also support research in object detection/tracking (e.g., 1 Mpx Automotive, FE108, DVS-Pedestrian) and image restoration under motion blur or low light (e.g., REVD, SDE, LLE-VOS). By offering rich multimodal data, these datasets highlight the strengths of event cameras in high-speed, low-light, and HDR scenarios, advancing beyond traditional vision systems.
% Event-based datasets provide critical benchmarks across a wide spectrum of applications, proving crucial for advancing general robotics and autonomous driving (e.g., MVSEC, DSEC, DDD20), SLAM/VIO (e.g., TUM-VIE, VECtor), and traffic monitoring (e.g., eTraM). 
% In addition, such resources facilitate targeted research in high-performance object detection and tracking (e.g., 1 Mpx Automotive, FE108, DVS-Pedestrian), as well as image and video restoration under challenging conditions like motion blur (e.g., REVD) and low light (e.g., SDE, LLE-VOS). By providing rich, multimodal data, these benchmarks underscore the unique advantages of event cameras in high-speed, low-light, and high-dynamic-range scenarios, pushing the boundaries of what is possible compared to traditional vision systems.
}

\vspace{-0.2cm}
\subsection{Synthetic data generation: simulators and approaches} 
% \vspace{-0.1cm}
While deep learning has advanced event-based vision, progress depends on large, diverse datasets, yet collecting dense ground truth for tasks like optical flow is costly \cite{gallego2020event}.
Simulators address this by generating synthetic event data with pixel-accurate annotations in controllable environments, enabling systematic training and evaluation. 
This section reviews major simulation approaches, categorized as physics-based rendering, video-to-event conversion, neural rendering, and AI-driven generation, as well as quality assessment methods (Tab. \ref{tab:simulators_and_generators}).
% While deep learning has advanced event-based vision, model performance hinges on large-scale, diverse training data. However, acquiring real-world datasets (Sec. 2.6) with the dense, per-pixel ground truth needed for tasks like optical flow is often prohibitively expensive \cite{gallego2020event}. 
% To overcome these limitations, simulators provide controllable environments to generate vast quantities of synthetic event data with perfect, pixel-accurate annotations, enabling systematic training and evaluation across diverse conditions. 
% This section reviews prominent simulation techniques, categorized as physics-based rendering, video-to-event conversion, neural rendering, and AI-driven generation, along with methods for quality assessment.

\begin{table*}[t]
\centering
\caption{Summary of event camera simulators and generation methods}
\vspace{-0.3cm}
{
\tiny
% Required packages: \usepackage{booktabs}, \usepackage[table]{xcolor}, \usepackage{array}, \usepackage{colortbl}
\begin{tabular}{
    >{\raggedright\arraybackslash}m{0.1\textwidth}|
    >{\raggedright\arraybackslash}m{0.13\textwidth}|
    >{\raggedright\arraybackslash}m{0.24\textwidth}|
    >{\raggedright\arraybackslash}m{0.17\textwidth}|
    >{\raggedright\arraybackslash}m{0.24\textwidth}
}
\toprule
\textbf{Simulator/Method} & \textbf{Type} & \textbf{Core mechanism} & \textbf{Key features/Output} & \textbf{Noteworthy aspects/Limitations} \\ 
\hline

% --- Row 1 ---
\textbf{ESIM} 
& \cellcolor{gray!10} Physics-based 
& Interpolates log-intensity from rendered video to trigger threshold-based events. 
& \cellcolor{gray!10} Event streams, intensity frames, depth maps. 
& Foundational open-source tool. Generates 'perfect' events but lacks realistic noise models.\\ 
% \hline

% --- Row 2 ---
\cellcolor{gray!10} \textbf{InteriorNet} 
& Physics-based (Integrated) 
& \cellcolor{gray!10} Renders large, photo-realistic indoor scenes and converts the high-framerate video to events. 
& Event streams, RGB-D, IMU, semantics, optical flow. 
& \cellcolor{gray!10} Mega-scale dataset with rich ground truth (RGB-D, IMU) and realistic physics. \\ 
% \hline

% --- Row 3 ---
\textbf{v2e Toolbox} 
& \cellcolor{gray!10} Video-to-Events 
& Converts video to events using a detailed DVS pixel model with various noise sources. 
& \cellcolor{gray!10} Realistic DVS event streams. 
& High realism, especially for low-light, by modeling sensor non-idealities (noise, bandwidth).  \\ 
% \hline

% --- Row 4 ---
\cellcolor{gray!10} \textbf{V2CE} 
& Video-to-Events (Learning-based) 
& \cellcolor{gray!10} Uses a 3D UNet to predict event voxels, then a statistical model (LDATI) for continuous timestamps. 
& High-fidelity, continuous event streams. 
& \cellcolor{gray!10} Claims SOTA. Solves 'temporal layering' problem by generating continuous, non-discrete timestamps. \\ 
% \hline

% --- Row 5 ---
\textbf{EvDNeRF} 
& \cellcolor{gray!10} Neural Rendering 
& Trains a dynamic NeRF directly on event data to synthesize novel event streams. 
& \cellcolor{gray!10} Predicted event streams, intensity, depth. 
& First dynamic NeRF for events; can render from novel viewpoints. Training may be unstable. \\ 
% \hline

% --- Row 6 ---
\cellcolor{gray!10} \textbf{Text-to-Events} 
& Generative AI 
& \cellcolor{gray!10} Generates events directly from text prompts using a latent diffusion model. 
& Synthetic event frames. 
& \cellcolor{gray!10} Bypasses intermediate video generation. Currently limited to specific domains (e.g., gestures).  \\ 
% \hline

% --- Row 7 ---
\textbf{SENPI} 
& \cellcolor{gray!10} Differentiable Simulator 
& Fully differentiable PyTorch library modeling the entire event generation pipeline. 
& \cellcolor{gray!10} High-fidelity pseudo-event tensors. 
& Allows co-optimization of sensor parameters and network architectures.  \\ 
% \hline

% --- Row 8 ---
\cellcolor{gray!10} \textbf{EQS} 
& Quality Assessment Metric 
& \cellcolor{gray!10} Measures realism by comparing latent features from a network processing real vs. synthetic data. 
& A differentiable realism score. 
& \cellcolor{gray!10} Differentiable metric for raw events that correlates with sim-to-real performance. Can be used as a loss.  \\ 

\bottomrule
\end{tabular}
}
\vspace{-0.3cm}
\label{tab:simulators_and_generators}
\end{table*}

% The foundational principle of many event camera simulators is to first render high-frame-rate conventional video from a 3D virtual environment and then algorithmically convert these intensity frames into event streams based on a model of the event camera’s pixel behavior.

% One of the pioneering and most widely adopted tools in this category is ESIM (Event-based Simulator), which synthesizes events by interpolating logarithmic intensity changes between high-framerate rendered images and triggering events when a contrast threshold is exceeded \cite{the_event_camera_dataset_and_simulator}. 
% This foundational approach provides “perfect” event data with microsecond resolution. 
% Later extensions have incorporated more photorealistic rendering, adaptive rendering schemes, and simple noise models. 
% In contrast to this high-fidelity approach, earlier methods sometimes generated events more simply by thresholding the difference between consecutive video frames, a computationally light but temporally less precise technique \cite{towards_a_framework_for_end_to_end}. 
% For generating vast, photorealistic datasets, frameworks like InteriorNet provide a comprehensive solution, creating large-scale indoor scenes with event streams alongside corresponding RGB-D, IMU, and semantic data, complete with physics-based object movement and realistic lighting conditions \cite{interiornet_mega_scale_multi_sensor}.
\textbf{Physics-based rendering and simulation.}
This approach typically involves rendering high-framerate video from a 3D environment and then converting these frames to events based on a pixel model. 
A pioneering tool, ESIM \cite{mueggler2017event}, synthesizes events by interpolating logarithmic intensity changes to provide “perfect” data with microsecond resolution. 
For large-scale generation, frameworks like InteriorNet \cite{li2018interiornet} create photorealistic indoor scenes with event streams and corresponding multi-modal data (RGB-D, IMU, semantics). 
In contrast, some simpler methods just threshold frame differences, a faster but less precise technique \cite{kaiser2016towards}.

\textbf{Video-to-event conversion.}
To leverage existing video archives, video-to-event (V2E) converters transform standard video into event streams. 
The v2e toolbox \cite{hu2021v2e}, for instance, implements a sophisticated pixel model that accounts for non-ideal behaviors like bandwidth limits, threshold mismatch, and noise, yielding highly realistic DVS events. 
Other robust tools, including the V2CE Toolbox \cite{zhang2024v2ce} and the Prophesee Video to Event Simulator \cite{propheseemetavision11}, also provide robust methods for converting frame-based video into the event domain.

% To leverage the vast amount of existing, annotated conventional video data, video-to-event (V2E) conversion tools have become popular. 
% These tools transform standard video into event streams, significantly broadening the data available for training. 
% The v2e toolbox is a prominent example, designed to produce highly realistic DVS events by implementing a sophisticated pixel model \cite{v2e_from_video_frames}. 
% It accounts for numerous non-ideal behaviors, including photoreceptor bandwidth limitations, threshold mismatch, and various noise sources like hot pixels and temporal shot noise, making it particularly effective for simulating low-light conditions \cite{v2e_from_video_frames}. 
% Other similar tools, such as the V2CE Toolbox \cite{V2CE} and the Prophesee Video to Event Simulator \cite{Prophesee: Video to event simulator (2023), https://docs.prophesee.ai/ stable/samples/modules/core_ml/viz_video_to_event_simulator.html, accessed: 2024-07-17}, also provide robust methods for converting frame-based video into the event domain.

\textbf{Neural rendering and AI-driven event generation.}
More recently, neural rendering and generative AI have emerged. Neural Radiance Fields (NeRFs) offer a new direction; EvDNeRF \cite{bhattacharya2024evdnerf}, the first dynamic NeRF trained on events, can synthesize event streams of dynamic scenes from novel viewpoints. 
Pushing this further, generative AI models like Text-to-Events \cite{ott2024text} bypass intermediate frames entirely, using a latent diffusion model to generate event data directly from text prompts, though currently limited to specific domains like gestures.

% More recently, sophisticated techniques leveraging neural rendering and generative AI have emerged, pushing the boundaries of synthetic data generation.

% A novel approach utilizes Neural Radiance Fields (NeRFs) to synthesize event streams from a learned 3D scene representation. 
% EvDNeRF \cite{evdnerf_reconstructing_event_data} stands out as the first dynamic NeRF model trained directly from event data to predict event streams of dynamic scenes, including both rigid and non-rigid motion. 
% This allows it to function as a scene-specific event simulator that can generalize to novel viewpoints and fine temporal resolutions.

% At the forefront of generative AI, the Text-to-Events model bypasses the need for intermediate intensity frames entirely by generating event data directly from textual descriptions \cite{text_to_events_synthetic}. 
% It employs a latent diffusion model guided by a sparse autoencoder, which is specifically trained to produce the sparse data characteristic of event cameras. 
% This prompt-based system offers a powerful new paradigm for creating labeled event datasets for specific domains, though it is currently focused on gesture generation.

\textbf{Differentiable simulators and quality assessment.}
Recent research thrusts are differentiable simulation and quality assessment. 
SENPI \cite{greene2025pytorch} exemplifies the former, offering a fully differentiable framework for end-to-end co-optimization of sensor parameters and a network’s architecture. 
To address sim-to-real gap, Event Quality Score (EQS) \cite{chanda2025event} provides a metric by comparing latent features from a network processing real versus synthetic data. 
The resulting score correlates with real-world performance and can serve as a loss function to improve simulators.

% As simulators grow in complexity, two key areas of research have emerged: making the simulation process itself optimizable and developing methods to quantify the realism of the generated data.
% This has led to differentiable simulators like SENPI \cite{a_pytorch_enabled_tool_for_synthetic}, a PyTorch framework that models the entire event generation pipeline. Its key innovation is differentiability, which allows for the end-to-end co-optimization of physical sensor parameters and neural network architectures. 
% Concurrently, to address the critical sim-to-real gap, the Event Quality Score (EQS)  \cite{event_quality_score_eqs} provides a quantitative metric for realism. 
% It assesses the similarity of latent features from a deep network when processing real versus synthetic event streams, yielding a score that correlates with real-world performance and can be used as a loss function to guide the development of more realistic simulators.

% \vspace{-0.5cm}

% 需要一张图

\vspace{-0.2cm}
\section{Abstraction: Event Representation} \label{3}
    % \subsection{list}
    %     \subsubsection{time surface}
    %     \subsubsection{SNN}
    %     \subsubsection{...}
    % \subsection{Comparision}

\begin{figure*}[t]   
    \centering
    \includegraphics[width=1\textwidth]{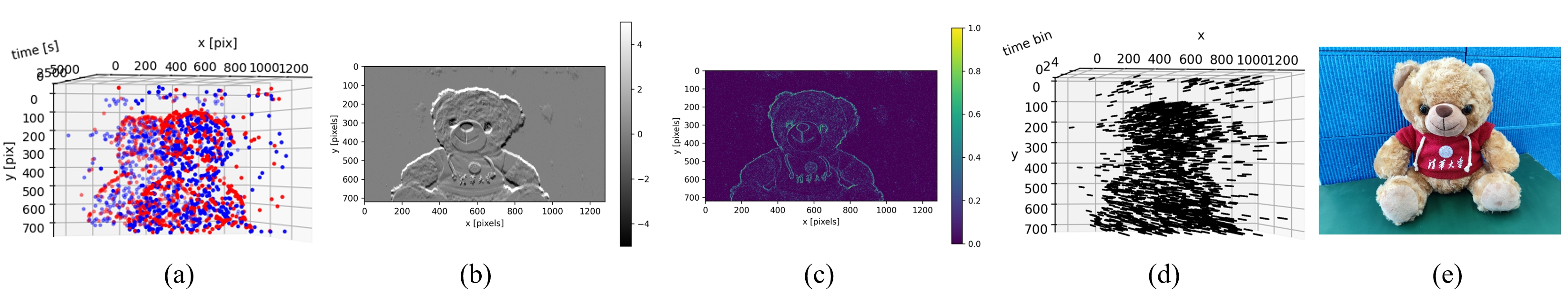}
    \vspace{-0.7cm}
    \caption{Event representations method. (a) Raw event. (b) 2D histogram. (c) Time surface. (d) Voxel grid. (e) RGB picture. }
    \vspace{-0.6cm}
    \label{representation}
\end{figure*}

% \notice{Figures about different representation}
% \notice{One paragraph writing style}
% Event data is often processed and transformed into various representations to extract meaningful information ("features") for solving specific tasks. Here, we review popular representations of event data, which range from simple, hand-crafted transformations to more elaborate methods.
Event data is often processed and transformed into various representations to extract meaningful information (features) for solving specific tasks. 
Here, we review popular representations of event data, which range from simple, hand-crafted transformations to more elaborate methods, as shown in \fig \ref{representation}.

\vspace{-0.2cm}
\subsection{Raw events}
% \vspace{-0.1cm}
Raw events offer high fidelity, retaining complete temporal and spatial information, making them ideal for event-driven processing, especially in applications using SNNs. 
While raw events provide detailed and precise information, they come with challenge of managing substantial data loads and ensuring proper alignment of events across time.

\textbf{Individual Events.}
Raw events $e_k\doteq(\mathbf{x}_k,t_k,p_k)$ are utilized by event-by-event processing methods such as probabilistic filters and Spiking Neural Networks (SNNs). These methods build additional information from past events or external knowledge and fuse it with incoming events asynchronously to produce an output 
\cite{wang2025eas, guo2025event}.

% Spiking Neural Networks (SNNs) are the neuromorphic counterpart to conventional neural networks, or Artificial Neural Networks (ANNs). While ANNs, particularly Deep Neural Networks (DNNs), are known for their accuracy, they often neglect computational complexity and energy consumption. SNNs excel in processing event camera data, which captures scene changes asynchronously, producing sparse, spatio-temporal data. By mimicking biological neural networks, SNNs integrate input spikes and fire when a threshold is reached. This spike-based processing is fault-tolerant and energy-efficient, only activating when necessary, unlike the continuous computation in ANNs.

\textbf{Event Packet.}
The event set \( E = \{ e_k \}^{N_e} \) retains precise timestamp and polarity information for each event. Selecting the appropriate packet size \( N_e \) is crucial to meet the assumptions of the algorithm (e.g., constant motion speed throughout packet’s duration), which varies depending on the task  \cite{lin2024compressed, li2024gaze}

% These networks are energy-efficient and well-suited for real-time tasks due to their spike-based processing, which mimics biological neural networks and only activates when necessary. However, the large volume of data generated by raw events can make direct processing complex, requiring advanced time management and synchronization mechanisms to handle the asynchronous nature of the data effectively. 

% \textbf{Advantages:}
% \begin{itemize}
%     \item High fidelity, retaining complete temporal and spatial information.
%     \item Suitable for event-driven processing, particularly SNNs, which are energy-efficient and ideal for real-time applications.
% \end{itemize}

% \textbf{Disadvantages:}
% \begin{itemize}
%     \item Large data volume, making direct processing complex.
%     \item Requires sophisticated time management and synchronization mechanisms.
% \end{itemize}

\vspace{-0.3cm}
\subsection{Event frame (2D Grid)}
\vspace{-0.1cm}
Events within a spatio-temporal neighborhood are converted into a 2D grid—often by counting events or accumulating polarity per pixel—forming an Event Frame compatible with standard image-based algorithms. While easy to implement, this representation can lose temporal information, suffer from motion blur in dynamic scenes, struggle under HDR conditions, and fail to fully exploit the sparsity of event data, reducing efficiency.
% Events in a spatio-temporal neighborhood are converted into an image (2D grid) through simple methods such as counting events or accumulating polarity pixel-wise. This image can then be fed into image-based algorithms.
% The Event Frame, a 2D grid representation, offers advantages like compatibility with existing image processing algorithms and ease of implementation. However, it can lose temporal information, causing motion blur in dynamic scenes, struggles in HDR environments, and may fail to fully leverage sparse nature of event data, leading to inefficiencies.

\textbf{Histogram.}
This representation converts events into histograms, offering an activity-driven sample rate. Although not fully aligned with the event-based paradigm, it has proven effective \cite{tang2025spike, kim2022real}. 
Traditional 2D histograms discretize events into bins, while the Activity-Aware Event Integration Module extends them with spatiotemporal operations, capturing finer details and improving dynamic-scene performance, particularly in semantic segmentation \cite{10477577}.
% This representation converts events into a histogram, providing a natural activity-driven sample rate. Although this practice is not ideal in the event-based paradigm, it has a significant impact \cite{tang2025spike, kim2022real}.
% While traditional 2D histograms capture spatiotemporal event data by discretizing the event stream into bins and counting occurrences over time, an AEIM (Activity-Aware Event Integration Module) enhances traditional 2D histograms by integrating complex spatiotemporal operations and attention mechanisms to capture finer scene details and prioritize high-confidence information, which reduces noise and improves performance in dynamic environments, making it especially effective for tasks of semantic segmentation \cite{10477577}.

\textbf{Time Surface.}
A Time Surface (TS) is a 2D map where each pixel stores a single time value (e.g., the timestamp of the last event at that pixel). Events are converted into an image whose "intensity" is a function of the motion history at that location, with larger values corresponding to more recent motion. TSs explicitly expose the rich temporal information of the events and can be updated asynchronously \cite{10401990, grimaldi2024robust, huang2023progressive}.

\vspace{-0.3cm}
\subsection{Spatio-temporal 3D grid representation}
\vspace{-0.1cm}
% A Voxel Grid is a space-time (3D) histogram of events, where each voxel represents a particular pixel and time interval. This representation preserves the temporal information of the events better by avoiding collapsing them onto a 2D grid. If polarity is used, the voxel grid becomes an intuitive discretization of a scalar field (polarity or brightness variation) defined on the image plane, with the absence of events marked by zero polarity. Each event's polarity may be accumulated on a voxel or spread among its closest voxels using a kernel, providing sub-voxel accuracy \cite{jiang2024learning, liu2023voxel, gehrig2024dense, mei2023deep, pan2024srfnet, wang2024evggs, cao2025embracing, xu2024dmr} .
A Voxel Grid is a 3D space-time histogram where each voxel corresponds to a pixel and time interval, preserving temporal information better than 2D projections.
With polarity, it forms a discretized scalar field on the image plane, using accumulation or kernel distribution for sub-voxel accuracy \cite{gehrig2024dense, cao2025embracing, xu2024dmr}.
This representation retains rich spatiotemporal detail for precise event tracking but requires more memory and computation, posing challenges for real-time use.

\vspace{-0.2cm}
\subsection{Customized representation}
% Customized event representations are tailored for specific tasks where traditional methods fall short, often combining spatial, temporal, or domain-specific features to improve efficiency and accuracy. For instance, in motion deblurring, \cite{wang2024asynchronous} introduces a representation using adaptive filtering to preserve high-frequency details, yielding sharper reconstructions. Similarly, \cite{chen2022ecsnet} proposes the 2D-1T Event Cloud Sequence (ECS), which separates spatial and temporal components to maintain sparsity and capture both geometry and motion, benefiting tasks like object classification and action recognition.
% These task-specific designs offer improved performance by leveraging relevant features and domain knowledge. However, they can be complex to design, harder to generalize across tasks, and risk introducing biases if not carefully constructed.

Customized event representations address task-specific limitations of traditional methods by combining spatial, temporal, or domain-specific features to boost efficiency and accuracy. 
For example, \cite{wang2024asynchronous} employs adaptive filtering for motion deblurring to preserve high-frequency details, while \cite{chen2022ecsnet} introduces the 2D-1T Event Cloud Sequence (ECS), which separates spatial and temporal components to retain sparsity and capture both geometry and motion, aiding recognition.
% Customized event representations are tailored for specific tasks where traditional methods fall short, often combining spatial, temporal, or domain-specific features to improve efficiency and accuracy. For instance, in motion deblurring, \cite{wang2024asynchronous} introduces a representation using adaptive filtering to preserve high-frequency details, yielding sharper reconstructions. Similarly, \cite{chen2022ecsnet} proposes the 2D-1T Event Cloud Sequence (ECS), which separates spatial and temporal components to maintain sparsity and capture both geometry and motion, benefiting tasks like object classification and action recognition.

\grs{
Task-specific designs improve performance by exploiting relevant features and domain knowledge, but their generalizability across tasks remains limited. 
For example, the adaptive filtering representation in \cite{wang2024asynchronous} is tightly coupled with a generative model and an extended kalman filter, making it powerful for dynamic tracking but hard to adapt to recognition or reconstruction tasks that require spatially structured outputs. 
Similarly, the ECS representation \cite{chen2022ecsnet} yields compact embeddings suitable for classification but lacks the explicit geometric and kinematic detail needed for precise tracking or mapping. 
Moreover, both are deeply embedded in their respective architectures, so transferring them to other paradigms demands significant redesign and retraining. 
Consequently, while customized representations demonstrate how tailoring to specific assumptions can achieve state-of-the-art performance, they also reveal the trade-off between specialization and generalization. 
Future research toward more unified or transferable event representations, along with standardized benchmarks for cross-task robustness, would be valuable for broadening applicability.
% These task-specific designs offer improved performance by leveraging relevant features and domain knowledge. Nevertheless, their generalizability across diverse tasks remains limited. The adaptive filtering representation in \cite{wang2024asynchronous}, for example, is tightly coupled with a parametric generative model and integrated into an Extended Kalman Filter framework, making it highly effective for dynamic target tracking but difficult to repurpose for recognition or reconstruction tasks that require spatially structured outputs. In contrast, the ECS representation \cite{chen2022ecsnet} produces compact feature embeddings well suited for classification, but these lack the explicit geometric and kinematic detail necessary for tasks such as precise tracking or mapping. Moreover, both representations are deeply embedded in their respective architectures, meaning that transferring them to other paradigms would entail significant redesign and retraining. As a result, while customized representations exemplify how tailoring to specific assumptions and objectives can yield state-of-the-art performance, they also highlight the inherent trade-off between specialization and generalization. Future research on more unified or transferable event representations would therefore be an important step toward broadening their applicability, and establishing standardized benchmarks for evaluating cross-task robustness would provide valuable guidance for advancing this direction.
}

\vspace{-0.2cm}
\section{Algorithm: Event processing} \label{4} 
\subsection{Event-based denoising}

\begin{wrapfigure}{r}{0.5\textwidth} % r 表示图在右边，l 表示图在左边
    \centering
    \begin{overpic}[width=1\linewidth,percent]{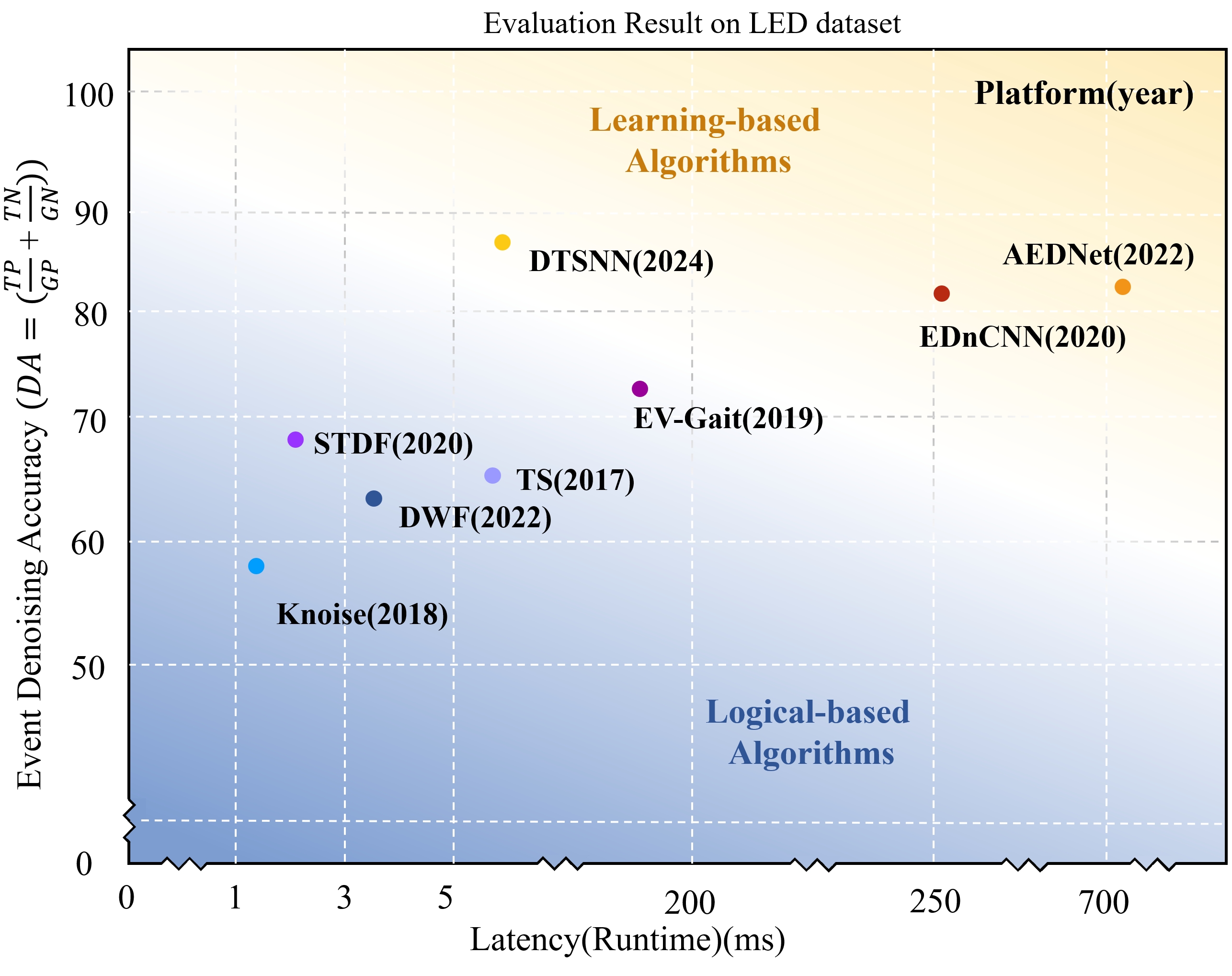}
        \put(39,42.5){\scriptsize\textbf{\cite{czech2016evaluating}}} % STDF
        \put(52,39.5){\scriptsize\textbf{\cite{lagorce2016hots}}} % TS
        \put(45,36){\scriptsize\textbf{\cite{khodamoradi2018n}}} % DWF
        \put(37,28.5){\scriptsize\textbf{\cite{guo2022low}}} % Knoise
        \put(67,44.5){\scriptsize\textbf{\cite{wang2019ev}}} % EV-Gait
        \put(91.5,51){\scriptsize\textbf{\cite{baldwin2020event}}} % EDnCNN
        \put(92,60.5){\scriptsize\textbf{\cite{fang2022aednet}}} % AEDNet
        \put(58,57){\scriptsize\textbf{\cite{duan2024led}}} % DTSNN
        
    \end{overpic}
    \vspace{-0.8cm}
    % \caption{Comparison of different event-based denoising algorithms in computational cost (Gflops) vs. performance (PSNR).
    \caption{\grs{Comparison of event-based denoising algorithms in computational cost vs. performance, all evaluated on the LED dataset.}
    % \notice{add reference}
    }
    \vspace{-0.3cm}
    \label{denoising}
\end{wrapfigure}

\textbf{Motivation.}
As event cameras see increasing use in high-speed, low-latency applications, effective denoising becomes crucial for reliable vision tasks. These bio-inspired sensors detect brightness changes with high temporal precision but are highly sensitive to noise, degrading event stream quality and affecting tasks like reconstruction and detection.
Noise arises from both external (e.g., ambient light changes) and internal sources (e.g., leakage currents), causing spurious events known as background activity (BA), which wastes bandwidth and reduces accuracy. 
Over time, hot pixels continuously emit false events, introducing hot noise.
Unlike conventional cameras that suppress noise via image integration, event cameras amplify it due to logarithmic encoding and differential sampling. 
This results in more BA, missed events, timing noise, and redundant trailing events after edge arrivals, complicating robust denoising efforts.

\textbf{Challenge.}
% Event cameras, while promising for high-speed and low-latency applications, face significant challenges in noise mitigation due to the diverse and complex nature of noise, limitations in data annotation, computational efficiency constraints, and non-uniform distribution of noise patterns.  
Event cameras face significant challenges in noise mitigation due to diverse noise sources, annotation limitations, computational constraints, and non-uniform noise distributions.
One major challenge lies in the diversity of noise types intrinsic to event cameras: 
\textit{(i) BA noise}, triggered by junction leakage currents or low-light conditions, generates spurious events.
\textit{(ii) Hot pixels}, common during prolonged usage or high-speed scenarios, produce persistent erroneous signals. 
\textit{(iii) Temporal noise} introduces stochastic timing variations, while \textit{(iv) structural noise} arises from edge inconsistencies or redundant trailing events. 
Their distinct spatiotemporal and statistical characteristics complicate simultaneous mitigation.
The lack of high-quality annotated datasets further limits supervised denoising methods, as creating paired data is labor-intensive, and synthetic datasets suffer from domain gaps. 
Multi-modal fusion with frame or IMU data helps, but struggles with extreme motion blur or lighting changes.

Computational efficiency is another bottleneck, many algorithms prioritize accuracy over latency, making them unsuitable for power-constrained systems, especially in low-light conditions where noise worsens. 
Moreover, the spatially and temporally non-stationary nature of noise challenges traditional methods that assume uniform noise. 
For example, BA noise often appears in localized bursts, while other noise types vary over time. 
Robust solutions must handle these irregularities for consistent performance.

\textbf{Literature review.}
% \notice{More citations?}
Event denoising has achieved notable advancements through diverse methodologies, including statistical approaches, filtering-based techniques, surface fitting and deep learning, significantly improving noise suppression in event-based vision (Fig. \ref{denoising}).
\grs{
To ensure fair comparison among these methods despite their diverse assumptions, we adopt the large-scale real-world LED dataset \cite{duan2024led}, which provides paired noisy/clean events under controllable illumination and noise levels. All methods are evaluated using the same metric denoising accuracy $DA=\frac{1}{2}(\frac{TP}{GP}+\frac{TN}{GN})$, which decomposes into signal retention (SR) and noise removal (NR), thus jointly capturing the two key aspects of denoising quality. As illustrated in Fig.~\ref{denoising}, this unified setting enables an objective and reproducible comparison of statistical, filtering-based, and deep learning strategies.
}
% These developments have significantly improved noise mitigation in event-based systems \fig \ref{denoising}.

\textit{Statistical methods}.
% Early event-denoising techniques primarily employed statistical methods to identify and remove outliers in event streams. 
% These methods typically assess event density within a local spatio-temporal neighborhood, classifying low-density events as noise \cite{feng2020event}. 
Early techniques relied on statistical analysis to identify outliers by evaluating event density in local spatio-temporal neighborhoods \cite{feng2020event}. 
Delbruck \textit{et al.} \cite{delbruck2008frame} pioneered density-based filtering, leveraging local context to suppress noisy events. 
Subsequent efforts \cite{khodamoradi2018n, guo2022low} enhanced these techniques with optimized event storage strategies to reduce computational complexity and improve processing efficiency.
However, these approaches often require manual parameter tuning to adapt to varying noise conditions, limiting their scalability and generalizability.

\textit{Filtering-based methods}.
To better exploit the temporal and asynchronous nature of event data, several filtering algorithms have been developed.
\textit{(i) Temporal filters} \cite{baldwin2019inceptive} remove redundant events by leveraging temporal correlations.
\textit{(ii) Spatial filters} \cite{ieng2014asynchronous} isolate motion-related events by analyzing pixel intensity changes.
\textit{(iii) Spatio-temporal filters} \cite{feng2020event, guo2021hashheat} combine both strategies to suppress background activity (BA) noise. For example, Liu \textit{et al.} \cite{liu2015design} showed that integrating spatial and temporal filtering reduces BA noise while preserving critical motion events.
% To better exploit the temporal and asynchronous nature of event data, several filtering-based algorithms have been developed.
% \textit{(i)} Temporal filters \cite{baldwin2019inceptive}  leverage event correlation to eliminate redundant activity.
% \textit{(ii)} Spatial filters \cite{ieng2014asynchronous} focus on pixel intensity changes to isolate events related to moving objects. 
% \textit{(iii)} Spatio-temporal filtering methods \cite{feng2020event, guo2021hashheat} combine these strategies to effectively suppress BA noise caused by intensity-independent events. 
% For instance, Liu \textit{et al.} \cite{liu2015design} demonstrated the advantages of integrating spatial and temporal filtering to reduce BA noise while preserving critical motion-related events.

% including temporal, spatial, and spatio-temporal filters. 
% \textit{(i)} Temporal filters \cite{baldwin2019inceptive} leverage the temporal correlation of events generated by object edges to eliminate redundant or ambiguous events.

\textit{Surface fitting techniques}.
Surface fitting methods offer an alternative approach, particularly effective for smoothing event data in continuous motion scenarios. 
Methods like EV-Gait \cite{wang2019ev} and the Guided Event Filter (GEF) \cite{duan2021guided} employ local plane fitting , optical flow estimation, and image gradients for noise smoothing. 
The time surface (TS) method \cite{lagorce2016hots, baldwin2019inceptive} transforming event streams into monotonically decreasing representations, effectively mitigates sparsity but exhibits limited performance in low light or highly dynamic scenes.
% to address sparsity issues. 
% These methods excel in scenarios with single-object motion but face challenges in low-light or highly dynamic environments.

\textit{Deep learning-based methods}. 
Recent advances in deep learning have revolutionized event denoising, enabling automated solutions by training on noisy-clean event pairs \cite{ruan2025pre,ruan2024distill}. Sparse feature learning (K-SVD \cite{xie2018dvs}, MLPF \cite{guo2022low}) methods focus on sparse feature extraction and event probability estimation. EDnCNN \cite{baldwin2020event} integrates frame and IMU data to classify events as signal or noise. EventZoom \cite{duan2021eventzoom} employs a U-Net architecture for efficient noise-to-noise denoising, achieving superior performance in handling noisy event streams. AEDNet \cite{fang2022aednet} processes raw DVS data while preserving inherent spatio-temporal correlations. EDformer \cite{jiang2024edformer} introduces an event-by-event transformer model, while EDmamba \cite{ruan2025edmamba} leverages state space modeling to achieve efficient noise-aware spatiotemporal denoising.

\vspace{-0.3cm}        
\subsection{Event-based filtering and feature extraction}
\vspace{-0.1cm}   
% \role{ciyu}
% \notice{Figure with accuracy and efficiency}，。
% \textbf{Motivation}.
% Unlike traditional frame-based systems that capture images at fixed intervals, event cameras generate data asynchronously, triggered only by significant changes in brightness. 
% This results in a sparse and highly dynamic output, necessitating specialized filtering techniques to isolate and retain the most meaningful events for downstream analysis, thereby enabling effective feature extraction.

% The primary objective of event filtering is to emphasize events that capture critical scene dynamics—such as motion and edges—while discarding irrelevant or redundant data. 
% This selective approach not only enhances the efficiency of algorithms for tasks like object detection, motion tracking, and scene reconstruction but also reduces computational overhead. 
% In fast-moving or highly dynamic environments, filtering ensures that only the most informative events are processed, improving both the accuracy and speed of downstream tasks.
% Moreover, effective filtering also plays a pivotal role in mitigating noise inherent in event data, which can otherwise compromise feature extraction for applications like object recognition and scene understanding. 
% This is particularly crucial in scenarios involving rapid motion or abrupt lighting changes, where noise can obscure essential information.
% By prioritizing relevant data, filtering ensures that critical events are preserved, facilitating more robust and accurate processing for real-time applications.

\textbf{Motivation}.
Unlike traditional frame-based cameras capturing images at fixed intervals, event cameras operate asynchronously, producing sparse data only when brightness changes occur. 
This dynamic output demands specialized filtering to isolate meaningful events for effective feature extraction.
Event filtering highlights important scene dynamics (\eg, motion and edges) while removing irrelevant or redundant data. 
This enhances efficiency and accuracy in tasks such as object detection, motion tracking, and scene reconstruction, especially in fast-moving or rapidly changing lighting conditions. 
By reducing noise and preserving critical features, filtering supports robust, low-latency processing essential for real-time applications.

\textbf{Challenge.}
The asynchronous nature of event data poses key challenges for filtering, as there’s no universal definition of a “significant” event. 
Variations in event frequency and timing complicate maintaining temporal coherence, which is vital for accurate motion representation during feature extraction. 
Balancing filtering complexity with real-time processing is also difficult, advanced methods may boost accuracy but risk adding latency.
Parameter tuning is another challenge, since optimal settings vary with environmental conditions and must adapt to dynamic, unpredictable scenes. This need for adaptability makes robust filtering crucial for reliable event-based vision. Overcoming these issues is essential to enhance feature extraction and ensure consistent real-world performance.

\begin{table*}[t]
\centering
\caption{Comprehensive Comparison of event filtering and feature extraction methods
\vspace{-0.3cm}
% \notice{2. Write each work individually and describe their cons./pros.?}
}
% \resizebox{\textwidth}{!}{
{
\tiny
% \scriptsize
% \footnotesize
% \arrayrulecolor{white}
\begin{tabular}{>{\centering\arraybackslash}m{0.20\textwidth}|>
    % {\raggedright\arraybackslash}m{0.18\textwidth}|>
    {\centering\arraybackslash}m{0.23\textwidth}|>
    {\centering\arraybackslash}m{0.23\textwidth}|>
    {\centering\arraybackslash}m{0.23\textwidth}}
% \hline
\toprule
\textbf{{\raisebox{0.25em}{Method}}} & \textbf{{\raisebox{0.25em}{Advantages}}} & \textbf{{\raisebox{0.25em}{Disadvantages}}} & \textbf{{\raisebox{0.25em}{Applications}}} \\ \hline
\cellcolor{gray!10} Frame-based methods \cite{harris1988combined,vasco2016fast} 
&   \begin{tabular}{@{}c@{}}
        High computational efficiency; \\
        Low memory consumption; \\
        Simple implementation.  
    \end{tabular}
&   \cellcolor{gray!10} \begin{tabular}{@{}c@{}}
        Poor accuracy in high-speed scenarios;\\
        Limited temporal precision;\\
        Loss of fine event details.  
    \end{tabular}
&   \begin{tabular}{@{}c@{}}
        Basic feature detection; \\
        Initial event processing. 
    \end{tabular}\\ 
% \hline
Surface of Active Events-based methods \cite{benosman2013event,8491018} 
&   \cellcolor{gray!10} \begin{tabular}{@{}c@{}}
        High temporal accuracy;\\
        Precise event timing preservation;\\
        Good feature localization.
    \end{tabular}
&   \begin{tabular}{@{}c@{}}
        High memory overhead;\\
        Reduced processing efficiency;\\
        Complex computational requirements.
    \end{tabular}
&   \cellcolor{gray!10} \begin{tabular}{@{}c@{}}
        Temporal-spatial feature detection;\\
        Optimization tasks.
    \end{tabular} \\ 
% \hline
\cellcolor{gray!10} Clustering-based methods \cite{hu2022ecdt,gao2024sd2event} 
&   \begin{tabular}{@{}c@{}}
        Linear time complexity;\\
        Efficient memory usage;\\
        Fast stream processing.
    \end{tabular} 
&   \cellcolor{gray!10} \begin{tabular}{@{}c@{}}
        Accuracy depends on parameters;\\
        Poor precision in complex scenes;\\
        Unstable performance.
    \end{tabular} 
&   \begin{tabular}{@{}c@{}}
        Dense event processing;\\ 
        Dynamic scene analysis.
    \end{tabular} \\ 
% \hline
Geometric transform-based methods \cite{gehrig2020eklt,seok2020robust,chui2021event} 
&   \cellcolor{gray!10} \begin{tabular}{@{}c@{}}
        High tracking accuracy;\\
        Precise motion estimation;\\
        Robust feature detection.
    \end{tabular}
&   \begin{tabular}{@{}c@{}}
        Heavy computational load;\\
        Low processing efficiency;\\
        High resource consumption.
    \end{tabular}
&   \cellcolor{gray!10} \begin{tabular}{@{}c@{}}
        Fast motion tracking; \\
        Extreme illumination scenarios.
    \end{tabular}\\ 
% \hline
\cellcolor{gray!10} Temporal filtering-based methods \cite{yi2023deep,wan2024event} 
&   \begin{tabular}{@{}c@{}}
        Fast processing speed;\\
        Efficient memory utilization;\\
        Good real-time performance.
    \end{tabular}
&   \cellcolor{gray!10} \begin{tabular}{@{}c@{}}
        Accuracy affected by noise;\\
        Precision loss in filtering;\\
        Detail preservation issues.
    \end{tabular}
&   \begin{tabular}{@{}c@{}}
        High temporal precision tasks; \\
        Real-time processing; \\
        Resource-constrained systems.
    \end{tabular}\\ 
% \hline
Asynchronous methods \cite{mueggler2017fast,alzugaray2020haste,chiberre2022long} 
&   \cellcolor{gray!10} \begin{tabular}{@{}c@{}}
        Low latency processing;\\
        High temporal accuracy;\\
        Efficient event handling.
    \end{tabular}
&   \begin{tabular}{@{}c@{}}
        Complex implementation; \\
        Resource intensive 
    \end{tabular}
&   \cellcolor{gray!10} \begin{tabular}{@{}c@{}}
        Low-latency applications;\\
        High-speed corner detection;\\
        Noise-heavy environments.
    \end{tabular}\\ 
% \hline
\cellcolor{gray!10}Hybrid methods \cite{li2019FAHarris,mohamed2021dynamic,harrigan2021rot} 
&   \begin{tabular}{@{}c@{}}
        High detection accuracy;\\
        Robust feature extraction; \\
        Multi-layer filtering.
    \end{tabular}
&   \cellcolor{gray!10}\begin{tabular}{@{}c@{}}
        Computationally heavy;\\
        Complex parameter tuning.
    \end{tabular}
&   \begin{tabular}{@{}c@{}}
        High-speed tasks; \\
        Multi-feature scenarios.
    \end{tabular}\\ 
% \hline
Neural networks-based methods \cite{afshar2020event,barchid2023spiking,yi2023deep,yao2023attention} 
&   \cellcolor{gray!10}\begin{tabular}{@{}c@{}}
        High feature accuracy; \\
        Tracks complex dynamics;\\
        Biologically inspired.
    \end{tabular}
&   \begin{tabular}{@{}c@{}}
        Training-intensive;\\
        High computational demands;\\
        Accuracy-speed tradeoffs.
    \end{tabular}
&   \cellcolor{gray!10} \begin{tabular}{@{}c@{}}
        Complex dynamic environments;\\
        Biological vision systems.
    \end{tabular}\\ 
% \hline
\cellcolor{gray!10}Frame-event hybrid methods \cite{best1992method,zhu2017event,rodriguez2020asynchronous} 
&   \begin{tabular}{@{}c@{}}
        High spatial-temporal accuracy;\\
        Precise feature matching;\\
        Robust performance.
    \end{tabular}
&   \cellcolor{gray!10}\begin{tabular}{@{}c@{}}
        Complex design;\\
        High computational cost;\\
        Complex resource management.
        \end{tabular}
&   \begin{tabular}{@{}c@{}}
        Robotics; \\
        Precision tasks; \\
        High-speed tracking.
    \end{tabular}\\ 
% \hline
\bottomrule
\end{tabular}
}%
% }
\vspace{-0.6cm}
\label{feature_extraction}
\end{table*}

\textbf{Literature review.}
% \notice{More citations?}
Event-based feature extraction has progressed from adapting frame-based methods to developing specialized event-driven approaches (Tab.~\ref{feature_extraction}). Early work modified traditional techniques, such as applying Harris corner detection \cite{harris1988combined} to event accumulation frames. A major advance was the Surface of Active Events (SAE) \cite{benosman2013event}, which records the timestamp of the latest event per pixel, preserving temporal precision while improving efficiency and feature extraction performance.
% Event-based feature extraction has evolved significantly, transitioning from adaptations of conventional frame-based methods to specialized event-driven approaches (Tab. \ref{feature_extraction}). 
% Early work extended traditional techniques, such as adapting Harris corner detection \cite{harris1988combined} to binary frames derived from event accumulation.
% A key breakthrough came with the introduction of the Surface of Active Events (SAE) \cite{benosman2013event}, a representation that stores the temporal information of the most recent events at each pixel. 
% This innovation bridged the gap between frame-based and event-driven paradigms, enabling algorithms to preserve the temporal precision inherent in event cameras while improving computational efficiency and feature extraction performance.

Feature extraction methods have advanced in both accuracy and efficiency. eFAST \cite{mueggler2017fast} shifted from gradient-based to faster comparison-based operations tailored for event data, while Arc* \cite{8491018} improved detection speed and repeatability through refined SAE filtering. Hybrid methods like FA-Harris \cite{li2019FAHarris} and TLF-Harris \cite{mohamed2021dynamic} balanced efficiency and robustness via candidate selection and multi-layer filtering. FEAST \cite{afshar2020event} introduced unsupervised extraction with spiking neuron-like units, and ROT-Harris \cite{harrigan2021rot} used tree-based processing to surpass traditional 2D approaches. Together, these works highlight a shift toward practical, scalable implementations.
% The evolution of extraction methodologies has seen notable improvements in both accuracy and efficiency. 
% The development of eFAST \cite{mueggler2017fast} marked a transition from computationally intensive gradient-based methods to faster comparison-based operations optimized for event data. Subsequent methods, such as the Arc* algorithm \cite{8491018}, enhanced detection speed and corner repeatability by refining SAE filtering techniques. 
% Hybrid approaches, including FA-Harris \cite{li2019FAHarris} and TLF-Harris \cite{mohamed2021dynamic}, integrated efficient candidate selection with robust multi-layer filtering, balancing computational complexity and accuracy. 
% FEAST \cite{afshar2020event} introduced an unsupervised feature extraction approach using spiking neuron-like units with individual selection thresholds, while ROT-Harris \cite{harrigan2021rot} leveraged tree-based processing to enhance feature extraction beyond traditional 2D methods. 
% These advancements reflect the field’s shift toward practical and scalable implementations.

Parallel to advancements in feature extraction, event filtering has emerged as a critical preprocessing step for robust feature extraction, leveraging the asynchronous and high-resolution temporal data of event cameras. 
Temporal filtering-based methods have significantly improved data quality and computational efficiency by prioritizing meaningful scene changes \cite{yi2023deep, wan2024event}. 
Clustering-based methods, such as eCDT \cite{hu2022ecdt}, dynamically group events to represent dense streams compactly while minimizing complexity. 
However, such methods often face challenges in maintaining robustness under varying conditions due to sensitivity to parameter configurations \cite{gao2024sd2event}.

% Parallel to advancements in feature extraction, event filtering has emerged as a critical preprocessing step for robust feature extraction, leveraging the asynchronous and high-resolution temporal data of event cameras. 
% Temporal filtering-based methods has significantly improved data quality and computational efficiency by prioritizing meaningful scene changes \cite{yi2023deep, wan2024event}. 
% Clustering-based methods, such as eCDT \cite{hu2022ecdt}, dynamically group events to represent dense streams compactly while minimizing complexity. 
% However, such methods often face challenges in maintaining robustness under varying conditions due to sensitivity to parameter configurations \cite{gao2024sd2event}.

Parametric filtering has further advanced event-based vision by employing geometric transformations to filter irrelevant data. 
For instance, EKLT \cite{gehrig2020eklt} aligns events to improve feature extraction, while curve-fitting techniques construct smooth spatio-temporal trajectories \cite{seok2020robust}, excelling in scenarios with rapid motion or extreme illumination.
Asynchronous methods, such as HASTE \cite{alzugaray2020haste}, process individual events in real time using hypothesis-driven transformations, and proximity-based trackers continuously refine feature locations, effectively suppressing noise and supporting low-latency applications \cite{chiberre2022long}.

% The integration of filtering and feature extraction methods has significantly enhanced the robustness and efficiency of event-based vision systems. 
% For example, combining filtering mechanisms with classical algorithms, such as Harris or FAST, has enabled robust corner detection, while shape detection methods using iterative closest point (ICP) and Hough transforms have improved performance in high-speed and high-dynamic-range scenarios \cite{best1992method}. 
% Modern tracking techniques leverage the low latency and high temporal resolution of event cameras by employing probability-based associations and spatio-temporal constraints, further enhancing feature stability and tracking accuracy \cite{zhu2017event, rodriguez2020asynchronous}.
Integrating filtering with feature extraction has greatly improved the robustness and efficiency of event-based vision systems. For instance, combining filters with classical algorithms like Harris or FAST enables reliable corner detection, while shape detection methods using ICP and Hough transforms enhance performance in high-speed, high-dynamic-range settings \cite{best1992method}. Modern tracking approaches also exploit the low latency and high temporal resolution of event cameras by incorporating probabilistic associations and spatio-temporal constraints, resulting in more stable features and improved tracking accuracy \cite{zhu2017event, rodriguez2020asynchronous}.

% Neural network-based filtering represents a promising frontier, combining asynchronous event processing with biologically inspired architectures. 
% Pulse-based neural networks \cite{barchid2023spiking}, for example, exploit high temporal resolution for precise feature tracking, demonstrating strong potential in dynamic and complex environments. 
% However, current neural network architectures still face challenges in achieving detection accuracy comparable to traditional convolutional models \cite{yi2023deep, yao2023attention}.
% The continued integration of filtering, feature extraction, and tracking methodologies is driving progress, expanding the application scope of event-based vision systems while maintaining computational efficiency and robustness in diverse real-world scenarios.
Neural network-based filtering integrates asynchronous event processing with biologically inspired architectures. Pulse-based neural networks \cite{barchid2023spiking} exploit the high temporal resolution of event data for precise feature tracking, showing strong potential in dynamic environments. Yet, current models still fall short of the detection accuracy achieved by convolutional networks \cite{yi2023deep, yao2023attention}. Ongoing efforts to unify filtering, feature extraction, and tracking are pushing event-based vision toward broader applicability while preserving efficiency and robustness in real-world scenarios.

\vspace{-0.2cm}
\subsection{Event-based matching}
% \role{xinyu}
\textbf{Motivation.}
% Matching refers to the identification of corresponding features between two or more event streams captured at different times or from different viewpoints. 
% These correspondences serve as a foundation for tasks such as visual odometry, video interpolation, and other event-based mobile sensing applications, enhancing accuracy and robustness in high-frequency dynamic environments. 
% Unlike RGB cameras, which capture images at fixed intervals, event cameras respond only to brightness changes, making them particularly well-suited for dynamic scenes. 
% They excel at capturing edges and texture information within the field of view while avoiding the redundant data accumulation typical of frame-based systems.
% Additionally, the temporal information embedded in event data enables the inference of object motion.
% The unique characteristics of event cameras effectively mitigate challenges faced by traditional cameras, including motion blur and low frame rates.
% Consequently, event cameras offer inherent advantages for matching-based tasks, with even single-modal event data demonstrating strong performance in these applications.
Matching involves identifying corresponding features between event streams captured at different times or viewpoints, forming the basis for tasks like visual odometry, video interpolation, and mobile perception. 
Event cameras, unlike RGB cameras that operate at fixed intervals, trigger only on brightness changes, making them ideal for dynamic scenes. 
They capture edges and textures efficiently while avoiding redundant data.
The precise temporal information in event data also enables accurate motion inference. By mitigating issues like motion blur and low frame rates common in traditional cameras, event cameras offer distinct advantages for matching tasks, even with single-modal data, demonstrating strong performance in high-frequency, fast-changing environments.

% Matching refers to identifying corresponding features between two or more event streams captured at different times or from different viewpoints. These matching results can be further applied to tasks such as visual odometry, video interpolation, and other visual perception tasks, providing higher accuracy and robustness in high-frequency dynamic environments.
% Since event cameras capture brightness changes, they are better suited than RGB cameras in moving scenes for capturing edges and texture information within the field of view, avoiding the collection of redundant information typical of traditional cameras operating at fixed frame rates. Additionally, the time elements contained in the event data can be used to infer the motion of objects. Furthermore, the inherent characteristics of event cameras, such as high frequency and high dynamic range, address common issues with traditional cameras, such as motion blur and low frame rates. As a result, event cameras have a natural advantage in matching tasks, and even single-modal event data performs well in these tasks.

\textbf{Challenge.}
% Event-based matching tasks, however, present several significant challenges. 
% First, the sparsity of event data in the pixel space and its non-uniform temporal distribution complicate the extraction of sufficient and robust features. 
% This issue is particularly pronounced in low-light conditions or static scenes, where fewer events are generated, making matching significantly more difficult. 
% These scenarios demand the development of specialized feature extraction algorithms tailored to the unique characteristics of event cameras for event-based matching.  
% Moreover, the ultra-high temporal resolution of event cameras generates a substantial volume of event data, leading to a significantly higher computational load compared to traditional frame-based cameras. 
% Addressing this challenge requires not only improving the efficiency of matching algorithms but also ensuring high accuracy with minimal latency to meet the real-time demands of mobile and resource-constrained computing applications.
Event-based matching faces major challenges due to the sparsity of events in pixel space and their uneven temporal distribution, which hinder robust feature extraction, especially in low-light or static scenes. 
This calls for specialized methods adapted to event camera properties. 
Moreover, the ultra-high temporal resolution produces massive data volumes, raising computational costs. 
Achieving real-time performance in resource-limited settings thus requires algorithms that balance efficiency and accuracy, minimizing latency while ensuring reliability in dynamic env..

\begin{table*}[t]
\centering
\tiny
\caption{Event-based Matching Algorithm Comparison}
\vspace{-0.3cm}
\label{matching}
\begin{tabular}{>{\centering\arraybackslash}m{0.18\textwidth} >{\centering\arraybackslash}m{0.18\textwidth} >
{\centering\arraybackslash}m{0.12\textwidth} >
{\centering\arraybackslash}m{0.18\textwidth} >
{\centering\arraybackslash}m{0.2\textwidth}}
\toprule
\renewcommand{\arraystretch}{2.0}
\textbf{Algorithm} & \textbf{Sample} & \textbf{Input} & \textbf{Advantages} & \textbf{Disadvantages} \\
\midrule
\multirow{3}{*}{\makecell{ \vspace{0.75ex} Local feature matching \\ \cite{Li_2024_CVPR,ren2024simpleeffectivepointbasednetwork,10342794,hou2022fefusionvprattentionbasedmultiscalenetwork,calonder2008keypoint,BAY2008346,6126544,calonder2010brief,10629077,wang2024fedetrkeypointdetectiontracking}}} &
  \multirow{3}{*}{\makecell{Optical Flow \\ Feature descriptor: SIFT/SURF/ORB}} &
  \multirow{3}{*}{Batch} &
  \multirow{3}{*}{\makecell{Low latency, noise resistance \\ Low computational resource}} &
  \multirow{3}{*}{\makecell{Difficult to handle global motion \\ Difficult to handle repeat texture \\Need further extraction method}} \\
% \cline{2-2}
    & & & &      \\
    & & & &      \\
\midrule
\multirow{4}{*}{\makecell{Optimization-based matching \\ \cite{Gallego_2018,9156854,Stoffregen2019EventCC}}} &
  \multirow{4}{*}{\makecell{Contrast maximization \\ Levenberg-Marquardt \\ Graph optimization}} &
  \multirow{4}{*}{Batch/Asynchronous } &
  \multirow{4}{*}{\makecell{Global consistency \\ Higher accuracy after iterations \\ Detailed design of optimizer}} &
  \multirow{4}{*}{\makecell{ Higher computational complexity \\ Higher time latency \\ Hard to converge \\ Sensitive to initial values}}  \\
% \cline{2-2}
    &  &       &     &        \\
% \cline{2-2}
    &    &       &     &     \\
    &    &       &     &     \\
\midrule
\multirow{3}{*}{\makecell{Neural network-based matching \\ \cite{wu2024lightweighteventbasedopticalflow,chen2024eventbasedmotionmagnification,deng2022voxel,rebecq2019events,schaefer2022aegnnasynchronouseventbasedgraph}}} &
  CNN &
  \multirow{3}{*}{Batch/Asynchronous} &
  \multirow{3}{*}{\makecell{Adaptive learning \\ Less manual intervention \\ Efficient inference}} &
  \multirow{3}{*}{\makecell{Large amounts of training data \\ Need parameter tuning \\ Poor generalization ability}} \\
% \cline{2-2}
    & SNN &       &     &        \\
% \cline{2-2}
    & GNN &       &     &        \\
\bottomrule
\end{tabular}
\vspace{-0.5cm}
\end{table*}

% \vspace{0.5em}
\textbf{Literature review.}
% \notice{More citations?}
In visual data processing, matching algorithms are essential for motion estimation, visual odometry, and feature tracking. They fall into three main categories—local feature, optimization-based, and deep learning-based matching—each with distinct strengths and limitations (Tab.~\ref{matching}).
% In visual data processing, matching algorithms play a pivotal role in tasks such as motion estimation, visual odometry, and feature tracking.
% These algorithms can be broadly classified into three categories: local feature matching, optimization-based matching, and deep learning-based matching, each offering distinct approaches, strengths, and limitations Tab. \ref{matching}.  

\textit{Local feature matching methods}. 
These methods focus on detecting and associating local features within visual data \cite{10.1177/0278364914554813}. Techniques such as optical flow \cite{Li_2024_CVPR, ren2024simpleeffectivepointbasednetwork} and descriptors like SIFT \cite{calonder2008keypoint}, SURF \cite{BAY2008346}, and ORB \cite{6126544} exemplify this approach. They offer low latency, robustness to noise, and low computational demands, making them well-suited for mobile, autonomous, and robotic applications \cite{10629077,wang2024fedetrkeypointdetectiontracking}.
However, local feature matching faces challenges with event streams, especially under large global motion, repetitive textures, or heavy reliance on specific feature extractors \cite{Whelan2015ElasticFusionDS}. Its performance also declines in scenes with sparse features or dramatic changes, underscoring the need for more advanced and robust techniques.

% These methods emphasize detecting and associating local specific features within visual data\cite{7035876,7354184,10.1177/0278364914554813,10.5555/3222346.3222583}. 
% Techniques such as optical flow \cite{Li_2024_CVPR,ren2024simpleeffectivepointbasednetwork,10342794,hou2022fefusionvprattentionbasedmultiscalenetwork} and feature descriptors like SIFT \cite{calonder2008keypoint}, SURF \cite{BAY2008346}, and ORB \cite{6126544,calonder2010brief} exemplify this category. 
% These methods are characterized by low latency, robustness to noise, and minimal computational requirements, making them particularly well-suited for resource-constrained applications such as mobile devices, autonomous driving, and robotic vision \cite{10629077,wang2024fedetrkeypointdetectiontracking}.
% However, local feature matching struggles with event stream and challenges like large-scale global motion, repeated textures, and a reliance on specific feature extraction methods \cite{Klein2007ParallelTA, Whelan2015ElasticFusionDS}.
% Performance degrades in scenarios with sparsely distributed features or drastic scene changes, also highlighting the need for more advanced techniques to enhance robustness and accuracy.  

\textit{Optimization-based matching methods}. 
These methods aim to ensure global consistency by solving optimization problems to align event data or trajectories. Techniques such as contrast maximization \cite{Gallego_2018, 9156854, Stoffregen2019EventCC}, the Levenberg-Marquardt algorithm, and graph-based optimization are widely used. They offer high flexibility and accuracy, supporting both batch and asynchronous processing, which makes them valuable in event camera applications.
However, these approaches are computationally demanding and sensitive to initial conditions—poor initialization can lead to local optima or non-convergence. Their high memory and processing requirements also limit real-time usability, especially for large-scale data. Achieving a balance between efficiency and accuracy remains a core challenge.

% These methods aim to achieve global consistency by solving mathematical optimization problems to align event data or trajectories.
% Popular techniques include contrast maximization \cite{Gallego_2018,9156854,Stoffregen2019EventCC}, the Levenberg-Marquardt algorithm, and graph-based optimization. 
% These methods are highly flexible and capable of achieving high accuracy through batch processing or asynchronous updates, making them valuable in event camera research. 
% However, they are computationally intensive and sensitive to initial parameter settings. 
% Poor initialization can lead to non-convergence or convergence to local optima. 
% Additionally, the high computational demands and memory requirements make these methods less suitable for real-time applications, especially when processing large-scale data.
% Balancing computational efficiency and matching accuracy remains a significant challenge in this domain.  

\textit{Deep learning-based matching methods}.
These methods mark a transformative shift by employing CNNs \cite{deng2022voxel}, SNNs \cite{wu2024lightweighteventbasedopticalflow,chen2024eventbasedmotionmagnification}, and GNNs \cite{schaefer2022aegnnasynchronouseventbasedgraph} to automatically learn representations and matching strategies, surpassing handcrafted descriptors in accuracy. 
GNNs are well-suited to the asynchronous nature of event streams and excel in high-speed, dynamic environments. 
However, these models are computationally demanding, rely on large annotated datasets, and struggle to generalize, while training remains time-consuming and requires careful hyperparameter tuning.

Selecting the most suitable matching algorithm depends on the specific task requirements, including computational constraints, real-time processing needs, data characteristics, and scene complexity.
As technology advances, these algorithms are likely to evolve and converge, enabling the development of more versatile and robust visual matching techniques capable of addressing a broader spectrum of applications.

\vspace{-0.2cm}
\subsection{Event-based mapping }
% \role{xinyu}
% \notice{Figure with accuracy and efficiency}
\textbf{Motivation}
Building on event matching, asynchronous event streams from different viewpoints can be fused across poses to incrementally reconstruct a dense 3D map. Event-based perception leverages spatial sparsity and high temporal resolution, reducing computation by focusing on regions of change and minimizing motion blur to preserve edges during fast motion. These properties enable real-time, efficient, and robust mapping in dynamic environments.

\textbf{Challenge}.
Using cameras—including event cameras—for mapping and depth estimation is feasible but challenging. Monocular cameras lack direct depth information, requiring techniques like multi-frame perspective changes or fusion with auxiliary sensors (e.g., IMUs) to infer depth. Stereo event camera setups can estimate depth more directly through disparity.
However, event cameras’ high temporal resolution leads to significant computational demands, as processing their continuous asynchronous data in real time is resource-intensive. Additionally, aligning event data with other sensor modalities complicates feature extraction and synchronization calibration. These challenges intensify in dynamic environments, where maintaining sensor consistency is vital for accurate depth estimation.

\begin{table*}[t]
\scriptsize
\caption{
% \notice{add reference?okk} 
Comparison of different mapping methods.
% in terms of advantages and disadvantages.
}
\vspace{-0.3cm}
\label{tab:comparison}
\centering
\begin{adjustbox}{width=\textwidth}
\begin{tabular}{>{\raggedright\arraybackslash}m{3.5cm}>
                {\raggedright\arraybackslash}m{6.3cm}>
                {\raggedright\arraybackslash}m{6.3cm}}
\toprule
\makecell{\textbf{Types of Mapping Methods}} & \textbf{Advantages} & \textbf{Disadvantages or Challenges} \\
\midrule
\multirow{2}{*}{\makecell{ \vspace{0.75ex}\textbf{Frame-based mapping}\cite{Mueggler2014Eventbased6P, Yuan2016FastLA, Bertrand2020EmbeddedEV, Kueng2016LowlatencyVO, Zhu2019NeuromorphicVO, Forster2014SVOFS}}} & 
\makecell[l]{
\ding{51} Improves the accuracy of depth estimation\\
\ding{51} Enhances the quality of map construction \\
\ding{51} Lower latency and computation cost
} & 
\makecell[l]{
\ding{55} Requires measuring and updating depth information \\ between current and initial detection frames, potentially \\ increasing computational load
} \\
\midrule
\multirow{1.5}{*}{\makecell{ \vspace{0.75ex}\textbf{Filter-based mapping} \cite{Weikersdorfer2012EventbasedPF, Chamorro2022EventBasedLS, Rebecq2018EMVSEM, Weikersdorfer2013SimultaneousLA} }} & 
\makecell[l]{
\ding{51} High robustness, suitable for rapidly changing environments \\
\ding{51} Dynamically adjusts the system’s map representation
} & 
\makecell[l]{
\ding{55} Requires a large number of parameters to represent camera \\ poses, potentially increasing computational complexity
} \\
\midrule
\multirow{2}{*}{\makecell{ \vspace{0.75ex}\textbf{\ \ \ Continuous-time mapping} \\ \cite{Mueggler2015ContinuousTimeTE, Liu2022AsynchronousOF,Gentil2020IDOLAF} }} & 
\makecell[l]{
\ding{51} Reduces parameter complexity \\
\ding{51} Improves mapping accuracy and efficiency \\
\ding{51} Simultaneously updates camera poses and 3D landmarks
} & 
\makecell[l]{
\ding{55} Requires handling continuous curve interpolation and \\ optimization problems\\
\ding{55} Increasing computational difficulty \\
\ding{55} High latency due to frequent states update
} \\
\midrule
\multirow{2}{*}{\makecell{ \vspace{0.75ex}\textbf{Spatio-temporal Consistency}\cite{Liu2021SpatiotemporalRF}}} & 
\makecell[l]{
\ding{51} Improves the accuracy of map construction \\
\ding{51} Optimizes motion parameters \vspace{0.3ex}
} & 
\makecell[l]{
\ding{55} Requires iteratively searching for the closest points and \\ applying a pruned ICP algorithm \\
\ding{55} High computational cost and latency
} \vspace{0.1ex}\\
\bottomrule
\end{tabular}
\end{adjustbox}
\label{mapping}
\vspace{-0.7cm}
\end{table*}

\textbf{Literature review}.
% \notice{More citations?}
Mapping plays a fundamental role in constructing a 3D representation of the environment based on visual features captured by cameras. 
With advancements in event-based vision, event cameras have become an increasingly valuable source of information in visual simultaneous localization and mapping, offering high-frequency, asynchronous data well-suited for real-time processing in dynamic scenes.
Various event-based mapping approaches have been proposed, each exhibiting distinct characteristics and advantages Tab. \ref{mapping}.

\textit{Frame-based mapping methods}.
Frame-based mapping methods often use event-derived 2D representations with depth filters to iteratively refine scene depth via feature triangulation. For example, \cite{Bertrand2020EmbeddedEV} estimates poses relative to planar structures by minimizing reprojection errors, while \cite{Zhu2019NeuromorphicVO} adopts the SVO algorithm \cite{Forster2014SVOFS} for pose estimation from event feature correspondences. 
These methods typically model depth with Gaussian–uniform filters, updating estimates from feature comparisons guided by poses. Additionally, \cite{Yuan2016FastLA} improves mapping by solving pose estimation as a least-squares problem on 2D–3D line constraints.

\textit{Filter-based mapping methods}.
Filter-based mapping techniques address the asynchronous nature of event data by continuously updating maps during camera tracking. Line-based vSLAM refines maps by measuring distances between incoming events and reprojected 3D lines, generating point cloud reconstructions. These methods often use the Hough transform to extract 3D line features, leveraging spatial correlations for robustness in dynamic scenes. For example, \cite{Weikersdorfer2012EventbasedPF} estimates poses via distances between back-projected event rays and planar points, while \cite{Weikersdorfer2013SimultaneousLA} improves precision with a probabilistic measurement function on planar surfaces. More recent strategies, such as \cite{Chamorro2022EventBasedLS}, update filter states by evaluating event-to-line distances, often combining EMVS \cite{Rebecq2018EMVSEM} with Hough-based line extraction to strengthen event-line associations and enhance spatial accuracy.

\textit{Continuous-Time mapping methods.}
To reduce the high parameter count in filter-based discrete pose representations, continuous-time mapping replaces discrete poses with smooth trajectory models such as B-splines or Gaussian processes, enabling interpolation from local control states. This lowers complexity and supports joint optimization of poses and landmarks, improving accuracy and efficiency. For example, \cite{Mueggler2015ContinuousTimeTE} employs B-splines, while \cite{Liu2022AsynchronousOF} use Gaussian process motion models to interpolate poses at arbitrary timestamps. \cite{Liu2022AsynchronousOF} further integrates incremental SfM for consistent refinement of control states and landmarks, and \cite{Liu2021SpatiotemporalRF} introduces a spatio-temporal constraint based on equal-time event pairs to enhance rotational accuracy. In practice, these methods iteratively establish correspondences and apply pruned ICP for spatial consistency, yielding more precise event-based maps.

\vspace{-0.2cm}
\section{Acceleration} \label{5}
Event-based perception generates data only when pixel intensity changes, drastically reducing redundancy and offering efficiency, low power, and minimal latency, which is well-suited for high-speed, energy-constrained applications.
With growing demand on mobile agents such as drones, vehicles, and wearables, real-time and efficient processing is critical but remains challenging under limited resources. 
Balancing accuracy and efficiency thus becomes central for deployment.
To address this, optimized hardware accelerators and specialized software must exploit event sparsity to cut computation while preserving accuracy (\fig \ref{hardware_acc}, \fig \ref{hardwar_software_acc}), enabling real-time event-based vision on mobile agents for advanced autonomous applications.

\vspace{-0.2cm}
\subsection{Hardware acceleration}
% Principle of FPGA
% Challenge: CPUs are unsuitable for real-time processing 
%   FPGA
%   GPU
% Implementation
\textbf{Motivation.}
Event-based vision transforms visual processing by generating data only from scene changes, achieving sparsity, high temporal resolution, and low latency, which is ideal for real-time, power-limited applications. Yet, conventional processors built for dense synchronous data struggle with sparse, asynchronous streams, making specialized hardware essential. 
Hardware accelerators bridge raw events to high-level tasks like detection, tracking, and reconstruction by exploiting sparsity and parallelism to cut redundancy, streamline data flow, and support massive parallel processing. 
This enables precise, low-latency, and energy-efficient event-based systems ready for practical deployment.

\textbf{Challenge}.
% In event-based vision systems deployed on mobile platforms, achieving computational efficiency, low power consumption, and low-latency processing is paramount.
Traditional CPUs, constrained by sequential execution, struggle to process sparse, asynchronous event streams in real time, often failing to meet the low-latency, high-throughput demands of tasks like detection and tracking on resource-limited platforms. 
GPUs, though powerful for dense parallel workloads, are optimized for structured image data; when applied to irregular event streams, they suffer from poor resource utilization, higher latency, and increased power draw.
In contrast, FPGAs offer a reconfigurable, massively parallel architecture that can be tailored to the unique characteristics of event data. 
Through custom dataflow designs, sparse convolutions, and asynchronous pipelines, they deliver low-latency, energy-efficient performance without relying on dense matrix operations. 
Their fine-grained hardware control and inherent parallelism make them especially well-suited for mobile and embedded platforms, where efficiency and performance must coexist.

\begin{wrapfigure}{r}{0.5\textwidth} % r 表示图在右边，l 表示图在左边
    \centering
    \vspace{-0.3cm}
    \begin{overpic}[width=1\linewidth,percent]{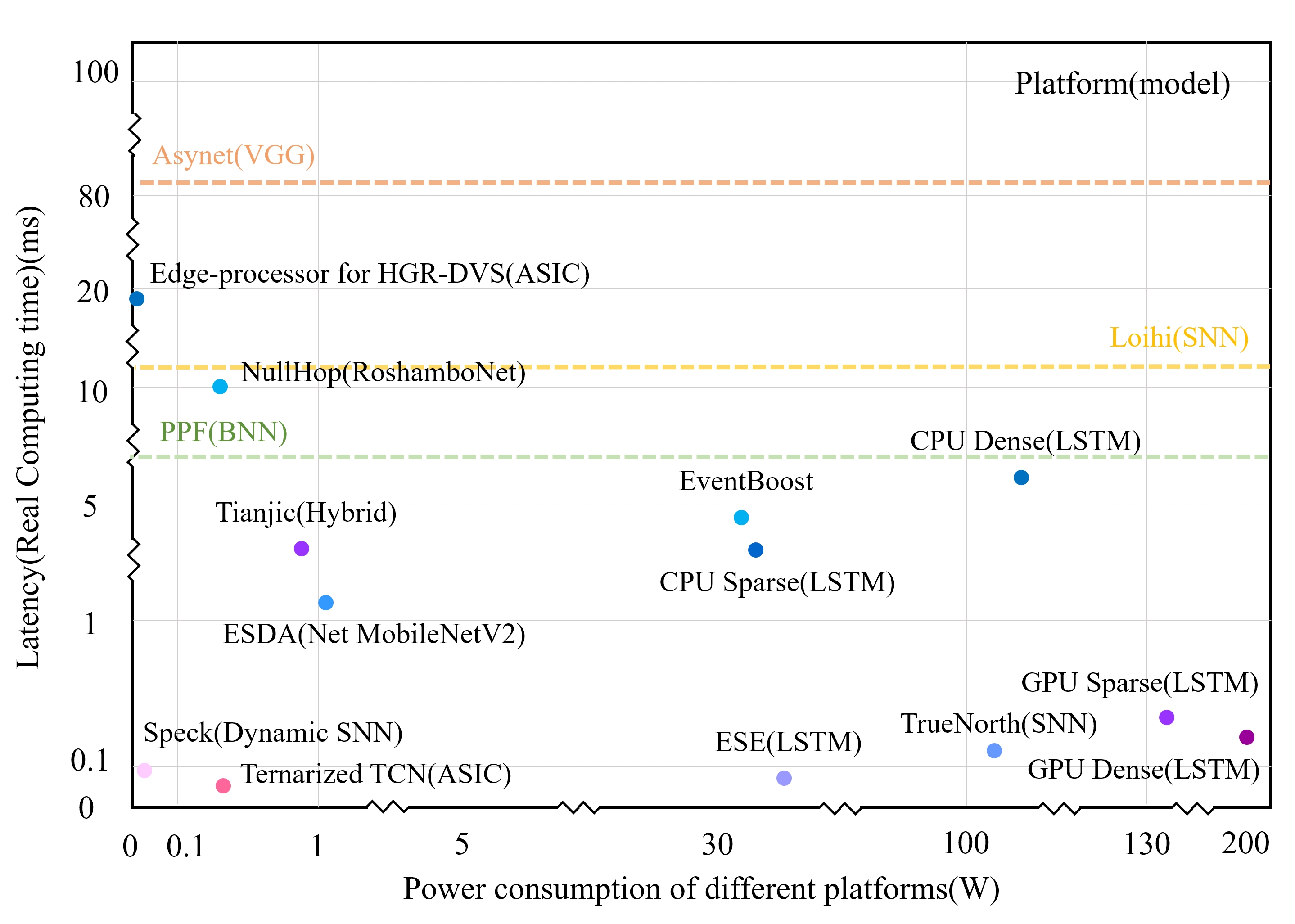}
        \put(24.5,58.5){\scriptsize \cite{hasssan2024spiking}} % VGG
        \put(45.7,49.5){\scriptsize \cite{Fu2024A}} % HGR-DVS
        \put(40.6,42){\scriptsize \cite{aimar2018nullhop}} % NullHop
        \put(22.5,37.5){\scriptsize \cite{cladera2020device}} % BNN
        \put(31,31){\scriptsize \cite{Pei2019Towards}} % Tianjic
        \put(40.6,21.7){\scriptsize \cite{Gao2024A}} % ESDA
        \put(31,14){\scriptsize \cite{Yao2024Spike-based}} % Speck
        \put(40,10.6){\scriptsize \cite{rutishauser2022ternarized}} % TCN
    
        \put(63,33.5){\scriptsize \cite{Cao2024EventBoost}} % Eventboost
        \put(78,44.5){\scriptsize \cite{8259423}} % Loihi
        \put(67,17){\scriptsize \cite{akopyan2015truenorth}} % TrueNorth
        \put(61.5,10.5){\scriptsize \cite{han2017ese}} % ESE
        \put(87.5,36.5){\scriptsize \cite{han2017ese}} % CPU Dense
        \put(69,25.5){\scriptsize \cite{han2017ese}} % CPU Sparse
        \put(89,20){\scriptsize \cite{han2017ese}} % GPU Sparse
        \put(74,10.5){\scriptsize \cite{han2017ese}} % GPU Dense
        
    \end{overpic}
    \vspace{-0.8cm}
    \caption{Power consumption of different platforms($W$) vs. Latency(Real computing time)($ms$) }
    \label{hardware_acc}
    \vspace{-0.5cm}
\end{wrapfigure}

\begin{figure*}[t]   
    \centering
    \includegraphics[width=0.88\linewidth]{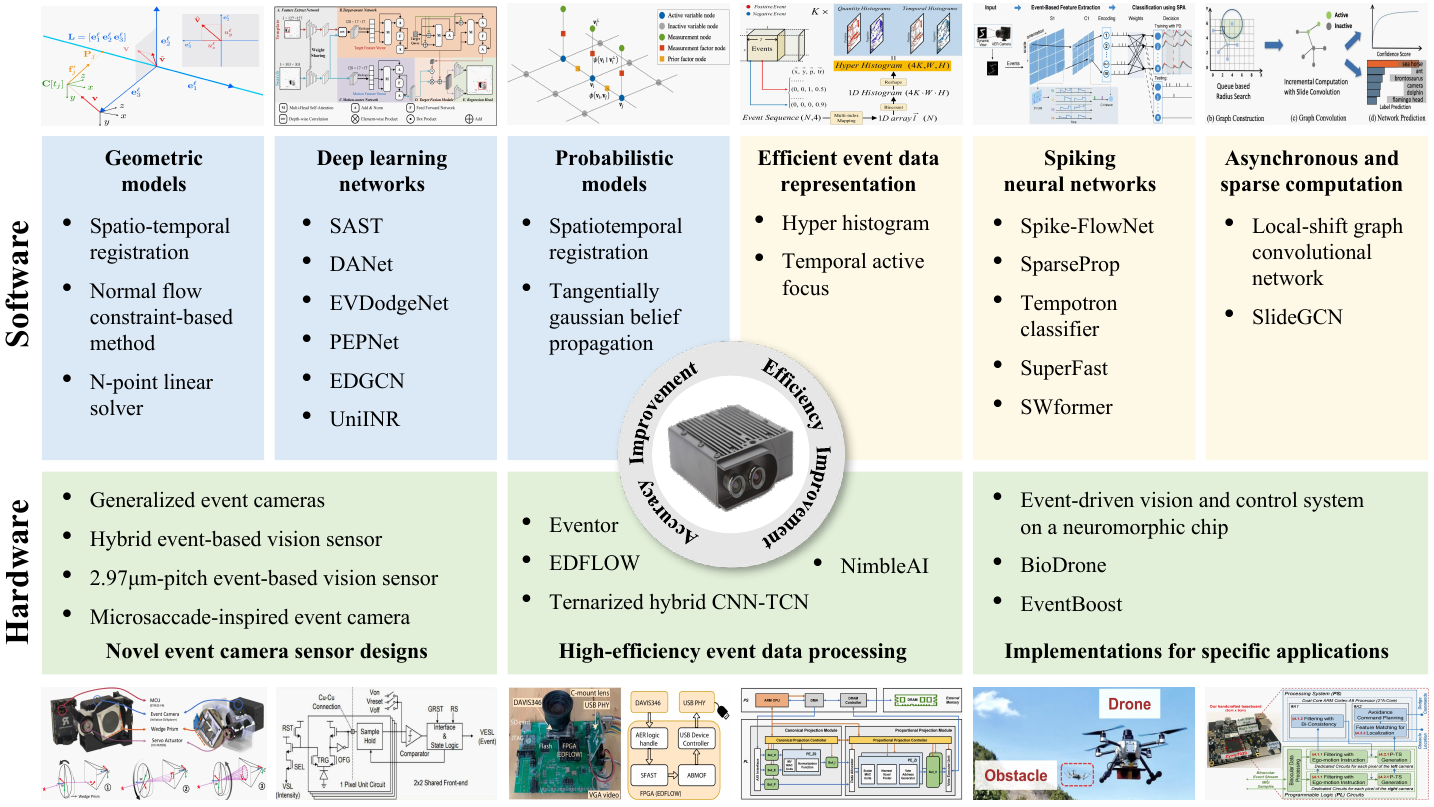}
    \vspace{-0.2cm}
    \caption{Specific hardware and software design for acceleration.}
    \label{hardwar_software_acc}
    \vspace{-0.6cm}
\end{figure*}

\textbf{Literature review}.
% \notice{more citation?}
Recent research on hardware acceleration for event-based vision can be broadly categorized into three main approaches: neuromorphic computing, event-driven deep neural network (DNN) acceleration, and hardware optimization techniques aimed at enhancing efficiency and reducing power consumption (\fig \ref{hardware_acc}, \fig \ref{hardwar_software_acc}).

\textit{Neuromorphic computing for event-based vision}.
Neuromorphic computing, inspired by the brain’s dynamic processing, mimics biological neurons and synapses to enable efficient event-driven computation.
SNNs typically run on custom neuromorphic chips for low-power, high-efficiency processing.
Notable devices include BrainScales \cite{5536970}, Spinnaker \cite{6515159}, Neurogrid \cite{6805187}, TrueNorth \cite{doi:10.1126/science.1254642}, Darwin \cite{MA201743}, and more recently, Loihi \cite{8259423}, Tianjic \cite{Pei2019Towards}, and Speck \cite{Yao2024Spike-based}.
A key challenge is improving energy efficiency via high-level brain-inspired mechanisms.
Among these, the asynchronous chip Speck \cite{Yao2024Spike-based} stands out as a sensing-computing SoC that fully leverages sparse, event-driven processing.
Operating at ultra-low power (0.70 mW in real-time), Speck demonstrates neuromorphic computing’s promise for power-constrained mobile and edge systems.

% Neuromorphic computing draws inspiration from the brain’s dynamic processing capabilities, mimicking biological neurons and synapses to enable efficient event-driven computation. 
% Spiking Neural Networks (SNNs) are typically deployed on custom neuromorphic chips to achieve low-power, high-efficiency processing. 
% Notable examples of such biomimetic computing devices include BrainScales \cite{5536970}, Spinnaker \cite{6515159}, Neurogrid \cite{6805187}, TrueNorth \cite{doi:10.1126/science.1254642}, Darwin \cite{MA201743}, and, more recently, Loihi \cite{8259423}, Tianjic \cite{Pei2019Towards}, and Speck \cite{Yao2024Spike-based}.
% A fundamental challenge in neuromorphic computing is how to further enhance energy efficiency through high-level brain-inspired mechanisms. 
% Among these efforts, the asynchronous chip "Speck" \cite{Yao2024Spike-based} stands out as a sensing-computing system-on-chip (SoC) that fully exploits event-driven processing with sparse and dynamic computation. 
% Speck operates at ultra-low power (0.70$mW$ in real-time applications), demonstrating the viability of neuromorphic computing for power-constrained mobile and edge systems.

\textit{Event-driven DNN acceleration}.
Neuromorphic computing provides a bio-inspired paradigm for event-driven processing, while another crucial research direction focuses on hardware acceleration tailored to event-based deep learning models. Unlike conventional GPUs, which fail to fully exploit the sparsity of event data, specialized architectures have been developed to achieve higher efficiency.
$(i)$ Sparse dataflow architectures: The combinable dynamic sparse dataflow architecture (ESDA) \cite{Gao2024A} realizes a configurable sparse dataflow model on FPGAs. By employing a unified sparse token–feature interface to interconnect parameterizable network modules, ESDA reduces both latency and power consumption during event-based DNN inference, making it well-suited for edge deployment.
$(ii)$ Optimized fusion for event-based vision: EventBoost \cite{Cao2024EventBoost} accelerates event–image fusion through a dedicated hardware accelerator on the Zynq SoC platform. By optimizing fusion workloads in real time, EventBoost mitigates inefficiencies typical of CPU- and GPU-based processing, significantly boosting performance for event-driven visual tasks.

\textit{Hardware optimization for efficient event processing.}
Beyond neuromorphic computing and event-driven DNNs, hardware acceleration is essential for unlocking the full potential of event-based vision, particularly in power-constrained and efficiency-critical scenarios. One optimization strategy aims to reduce power consumption directly at the sensor interface. For example, \cite{RUTISHAUSER2023717} proposes a dedicated on-chip DVS interface that aggregates asynchronous event streams into ternary event frames, substantially lowering the energy demands of subsequent processing stages.

In addition to such front-end innovations, a variety of FPGA-based accelerators have been explored to enhance efficiency across the entire event-vision pipeline by exploiting the inherent parallelism of event data. For instance, \cite{Fu2024A} presents a reconfigurable processing element architecture that integrates both a median filter core and an AI accelerator core for CNN inference within a system-on-chip (SoC). Leveraging Reconfigurable Multiple Constant Multiplication (RMCM) for efficient resource sharing, this design achieves an energy consumption of only 593.4 nJ per inference under 65 nm technology, significantly reducing computational cost. Similarly, \cite{9727106} develops a complete event-driven optical flow camera system with FPGA-based acceleration for key modules such as event-driven corner detection and adaptive block matching optical flow. Expanding toward heterogeneous architectures, \cite{li2022eventorefficienteventbasedmonocular} demonstrates an FPGA/ARM platform for Event-based Monocular Multi-View Stereo (EMVS), where algorithm restructuring and mixed-precision quantization boost throughput while minimizing memory footprint. Finally, \cite{Cao2024EventBoost} proposes a Zynq SoC–based event–vision fusion accelerator that employs hardware–software co-design to distribute tasks between FPGA and CPU, maximizing parallelism and efficiency in computationally demanding fusion algorithms.

\textit{Application-specific hardware acceleration design.}
The development of specialized hardware architectures tailored to accelerate specific tasks has become a central focus in event-based vision research. By optimizing hardware resources for domain applications, these solutions achieve substantial performance gains. For instance, \cite{Fu2024A} introduces an energy-efficient processor-on-chip for hand gesture recognition.
A key driver is aerial robotics, where stringent constraints on size, weight, and power demand highly optimized systems. \cite{xu2023taming} presents FPGA-based hardware–software co-designs that exploit FPGA reconfigurability to implement efficient pipelines for drone tasks such as navigation and obstacle avoidance. Complementarily, \cite{9560881} demonstrates a neuromorphic approach using Intel Loihi, where event data feed directly into SNNs for end-to-end drone control, achieving high responsiveness with low power.
These directions highlight the synergy between specialized hardware and event-based vision. As edge applications—such as autonomous driving, robotics, and AR—demand low-power, real-time performance, domain-optimized accelerators will be pivotal for practical deployment.

\grs{
A key consideration in hardware acceleration is trade-off between accuracy and computational latency. Hardware-specific accelerators often reduce latency substantially by exploiting task-specific parallelism and sparse dataflow, but this comes at the cost of reduced architectural flexibility and higher development overhead. Recent case studies illustrate this balance clearly. For instance, EventBoost \cite{Cao2024EventBoost} employs a software-hardware co-design on a Zynq SoC to accelerate event-visual fusion for drone localization, achieving a 24.33\% improvement in accuracy compared to state-of-the-art systems while maintaining only 30 ms end-to-end latency, thereby meeting real-time constraints on resource-constrained agents. 
Similarly, BioDrone \cite{zhao2024biodrone} integrates an FPGA-based processing pipeline for autonomous drone navigation. Experimental results show that FPGA accelerator reduces per-frame processing latency from nearly 20 ms on CPU to 2.2 ms, a nearly 10$\times$ speed-up, while sustaining almost identical perception accuracy (within 1–2\% deviation) compared to CPU baselines. These studies demonstrate fundamental trade-off in hardware-specific acceleration: latency can be drastically reduced without significant accuracy loss, but achieving this requires tight co-design between algorithms and hardware. As a result, while FPGA and ASIC solutions provide a practical pathway toward real-time, power-efficient event-based vision deployment, their specialized nature may limit applicability across tasks.
}

\vspace{-0.2cm}
\subsection{Software acceleration}
% NeRF
% Bio-inspired algorithm
\textbf{Motivation}.
Deploying event cameras on mobile agents requires software acceleration to handle sparse, asynchronous data in real time under strict resource limits.
By transforming raw events into structured representations through clustering, compression, or sparse matrix operations, software acceleration streamlines denoising, filtering, feature extraction, and inference. Exploiting sparsity, reducing redundancy, and applying techniques like adaptive sampling and dynamic memory management are key to achieving energy-efficient, real-world mobile applications.

\textbf{Challenge}.
Effective software acceleration for event cameras on mobile agents faces several challenges. First, sparse and asynchronous nature of event data makes dense-input algorithms inefficient, requiring specialized designs. Second, integration with deep learning frameworks is nontrivial, as most models are tailored for frame-based inputs. Third, limited on-device resources demand optimized memory management and throughput to balance speed and power. Finally, current benchmarks overlook unique characteristics of event data, calling for dedicated evaluation metrics.

\textbf{Literature review.}
% \notice{more citation?}
% In the field of event-based vision systems, software acceleration is crucial for enhancing the efficiency and performance of the event processing pipeline. 
% By optimizing specific stages of event handling, researchers can significantly improve computational performance, reduce latency, and optimize resource utilization. 
% Software acceleration can be applied to various stages of the pipeline, including event sampling, preprocessing, feature extraction, event analysis, decision making, and result output.
% Each stage of the pipeline faces unique challenges and several software acceleration methods have been proposed for these stages:
In event-based vision systems, software acceleration is essential for improving the efficiency and performance of the event processing pipeline.
By optimizing key stages of event handling, researchers can significantly enhance computational speed, reduce latency, and better utilize limited resources.
Software acceleration techniques target various pipeline stages, such as event sampling, preprocessing, feature extraction, and event analysis, as shown in \fig \ref{hardwar_software_acc}.
Each stage presents distinct challenges, and numerous specialized acceleration methods have been developed to address them:

 \textit{Event sampling.}
Event cameras capture raw data with precise timestamps and spatial locations. Unlike uniform sampling, adaptive sampling prioritizes significant scene changes to improve data efficiency. The key challenge is balancing temporal resolution with sparsity to prevent information loss or redundancy. \cite{wang2025eas} addresses this by introducing a recurrent convolutional SNN-based adaptive sampling module that dynamically adjusts rates based on spatio-temporal event patterns, enhancing overall efficiency.

\textit{Event pre-processing.} Events need to be denoised, filtered, and formatted to improve subsequent processing. 
Common challenges include false positives and inefficient filtering. 
Enhanced filtering algorithms, such as adaptive median filters, can retain important features while reducing computational overhead.
A lightweight, hardware-friendly neural network architecture, 2-D CNN, is introduced in \cite{Fu2024A} for DVS gesture recognition, using a customized median filter to enhance signal-to-noise ratio and reduce hardware complexity.

\textit{Feature extraction.} Useful features, including spatial, temporal, and frequency features, are extracted from preprocessed data. 
Traditional techniques may not effectively capture event data's unique properties. 
Software acceleration, particularly through deep learning models tailored for sparse data, can optimize feature extraction while reducing processing time.
The FARSE-CNN model, proposed in \cite{Santambrogio2024FARSE-CNN:}, integrates sparse convolutional and asynchronous LSTM modules for efficient event data processing. 
\cite{Fang2024Spiking} introduces SWformer, an attention-free architecture leveraging sparse wavelet transforms to capture high-frequency patterns, resulting in improved energy efficiency and performance.

\textit{Event analysis.} This step involves pattern recognition, classification, or regression, commonly for tasks like object detection or tracking. 
The sparsity of event data can lead to high computational costs. Optimized CNNs and SNNs can enhance the efficiency and effectiveness of analysis.
A fast linear solver for camera motion restoration, developed in \cite{Ren2024Motion}, addresses geometric problems and adapts to sudden motion changes, providing a robust solution for event data.

Moreover, existing methods improve event processing by optimizing event representations, integrating geometric and probabilistic models, and employing deep learning techniques.
They also leverage the asynchronous and sparse characteristics of event data to accelerate computation without sacrificing accuracy.
% enabling real-time performance even on resource-constrained platforms.

\textit{Efficient event data representation}.
Efficient event data representation methods convert sparse, asynchronous event streams into structured formats for effective processing with reduced overhead. Recent approaches include event stacking \cite{8954323}, Temporal Activity Focus (TAF) \cite{10109007}, and Hyper Histogram (HH) \cite{Peng2023BetterAF}. TAF adaptively adjusts time window length and resolution based on spatial and polarity cues, enhancing flexibility. HH builds multiple histograms from event polarity and temporal statistics, integrating them into 3D tensors to preserve fine-grained details.

\why{
\textit{Geometric models-based methods}.
Geometric models leverage spatial relationships within scenes to improve the performance of event-based visual algorithms. By integrating geometric constraints from event data, these models reduce error accumulation, enhancing both accuracy and computational efficiency.
Recent progress has been notable in event-based visual odometry \cite{Liu2021SpatiotemporalRF}, motion estimation \cite{Ren2024Motion, 10656961, wu2024motion, zhou2024resflow, zhou2025spatially}, and time-to-collision estimation \cite{li2024eventaidedtimetocollisionestimationautonomous}.}
% advancing the precision of these critical tasks.

% This paper integrates video diffusion models with event camera data, using alignment via reinforcement learning to predict future object motion with high temporal precision.

\textit{Probabilistic models-based methods}.
Probabilistic models offer a principled way to represent data and quantify uncertainty, making them well-suited for the noisy, asynchronous nature of event streams. By leveraging probabilistic inference and optimization, they improve both robustness and accuracy in event-based vision. Recent studies demonstrate their effectiveness in object classification \cite{liu2020effectiveaerobjectclassification}, where class likelihoods are directly maximized from event data, and in optical flow estimation \cite{10204349}, where noise modeling and belief propagation enhance precision.
\textit{Deep learning-based methods}.
Deep learning has greatly advanced event-based vision by extracting complex spatio-temporal patterns for accurate prediction and decision-making.
Recent efforts focus on efficiency, such as lightweight networks and sparse Transformers \cite{peng2024sceneadaptivesparsetransformer} for object detection, and deep models for dynamic obstacle avoidance \cite{sanket2020evdodgenetdeepdynamicobstacle} and tracking \cite{10299598}. Advanced methods like GCNs and cross-representation distillation \cite{deng2023dynamicgraphcnncrossrepresentation} further improve scene understanding, while point cloud-based networks enhance pose relocalization \cite{ren2024simpleeffectivepointbasednetwork}, and unified implicit neural representations support rolling shutter image restoration \cite{lu2024uniinreventguidedunifiedrolling}. 
% Complementing these, SNNs provide energy-efficient, event-driven computation, showing strong performance in object classification \cite{liu2020effectiveaerobjectclassification}, optical flow estimation \cite{lee2020spikeflowneteventbasedopticalflow}, high-speed reconstruction \cite{9962797}, and recurrent simulation \cite{engelken2023sparsepropefficienteventbasedsimulation}.

% and novel designs like the spiking wavelet transformer \cite{fang2024spikingwavelettransformer}.

\grs{
% A particularly promising direction involves lightweight neural architectures specifically designed for event cameras, which exploit the sparse and asynchronous nature of event data to achieve efficient processing. These models move beyond simply compressing existing networks, instead co-designing architectures and learning principles with the unique properties of event streams.
% IDNet\cite{wu2024lightweight} replaces costly 4D correlation volumes with an iterative motion-compensation loop, where a lightweight ConvGRU progressively refines residual flow on deblurred event traces, achieving accuracy with drastically fewer parameters and memory.
% FARSE-CNN\cite{santambrogio2024farse} introduces a fully asynchronous, recurrent sparse-CNN that eschews dense event accumulation; through spatio-temporal compression modules, it learns hierarchical features directly from sparse event streams, attaining state-of-the-art efficiency with minimal computational load.
% Ultralight Polarity-Split SNN\cite{xu2025ultralight} adopts polarity-split encoding and a learnable spatio-temporal loss to achieve event-stream super-resolution with minimal model size and latency, enabling efficient embedding in-event camera systems.
A promising direction lies in lightweight neural architectures tailored to the sparse and asynchronous nature of event streams. Rather than compressing conventional models, these approaches co-design network structures and learning principles with the sensing modality itself.
IDNet \cite{wu2024lightweighteventbasedopticalflow} replaces expensive 4D correlation volumes with an iterative motion-compensation loop, where a lightweight ConvGRU progressively refines residual flow, achieving competitive accuracy with far fewer parameters and memory.
FARSE-CNN \cite{Santambrogio2024FARSE-CNN:} proposes a fully asynchronous recurrent sparse-CNN, incorporating spatio-temporal compression modules to learn hierarchical features directly from events, yielding state-of-the-art efficiency at low computational cost.
Ultralight Polarity-Split SNN \cite{xu2025ultralight} leverages polarity-split encoding and a learnable spatio-temporal loss for event-stream super-resolution, providing low-latency inference with minimal model size and enabling on-sensor deployment.
% Spike-FlowNet\cite{lee2020spike} integrates spiking neural networks with event streams, exploiting spike timing to encode motion and replacing dense floating-point operations with sparse spike transmissions, yielding substantial energy efficiency. % 2020
% e-TLD\cite{ramesh2020tld} adapts the TLD paradigm to event data via sparse keypoint sampling and probabilistic temporal coherence, updating the model only when necessary to minimize computation while preserving tracking reliability. % 2020
}

\textit{Asynchronous and sparse computation-based acceleration.}
Asynchronous and sparse computation aligns with event cameras’ data characteristics by updating only when changes occur, selectively processing active regions. This reduces computational cost and boosts efficiency. Typical approaches focus on pixel brightness changes or region-specific features. Recent advances include sparse convolutional networks for asynchronous streams \cite{messikommer2020eventbasedasynchronoussparseconvolutional}, graph-based frameworks \cite{9710541}, and local shift operations for optimized event handling \cite{10.1609/aaai.v37i2.25336}.

\vspace{-0.2cm}
\section{Application: Mobile agent-based task } \label{6}
\begin{wrapfigure}{r}{0.6\textwidth} % r 表示图在右边，l 表示图在左边
    \centering
    \vspace{-0.5cm}
    \includegraphics[width=1\linewidth]{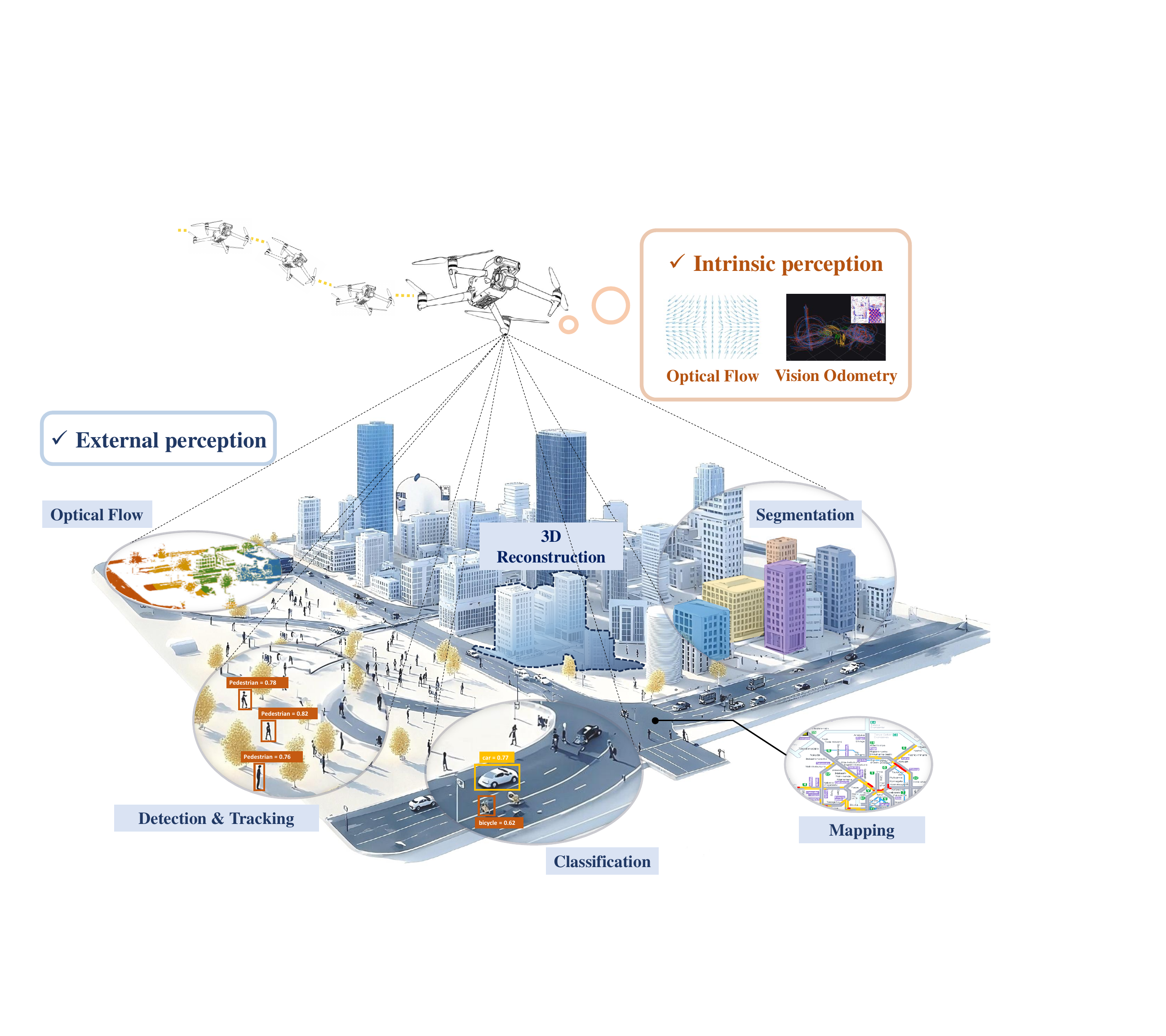}
    \vspace{-0.8cm}
    \caption{Event camera-based perception tasks on mobile agents}
    \label{application}
    \vspace{-0.3cm}
\end{wrapfigure}

% \\ \role{pengtao}
% \notice{finish the citation.}
% \notice{Figures on relationship of different applications}
This section explores the various tasks associated with mobile agents, emphasizing the use of event cameras and other perception technologies to improve performance in dynamic environments. 
These tasks are categorized into intrinsic perception, external perception, and event-based SLAM, each offering distinct applications and advantages (\fig \ref{application}).
% This section explores the various tasks associated with mobile platforms, focusing on the utilization of event cameras and other sensing technologies to enhance performance in dynamic environments. The tasks are categorized into intrinsic sensing, external sensing, and event-based SLAM, each with its specific applications and advantages.

% \begin{figure*}[t]   
%     \centering
%     \includegraphics[width=0.58\linewidth]{Figs/redraw_ruishan_11.pdf}
%     \vspace{-0.4cm}
%     \caption{Event camera-based sensing tasks on mobile platforms}
%     \label{application}
%     \vspace{-0.6cm}
% \end{figure*}

\vspace{-0.2cm}
\subsection{Intrinsic perception tasks}
% 对无人机内部状态的感知，可以分成几个任务写，比如自身位置估计这种，咱们之前总结过,
% Intrinsic sensing involves the use of internal sensors on mobile platforms to collect data about the immediate environment.
% This encompasses visual odometry, and optical flow, with event cameras providing high temporal resolution and low-latency data that enhance accuracy and efficiency.

% Intrinsic sensing uses onboard sensors to capture the immediate environment, encompassing visual odometry and optical flow, with event cameras offering high temporal resolution and low latency to improve accuracy and efficiency.

\textbf{Vision odometry}.
% Event-based visual odometry leverages event stream to estimate motion with greater precision and lower computational cost than traditional frame-based methods \cite{10254473}. 
% Typical approaches extract pose or motion parameters directly from event streams. 
% For example, \cite{7805257} proposes angular velocity estimation via contrast maximization using event edges, while \cite{9706311} improves attitude estimation in high-rotation scenarios through enhanced aggregation functions. 
% The DEVO system \cite{zuo2022devodeptheventcameravisual} integrates depth data for robust odometry in challenging conditions.
% Moreover, \cite{10107754} introduces the first event-based stereo visual-inertial odometry system combining stereo event cameras, standard cameras, and IMUs, using spatiotemporal correlation and motion compensation to improve accuracy in dynamic scenes.
Event-based visual odometry exploits asynchronous event streams to estimate motion with higher precision and lower computational cost than frame-based methods \cite{10254473}. 
Early work focused on direct motion extraction, such as angular velocity estimation via contrast maximization on event edges \cite{7805257} and improved attitude estimation under high rotation using enhanced aggregation functions \cite{9706311}. 
More advanced systems integrate complementary modalities: \cite{zuo2022devodeptheventcameravisual} incorporates depth for robust odometry in challenging conditions, while first event-based stereo visual-inertial system \cite{10107754} fuses stereo events, standard frames, and IMUs through spatiotemporal correlation and motion compensation, significantly boosting accuracy in dynamic scenes.
\why{
Although these methods suggest proof-of-concept readiness, their large-scale deployment in unstructured real-world environments remains limited.
Advanced methods have taken a step forward by attempting GPS-denied navigation for unmanned systems, comparing real-time terrain fingerprints generated from event camera outputs with pre-stored fingerprints derived from satellite imagery \cite{Neuromorphic}.
}

\textbf{Optical flow}.
Event cameras are highly suitable for optical flow estimation, supporting real-time processing in HDR environments critical for motion tracking. Existing methods fall into two main categories:
\textit{(i) Deep learning-based approaches}, which leverage neural networks to infer flow from event streams—for example, a hierarchical SNN for global motion perception \cite{8660483}, though they often require event-to-frame preprocessing.
\textit{(ii) Traditional computer vision approaches}, which exploit the spatio-temporal structure of event data. These include normal flow estimation fused with IMU data for high-dynamic velocity estimation \cite{lu2024eventbasedvisualinertialvelometer} and contrast maximization methods with tailored reward functions \cite{Stoffregen2019EventCC}.
\why{
Event-based optical flow has shown promise in drone status estimation, and deep learning methods perform well in lab settings; however, their robustness to sensor noise and illumination variability in real-world deployments remains an open research problem.
}

\vspace{-0.2cm}
\subsection{External perception tasks}
% \vspace{-0.15cm}
% 对无人机外部状态的感知，可以分成几个任务写，比如障碍物位姿估计这种，咱们之前总结过
% External sensing leverages sensors to capture the broader environment of a mobile platform, supporting tasks such as mapping, detection and tracking, obstacle avoidance, optical flow, classification, 3D reconstruction, and segmentation—all benefiting from the high temporal resolution of event cameras.

% External sensing involves using sensors to perceive the broader environment around a mobile platform. Typical tasks include mapping, object detection and tracking, obstacle avoidance, optical flow estimation, classification, 3D reconstruction, and segmentation. These applications benefit from the high temporal resolution of event cameras.

% External sensing refers to the use of sensors to gather data about the broader environment surrounding the mobile platform. 
% This encompasses a range of tasks, including mapping, object detection and tracking, obstacle avoidance, optical flow, classification, 3D reconstruction, and segmentation, all of which benefit from the high temporal resolution and low latency provided by event cameras.

\textbf{Mapping}.
% Event cameras enhance 3D map construction accuracy and efficiency by capturing rapid environmental changes. 
% Feature-based methods are common, extracting point or line features from scenes.
% For example, \cite{9361194} detects straight lines in railway settings, combining odometry for infrastructure mapping. 
% However, these methods rely on feature stability and repeatability, which can be challenging in sparse or occluded environments.
% To overcome such limitations, recent approaches explore alternative strategies. 
% BeNeRF \cite{li2024benerfneuralradiancefields} reconstructs neural radiance fields (NeRF) using a single blurred image paired with an event stream, removing the dependence on extensive camera pose data and enabling learned scene representations for mapping. 
% AsynHDR \cite{wu2024eventbasedasynchronoushdrimaging} employs LCD modulation to achieve HDR imaging, enriching scene information for improved 3D reconstruction.
Event cameras improve 3D map construction by capturing rapid environmental changes with high temporal resolution. 
Feature-based methods, such as extracting points or lines, remain common. 
For instance, \cite{9361194} detects straight lines in railway settings and combines them with odometry for infrastructure mapping. However, these methods depend on stable, repeatable features, which may be unreliable in sparse or occluded scenes. To address this, recent works adopt alternative strategies: BeNeRF \cite{li2024benerfneuralradiancefields} reconstructs NeRFs from a single blurred image and an event stream, reducing reliance on extensive pose data, while AsynHDR \cite{wu2024eventbasedasynchronoushdrimaging} leverages LCD modulation for HDR imaging, enriching scene information for more accurate 3D reconstruction.
\why{
In summary, feature-based and neural approaches show promise for 3D reconstruction in dynamic environments, but many rely on assumptions about feature stability or controlled settings, indicating they are largely at the proof-of-concept stage rather than fully deployment-ready.
}

\textbf{Object detection \& tracking}.
The high temporal resolution and low latency of event cameras make them particularly well-suited for precise object detection and tracking in dynamic environments \cite{gehrig2024low, gehrig2023recurrentvisiontransformersobject}. Existing methods can be broadly categorized into two groups: 
% event stream-based detection and tracking, and fusion-based methods that incorporate additional data sources such as frame images or IMU measurements.
\textit{(i) Event stream-based detection and tracking methods.} 
These approaches primarily leverage the spatio-temporal characteristics of event data \cite{ding2025hawkeye, chen2025count}. 
For example, \cite{8088365} proposes a stereo event-based tracking algorithm that addresses occlusion by combining 3D reconstruction with cluster tracking. 
% EventCap \cite{9157340} and EventEgo3D \cite{millerdurai2024eventego3d3dhumanmotion} demonstrate monocular event-based 3D human motion capture.
\why{
\cite{zhu2022learning} proposes an end-to-end event cloud-based object tracking framework using density-insensitive key-event sampling, graph-based embedding, and motion-aware likelihood prediction.
}
% showcasing the potential of event cameras in pose estimation and tracking. 
% In \cite{9197341}, corner tracking is used to filter noise and identify events from moving intruders, enabling robust object tracking. 
EVPropNet \cite{sanket2021evpropnetdetectingdronesfinding} tracks drone propellers using event data, while EDOPT \cite{10611511} performs six-degree-of-freedom object pose tracking solely with event cameras.
\textit{(ii) Fusion-based detection and tracking methods.} 
These methods combine event data with additional sensors to enhance accuracy and robustness. 
% In \cite{8793924} and \cite{10195232}, frame images are fused with event streams to improve target detection. 
\why{
\cite{zhu2025modeling} proposes a two-stage gaze estimation framework using event and frame data with anchor-based state shifts and denoising distillation.
\cite{zhu2023cross} explores object tracking from RGB and event data by leveraging a pre-trained ViT with mask modeling and orthogonal high-rank loss to enhance inter-modal token interaction.
\cite{chen2024crossei} proposes a motion-adaptive event sampling and bidirectional-enhanced fusion framework to align event and image data for more accurate object tracking.
}
High-frequency drone localization using mmWave radar and event streams is demonstrated in \cite{wang2025ultra}, and \cite{luo2024eventtracker} integrates depth camera data with events for obstacle tracking.
\why{
In summary, some event-only methods and fusion-based approaches demonstrate impressive performance in dynamic and high-speed conditions. 
Fusion-based systems that combine RGB, depth, or radar data appear closest to real-world deployment, especially in drone navigation and obstacle tracking.
In the industry, Meituan has explored combining event cameras with mmWave radar for drone localization \cite{wang2025ultra}. 
Tobii, a global leader in eye-tracking, together with Meta, has investigated event-camera-based gaze tracking \cite{tobii, stoffregen2022event}. 
Meanwhile, SAAZ Micro Inc. and Neurobus have applied event cameras to drone detection \cite{saaz, neurobus}.
}

\textbf{Optical flow}.
% Event cameras excel in external optical flow estimation, enabling accurate motion tracking in high-speed scenarios \cite{kosta2023adaptivespikeneteventbasedopticalflow, liu2023tmatemporalmotionaggregation, 10655490, ponghiran2023eventbasedtemporallydenseoptical, 10204349}. 
% Unlike internal perception, which estimates the camera's own motion, external optical flow focuses on understanding the movement of objects in the environment. 
% Typical applications include object capture and robotic control.
% Recent works explore diverse strategies to enhance optical flow estimation with event cameras. RPEFlow \cite{10376957} improves accuracy by fusing event data with RGB images and point clouds through cross-modal attention. 
% Contrast maximization (CM)-based methods \cite{Shiba_2022, 10517639} offer a principled alternative, achieving SOTA results with innovations such as multi-reference focus loss and time-aware flow modeling. 
% TEGBP \cite{10204349} further proposes an efficient approach for estimating dense optical flow from sparse event data. 
% These advances highlight the growing potential of event cameras for robust external motion analysis.
Event cameras excel in external optical flow estimation, enabling robust motion tracking of dynamic objects in high-speed scenarios \cite{kosta2023adaptivespikeneteventbasedopticalflow, ponghiran2023eventbasedtemporallydenseoptical}. Unlike internal perception, which focuses on self-motion estimation, external optical flow targets the movement of surrounding objects, supporting applications such as object capture and robotic control. Existing approaches can be broadly categorized into multimodal fusion, contrast maximization, and event-specific dense flow estimation. For instance, RPEFlow \cite{10376957} enhances accuracy by fusing events with RGB images and point clouds via cross-modal attention. Contrast maximization (CM)-based methods \cite{Shiba_2022, 10517639} provide a principled formulation and achieve state-of-the-art results with innovations like multi-reference focus loss and time-aware flow modeling. Complementarily, TEGBP \cite{10204349} introduces an efficient framework for deriving dense optical flow directly from sparse event data. These developments highlight the growing versatility of event cameras for external motion analysis across diverse environments.
\why{
Despite these advances, most methods remain at the proof-of-concept stage, validated primarily in controlled laboratory settings. 
It is worth noting that this technology shows significant potential for applications such as fall detection, crowd detection and tracking, and traffic data acquisition \cite{eventiot}.
}

% Event cameras excel in external optical flow calculations, enabling accurate motion tracking in high-speed scenarios \cite{kosta2023adaptivespikeneteventbasedopticalflow},\cite{liu2023tmatemporalmotionaggregation},\cite{10655490},\cite{ponghiran2023eventbasedtemporallydenseoptical},\cite{10204349}. 
% Unlike internal perception, which focuses on the camera's motion, optical flow information here is used to understand and predict the movement of external objects. 
% Typical applications include high-speed object capture and robotic control.

% RPEFlow\cite{10376957} enhance accuracy by combining event data with RGB and point clouds, leveraging cross-modal attention for dynamic scenes.  Contrast Maximization (CM) based methods\cite{Shiba_2022}\cite{10517639}  offer a principled alternative, achieving state-of-the-art performance through innovations like multi-reference focus loss and time-aware flow modeling.  Distinct from these, TEGBP\cite{10204349} provides an efficient, incremental estimation of full optical flow from sparse event data.  Collectively, these advancements demonstrate the diverse and potent strategies emerging to harness event cameras for robust external motion analysis.

\textbf{Classification}.
Event cameras’ ability to capture rapid changes in lighting and motion greatly improves image classification \cite{10377903, zheng2024eventdanceunsupervisedsourcefreecrossmodal}. 
Event-based classification methods leverage subtle motion and illumination variations with specialized models.
For instance, \cite{9337225} uses graph convolutional networks on event-derived graph structures for classification, while \cite{9010397} applies deep learning to recognize individuals from gait patterns. 
Event cameras also aid microscopic object classification using SNNs \cite{neftci2019surrogategradientlearningspiking}, and \cite{9844855} captures facial micro-expressions for emotion recognition, showcasing event data’s sensitivity to subtle cues.
\why{
These methods have been validated on specific datasets.
% for micro-expression recognition and gait analysis.
While promising, their generalization to large-scale, unconstrained environments remains a challenge.
It is worth noting that such approaches hold significant potential for applications in medical diagnostics \cite{medical}.
}

\textbf{3D Reconstruction}.
Event cameras enable precise 3D reconstruction in dynamic scenes by leveraging high temporal resolution and sparse data. 
Real-time monocular reconstruction is shown in \cite{Kim2016RealTime3R}, while \cite{Zhou_2018} used stereo event cameras for semi-dense 3D reconstruction, balancing accuracy and efficiency. Combining structured light with event cameras allowed high-speed 3D scanning with less data redundancy \cite{Huang:21}. 
A polarization-based method in \cite{muglikar2023event} maintains precision even at low event rates. Specific applications include real-time 3D hand gesture estimation \cite{9711380}, 3D human pose and shape estimation \cite{zou2021eventhpeeventbased3dhuman}, and non-rigid object reconstruction handling complex motions \cite{xue2022eventbasednonrigidreconstructioncontours}.
\why{
In summary, real-time monocular and semi-dense stereo reconstructions have been achieved in controlled scenarios. 
Non-rigid object reconstruction and human pose estimation highlight potential, but deployment in complex scenes is still limited.
}

\textbf{Segmentation}.
Event cameras improve image segmentation performance by enabling accurate and efficient segmentation in rapidly changing environments \cite{jing2024hplesshybridpseudolabelingunsupervised, kong2024openesseventbasedsemanticscene}. Segmentation approaches based on event data can be divided into:
\textit{(i) Motion-based segmentation methods} leverage the motion information captured by event cameras to segment scenes. 
For example, \cite{9010722} proposes a motion compensation-based iterative optimization algorithm that segments scenes into independently moving objects, effectively exploiting the event camera’s high sensitivity to motion changes for dynamic target segmentation.
\textit{(ii) Deep learning-based semantic segmentation methods} utilize deep neural networks for feature extraction and semantic understanding. 
\grs{
Existing works in this category explore how to effectively combine event data with RGB frames or adapt pre-trained segmentation models to the event domain. For instance,} CMDA \cite{xia2023cmdacrossmodalitydomainadaptation} exploits HDR of event cameras to complement the limited dynamic range of frame cameras, achieving robust semantic segmentation in challenging nighttime scenes.
\grs{
More recently, SAM-Event-Adapter \cite{yao2024sam} introduces a lightweight adapter to bridge event data with the Segment Anything Model (SAM), enabling zero-shot semantic segmentation and demonstrating strong generalization across diverse event datasets. Similarly, \cite{chen2024segment} proposes a multi-scale feature distillation method to align embeddings from event data with RGB images, further facilitating the adaptation of SAM for robust and universal object segmentation in the event domain.
}
% \why{
% \cite{chen2024segment} proposes a multi-scale feature distillation method to align embeddings from event data with RGB images, enabling the adaptation of the Segment Anything Model (SAM) for robust and universal object segmentation in the event domain.
% }
\why{
Although mostly evaluated on datasets, these methods have strong potential in IoT applications, such as crowd detection and traffic data acquisition, as well as in medical contexts like high-speed particle tracking in microfluidic devices, where accurate object segmentation is essential.
}
\vspace{-0.2cm}
\subsection{Simultaneous Localization and Mapping (SLAM)}
Event-based SLAM leverages the high temporal resolution and low latency of event cameras for robust localization and mapping in dynamic environments \cite{niu2025esvo2, 10655500}. 
A key enhancement is multimodal fusion, combining complementary sensors to overcome individual limitations.
For instance, Ultimate SLAM \cite{8258997} fuses events, frames, and IMU data for HDR and high-speed scenarios, while Implicit Event-RGBD Neural SLAM \cite{10655500} integrates event and RGB-D data to handle motion blur and lighting changes.
Advances in feature representation further boost robustness. 
Line-based SLAM methods \cite{Chamorro2022EventBasedLS} mitigate feature loss in fast motion or low-light using PTAM frameworks, while optimization-based approaches like CMax-SLAM \cite{10474186} apply contrast maximization for precise rotational motion estimation through event-based global bundle adjustment.
\why{
Despite these promising results, most event-based SLAM methods remain at the proof-of-concept stage, validated mainly in controlled or semi-controlled scenarios. 
Only a few methods have been demonstrated in real-world dynamic environments \cite{niu2025esvo2, zhou2021event}, indicating that full deployment readiness is still limited.
}
\vspace{-0.3cm}
\section{Future direction and discussion} \label{8}

% \begin{figure}[htbp]
%     \centering
%         \includegraphics[width=1\columnwidth]{Figs/crop_future_v1.pdf}
%     \vspace{-0.5cm}
%      \caption{Future directions.}
%     \label{future}
%     \vspace{-0.5cm}
% \end{figure} 

% Despite their potential to overcome some limitations of frame-based cameras and enable new applications in previously inaccessible scenarios, 
% Although there has been extensive research on event cameras, they are still in the early stages of development for mobile sensing.
% Significant opportunities remain for advancement in both research and industry, with several trade-offs to consider, such as latency, power consumption, and accuracy. 
\why{
Despite extensive research on event cameras, their application in mobile embodied perception remains in early stages, with significant opportunities for advancement while balancing trade-offs in latency, power consumption, and accuracy.
% Below, we outline key future directions for event-based mobile sensing.
% , as shown in \fig \ref{future}

% \begin{figure}[t]
%     \centering
%         \includegraphics[width=0.8\columnwidth]{Figs/redraw_ruishan_28.pdf}
%     \vspace{-0.3cm}
%      \caption{Future directions.}
%     \label{future}
%     \vspace{-0.5cm}
% \end{figure} 

\noindent \textbf{(1) Improving event cameras using optics devices.}
Event cameras respond only to illumination changes and remain inactive in fully static scenes, limiting continuous perception. 
While some hardware-based solutions exist, they often reduce energy efficiency due to added mechanical components \cite{he2024microsaccade, bao2024temporal}. 
Future work could explore dynamic optical elements, such as electro-optic materials, to induce controlled illumination changes via optical phase arrays, enabling detection in static environments. 
Hybrid optical-electronic systems, combining event cameras with active illumination or computational imaging, may further enhance performance \cite{yu2024eventps, muglikar2023event}. 
Key challenges remain, including designing diverse scanning patterns without blind spots, developing easily implementable non-mechanical illumination devices, and improving event signal quality under high-speed illumination.

\noindent \textbf{(2) Designing neuromorphic hardware to facilitate event processing.}
% Current hardware faces significant challenges in processing event-based data. CPUs suffer from frequent context switching, reducing throughput, while GPUs remain poorly suited for asynchronous, high-frequency events. FPGAs offer parallelism and low latency but lack full end-to-end pipeline optimization. Although dedicated accelerators have been explored, the massive volume of events (e.g., thousands per millisecond) can still overwhelm I/O bandwidth, and burst-induced latency often exceeds worst-case budgets, introducing jitter that destabilizes control loops. Programmability is further limited, as mapping high-level neural models to diverse neuromorphic ISAs is still largely manual and brittle, complicating debugging and optimization.
% Future efforts should focus on specialized neuromorphic hardware explicitly designed for event-driven computing. Architectures based on SNNs or asynchronous pipelines could better exploit event sparsity, while high-throughput interconnects with QoS, priority channels for control-critical data, and traffic shaping are needed to mitigate I/O bottlenecks. To ensure temporal determinism, hard real-time scheduling, bounded-latency NoCs, and deadline-aware routers are essential. Equally important are stable intermediate representations (IRs) and robust compiler toolchains, which can unify high-level model descriptions and improve programmability across heterogeneous neuromorphic platforms.
Current hardware faces challenges in processing event-based data. CPUs suffer from frequent context switching, while GPUs are ill-suited for asynchronous, high-frequency events. FPGAs provide parallelism and low latency but lack end-to-end pipeline optimization. 
Although some efforts have developed dedicated accelerators \cite{Cao2024EventBoost, xu2023taming, Pei2019Towards}, the massive data volume in event stream (e.g., thousands of events in milliseconds) can easily overwhelm I/O bandwidth, and end-to-end latency often violates worst-case budgets under event bursts, causing jitter that disrupts control loops. Moreover, mapping high-level neural models to neuromorphic ISAs remains manual and brittle, limiting programmability and debugging.
Future work should pursue specialized neuromorphic hardware tailored for event-driven computing. Architectures based on SNNs or asynchronous pipelines can better exploit event sparsity, while higher-throughput interconnects with QoS, priority channels for control streams, and traffic shaping can alleviate I/O bottlenecks. Hard real-time scheduling, bounded-latency NoCs, and deadline-aware routers are needed to guarantee temporal determinism. Finally, stable intermediate representations (IRs) and compiler toolchains are essential to improve programmability across diverse neuromorphic platforms.

\noindent \textbf{(3) Leveraging the complementary strengths of event camera and other sensors.}
Event cameras complement frame cameras, radar, and LiDAR by providing high temporal resolution and low latency, which mitigate motion blur and enhance perception in dynamic environments. 
While frame cameras offer texture and color \cite{gehrig2024low, zhao2025urbanvideo}, radar ensures robustness under poor lighting and weather \cite{wang2025ultra}, and LiDAR supplies precise 3D geometry \cite{cui2022dense, zhou2024bring}, their integration with event streams enables improved visual odometry, detection, reconstruction, and SLAM. 
Yet, most existing fusion methods remain loosely coupled and overlook the intrinsic properties of raw multimodal data—for instance, frame images are spatially dense but temporally sparse, event and radar data are spatially sparse yet temporally dense, and LiDAR point clouds are sparse in both space and time. 
Future research should therefore develop tightly coupled deep learning– or optimization-based frameworks that explicitly exploit these raw data complementary characteristics, alongside optimized hardware designs for real-time, energy-efficient robotic applications.

\noindent \textbf{(4) Bio-inspired algorithm design.}
Event cameras inherently exhibit neuromorphic traits, making them well-suited for bio-inspired algorithms. 
Integrating SNNs, which process discrete spikes like the brain, enables sparse, low-power, and precise event-driven perception for tasks such as recognition and tracking \cite{ghosh2009spiking}. 
Bio-inspired models based on primate vision improve segmentation \cite{xu2023taming}, while neuromorphic control systems emulate sensorimotor loops for fast decisions \cite{liu2022toward}.
Nevertheless, most current innovations primarily focus on mimicking the biological mechanisms, developing sophisticated algorithms or hardware for high-level recognition and scene understanding. 
While inspiring, such approaches may not be optimal for tasks requiring strict real-time performance due to their significant computational overhead, or in some cases, may be impractical to realize. 
Future research should therefore explore alternative nature-inspired strategies—rather than strictly replicating biological mechanisms—by designing bio-inspired models that enable efficient event processing, and by optimizing neuromorphic hardware to effectively support these systems.

\noindent \textbf{(5) Hardware-software co-optimization techniques.}
Efficient event data processing requires tight hardware-software co-design.
Traditional hardware struggles with asynchronous, sparse event streams. 
Dedicated neuromorphic hardware—such as SNN chips, FPGA-based processors, and ASICs—can better support event-driven computation, improving efficiency \cite{Cao2024EventBoost, xu2023taming}.
On the software side, event-driven programming models and lightweight architectures are needed to reduce latency and redundancy, as conventional deep learning frameworks are not well-suited for event data \cite{gehrig2024low}.
However, most existing approaches focus on either hardware or software in isolation, overlooking cross-layer optimizations. Techniques such as dynamic resource allocation, memory access optimization, and hardware-aware model compression could further improve speed and energy efficiency. Future research should therefore emphasize holistic co-design of algorithms and neuromorphic hardware to realize real-time, low-power event-based vision systems for robotics, autonomous driving, and edge intelligence.
}

\vspace{-0.4cm}
\section{Conclusion} \label{9}
Event-based vision is a transformative approach for mobile embodied perception, offering high temporal resolution, low latency, and energy efficiency. This survey reviews event camera principles, event representations, algorithms, hardware/software acceleration, and diverse mobile applications. Despite advantages, challenges remain in event processing, sensor fusion, and real-time use on resource-limited agents. Future work should enhance hardware with advanced optics, develop neuromorphic processors for asynchronous data, and apply bio-inspired algorithms to boost perception. Integrating event cameras with LiDAR and radar will further broaden applications in dynamic settings. We hope this survey inspires research and practical deployment in the mobile embodied perception field.

\bibliography{ref}

% %%
% %% If your work has an appendix, this is the place to put it.
% \appendix









\end{document}